\documentclass[preprint,5p]{elsarticle}

\usepackage{amssymb}
\usepackage{amsmath}

\journal{Neural Networks}

\usepackage[utf8]{inputenc} 
\usepackage[T1]{fontenc}    
\usepackage{hyperref}       
\usepackage{url}            
\usepackage{booktabs}       
\usepackage{amsfonts}       
\usepackage{nicefrac}       
\usepackage{microtype}      
\usepackage{amsmath,amssymb,amsfonts}
\usepackage{graphicx}
\usepackage{textcomp}
\usepackage[table]{xcolor}
\usepackage{tikz}
\usetikzlibrary{tikzmark,calc}

\usepackage[ruled,vlined]{algorithm2e} 
\usepackage{setspace} 
\usepackage{lipsum}
\usepackage{bbm}
\usepackage{xcolor}
\usepackage{bm}
\usepackage{tabularx}
\usepackage{array}
\usepackage{booktabs}
\usepackage{cases}
\usepackage[skip=1pt, belowskip=0pt]{caption}
\usepackage{subcaption} 
\usepackage{amsthm}
\usepackage{arydshln}
\usepackage{enumitem}
\usepackage{mathtools}
\usepackage{amsmath}
\usepackage{comment}
\usepackage{enumitem}  
\usepackage{multirow}
\usepackage[normalem]{ulem} 
\usepackage{float}

\makeatletter
\def\BibTeX{{\rm B\kern-.05em{\sc i\kern-.025em b}\kern-.08em
    T\kern-.1667em\lower.7ex\hbox{E}\kern-.125emX}}
\allowdisplaybreaks
\definecolor{myblue}{rgb}{0,0,0}

\definecolor{myblue2}{rgb}{0.0328, 0.0539, 0.4758}
\definecolor{mygreen2}{rgb}{ 0.0328 0.4758 0.0539} 
\definecolor{mygreen3}{rgb}{ 0.0328 0.1758 0.0539} 
\definecolor{myred}{rgb}{0.4758, 0.0328, 0.0539}
\definecolor{myred2}{rgb}{0.75, 0.0328, 0.0539}
\definecolor{mypurple}{rgb}{0.7, 0.0328, 0.7539}

\newcommand{\balert}[1]{{\color{myblue}{#1}}}

\newcommand{\zerobf }{\boldsymbol 0}
\DeclareMathOperator{\E}{\mathbb{E}}

\newcommand{\Thetabf}{\bf{\Theta}}

\newtheorem{prop}{Proposition}
\newtheorem{rem}{\bf{Remark}}

\theoremstyle{remark}

\newtheorem{asmptn}{\bf{Assumption}}

\newenvironment{remark}
{\par\noindent \rem \begin{itshape}\noindent}
{\end{itshape} \vspace{3pt}}

\newcommand{\Abf}{\bm{A}}
\newcommand{\Bbf}{\bm{B}}
\newcommand{\Cbf}{\bm{C}}
\newcommand{\Dbf}{\bm{D}}

\newcommand{\AbfR}[1]{\bm{A}_{r_{#1}}} 
\newcommand{\BbfR}[1]{\bm{B}_{r_{#1}}} 
\newcommand{\CbfR}[1]{\bm{C}_{r_{#1}}} 
\newcommand{\DbfR}[1]{\bm{D}_{r_{#1}}} 
\newcommand{\AbfRo}{\bm{A}_{r}} 
\newcommand{\BbfRo}{\bm{B}_{r}} 
\newcommand{\CbfRo}{\bm{C}_{r}} 
\newcommand{\DbfRo}{\bm{D}_{r}} 
\newcommand{\Abfc}{\bm{A}_{c}} 
\newcommand{\Bbfc}{\bm{B}_{c}} 
\newcommand{\Cbfc}{\bm{C}_{c}} 
\newcommand{\Dbfc}{\bm{D}_{c}} 

\newcommand{\bS}{b_S}
\newcommand{\bT}{b_T}

\newcommand{\Qone}{Q_1}
\newcommand{\Qtwo}{Q_2}

\newcommand{\Wbf}{\bm{W}}
\newcommand{\Vbf}{\bm{V}}

\newcommand{\fbf}{\bm{f}}
\newcommand{\fbfbar}{\bm{\bar{f}}}
\newcommand{\ybf}{\bm{y}}

\newcommand{\mubfx}{\bm{\mu_m}}
\newcommand{\R}{\mathbb{R}}
\newcommand{\Z}{\mathbb{Z}}

\newcommand{\ctR}[1]{T_{r_{#1}}} 
\newcommand{\ctRo}{T_{r} }
\newcommand{\dtR}[1]{N_{r_{#1}}}
\newcommand{\vbfR}[1]{{\bf{v}}_{r_{#1}} }

\newcommand{\hS}{h_{(S)}}
\newcommand{\hT}{h_{(T)}}
\newcommand{\alphaS}{\alpha_{(S)}}
\newcommand{\alphaT}{\alpha_{(T)}}

\newcommand{\betaS}{\beta_{(S)}}
\newcommand{\betaT}{\beta_{(T)}}
\newcommand{\DeltaT}{\Delta_{T}}
\newcommand{\DeltaS}{\Delta_{S}}
\newcommand{\DeltaRatio}{\rho}
\newcommand{\NRatio}{\bar{\rho}}

\newcommand{\FuncF}{\mathcal{F}} 
\newcommand{\FuncG}{\mathcal{G}}

\newcommand{\vsig}[1]{v_{#1}} 
\newcommand{\vsigHat}[1]{\hat{v}_{#1}} 
\newcommand{\msefnc}{\operatorname{MSE}} 
\newcommand{\varfnc}{\operatorname{Var}} 
\newcommand{\covfnc}{\operatorname{Cov}} 
\newcommand{\dtN}[1]{N_{#1}} 
\newcommand{\dtNS}{N_{S}} 
\newcommand{\dtNT}{N_{T}} 
 
\newcommand{\partof}{\rightarrow}

\newcommand{\muS}{\mu_S}
\newcommand{\muT}{\mu_T}

\newcommand{\sigmaS}{\sigma_S}
\newcommand{\sigmaT}{\sigma_T}

\newcommand{\NS}{N_S}
\newcommand{\NT}{N_T}

\newcommand{\Umem}[1]{u[#1]}
\newcommand{\Uad}[1]{v[#1]}
\newcommand{\UmemNoArg}{u} 
\newcommand{\UadNoArg}{v} 

\newcommand{\Sil}[1]{s_{out}[#1]}
\newcommand{\Sl}[1]{\bm{s}_{out}[#1]}
\newcommand{\Slbefore}[1]{\bm{s}_{in}[#1]}

\newcommand{\Slo}{\bm{s}_{out}}
\newcommand{\Slbeforeo}{\bm{s}_{in}}

\newcommand{\vbf}{\bm{v}}
\newcommand{\sbf}{\bm{s}}

\newcommand{\Ibf}{\bm{I}}
\newcommand{\Hbf}{\bm{H}}
\newcommand{\Gbf}{\bm{G}}

\newcommand{\Hbfv}{\bm{H}_{v}}
\newcommand{\Hbfs}{\bm{H}_{f}}
\newcommand{\Hbff}{\bm{H}_{i}}
\newcommand{\Hbfr}{\bm{H}_{r}}

\newcommand{\HbfvS}{\bm{H}_{v, (S)}}
\newcommand{\HbfsS}{\bm{H}_{f, (S)}}
\newcommand{\HbffS}{\bm{H}_{i, (S)}}
\newcommand{\HbfrS}{\bm{H}_{r, (S)}}
\newcommand{\HbfiS}{\bm{H}_{k, (S)}} 
\newcommand{\HbfvT}{\bm{H}_{v, (T)}}
\newcommand{\HbfsT}{\bm{H}_{f, (T)}}
\newcommand{\HbffT}{\bm{H}_{i, (T)}}
\newcommand{\HbfrT}{\bm{H}_{r, (T)}}
\newcommand{\HbfiT}{\bm{H}_{k, (T)}} 

\mathchardef\myhyphen="2D

\begin{document}

\begin{frontmatter}



\title{Zero-shot Temporal Resolution Domain Adaptation for Spiking Neural Networks} 


\author[0]{Sanja~Karilanova}
\author[1,2]{Maxime~Fabre}
\author[1]{Emre~Neftci}
\author[0]{Ayça~Özçelikkale}
\affiliation[0]{organization={Uppsala University},
            country={Sweden}}
\affiliation[1]{organization={RWTH and Forschungszentrum Jülich},
            country={Germany}}
\affiliation[2]{organization={University of Groningen},
            country={Netherlands}}

    

\begin{abstract}
Spiking Neural Networks (SNNs) are biologically-inspired deep neural networks that efficiently extract temporal information while offering promising gains in terms of energy efficiency and latency when deployed on neuromorphic devices. SNN parameters are sensitive to temporal resolution, leading to significant performance drops when the temporal resolution of target data during deployment is not the same as that of the source data used for training, especially when fine-tuning with the target data is not possible during deployment. To address this challenge, we propose three novel domain adaptation methods for adapting neuron parameters to account for the change in time resolution without re-training on target time resolution. The proposed methods are based on a mapping between neuron dynamics in SNNs and State Space Models (SSMs) and are applicable to general neuron models. We evaluate the proposed methods under spatio-temporal data tasks, namely the audio keyword spotting datasets SHD and MSWC, and the neuromorphic image NMINST dataset. Our methods provide an alternative to-and in most cases significantly outperform-the existing reference method that consists of scaling only the time constant. Notably, when the temporal resolution of the target data is double that of the source data, applying one of our proposed methods instead of the benchmark achieves classification accuracy of $89.5\%$ instead of $53.0\%$ on SHD, $93.6\%$ instead of $38.8\%$ on MSWC and  $98.5\%$ instead of $97.2\%$ on NMNIST. Moreover, our results show that high accuracy on high temporal resolution data can be obtained by time-efficient training on lower temporal resolution data. 
\end{abstract}
\end{frontmatter}

\section{Introduction}
\label{sec:introduction}
\begin{figure*}
    \centering
    \includegraphics[width=1\linewidth]{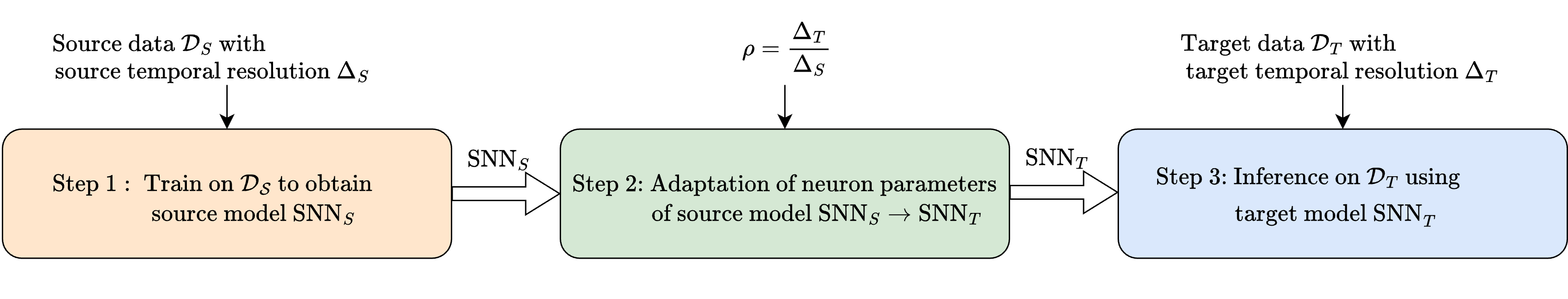} 
    \caption{Overview of the set-up. We investigate how models trained on data with a given source temporal resolution can be adapted for data with a different target temporal resolution, both in the Fine-to-Coarse and Coarse-to-Fine deployment directions. }
    \label{fig:main_set_up_summary}
\end{figure*}

Domain adaptation addresses the scenarios in which a model is confronted with a domain shift after deployment, while its task remains the same, such as face recognition under different lightning conditions or poses   \cite{domain_adaptation_brief_review, liang2023comprehensivesurveytesttimeadaptation}. The domain shift refers to a change in feature space between the source domain on which the model was trained, and the target domain on which the model is evaluated. The change in feature space has been studied from various aspects, particularly for visual data \cite{Patel_2015, zubić2024state}, but has rarely been explored with respect to temporal resolution shift for temporal data. Our work focuses precisely on this under-explored aspect, investigating feature discrepancies caused by differences in temporal resolution between the source and target data.
While domain adaptation typically involves transferring knowledge between the source and target data by re-training parts of the model or the entire model using samples from target domain, \cite{Patel_2015, Sun_2019_CVPR} \cite{beei}, we focus on the extreme case of no retraining on  target domain.

In a conventional data acquisition set-up,  where values of signals of interest are recorded at regular intervals, temporal resolution is primarily determined by the time interval between successive samples. For instance, the frame rate of a video stream or the sampling rate of an audio recording determines the temporal resolution. 
In the case of event-based sensing, e.g. \cite{Gallego2022}, where changes in the scene are recorded with time-stamps, temporal resolution is determined by the temporal accuracy of time-stamps.

Various practical limitations lead to data with varying temporal resolution.
For instance, limited memory capacity and power constraints often require lower sampling rates of data acquisition or storage devices \cite{Dieter_2005, ur2016BigDataCompress}. Additionally, sensor quality and bandwidth limitations further restrict temporal resolution, as high-resolution sensors and real-time transmission demand more resources \cite{chen2021distributed, park2021communication}.
Consequently, changes in the temporal resolution of data have been the focus of a wide range of studies, such as increasing of temporal resolution of videos \cite{Zuckerman2020}, optimization of sensor networks' sensing rates \cite{Ankur2004}, processing of irregularly sampled data \cite{rubanova2019latent},  optimal combination of high- and low-rate time-series data \cite{Bae2024}. In the case of event-based cameras, effect of bandwidth and event-rate constraints  on the temporal resolution have been studied \cite{gehrig2022highresolutioneventcamerasreally, zubić2024state}.

In this paper,  we are interested in the scenario where the temporal resolution of the source data which the model has been trained on, and the target data which the model is to be used on is different. 
We investigate both the scenario of Fine-to-Coarse deployment, where the change is from high to low resolution, and Coarse-to-Fine deployment, where the change is from low to high resolution.
We consider the setting where it is not possible to fine-tune the model at the edge on the data with the target temporal resolution, for instance when real-time inference is needed on the current sample from target domain. Hence, we consider temporal resolution domain adaptation with no re-training on the target dataset, see Figure \ref{fig:main_set_up_summary}.

Applications of these scenarios appear in different edge processing scenarios, where small resource-constrained devices with different sensing and computing capabilities must process their data at the edge of
a wireless system with varying degrees of support from a central processing or coordination unit \cite{chen2021distributed, park2021communication}.  
A possible application of Coarse-to-Fine deployment is encountered in communication constrained distributed scenarios with a central learning unit. 
In such scenarios, sensor data is sent to a central learning unit with reduced temporal resolution due to bandwidth or power limitations \cite{chen2021distributed, park2021communication}.
As a result, the central model is trained on coarse data and then deployed to edge devices, where higher-resolution data is available for inference.
A possible application of Fine-to-Coarse deployment is encountered where the sensors have to switch to low sampling rates to reduce power consumption \cite{Dieter_2005,Ankur2004}. This leads to having target data at edge with coarser temporal resolution than the source data used for training the deployed models at the central learning unit.

We consider the above temporal domain adaptation problems under Spiking Neural Networks (SNN) models.  
SNNs are biologically-inspired neural networks with stateful recurrent neurons, and binary spike-based communication \cite{MAASS19971659}. SNNs provide an attractive framework for processing data with encoded temporal information due to their ability to represent and learn spatiotemporal features \cite{bittar2022surrogate, yik2024neurobench, Braille_SNN, eshraghian2023trainingspikingneuralnetworks, ma2025spikingneuralnetworkstemporal}.
Like most machine learning algorithms, the current mainstream pipeline for SNNs relies on training at high-power computational units, such as GPUs, and then deploying and running on low-power edge devices \cite{10177729}. SNNs, when deployed at neuromorphic edge hardware, are especially valuable because they are designed to mimic the brain’s energy-efficient processes, offering significant reductions in power consumption and latency \cite{9395703, caccavella2023lowpower, Braille_SNN}.
SNNs are particularly effective in event-based systems, where event-based sensors, e.g. event-cameras, record changes only as they occur, rather than at fixed time intervals, leading to optimized energy use and reduced data redundancy \cite{Gallego2022}.

A common practical training set-up for SNNs is the standard digital clock-based SNNs, where the temporal resolution of the data corresponds directly to the processing resolution at which the model operates. 
In this setup, the possible discrepancies in the temporal resolution of the source and target data constitute a significant challenge since trained SNN model parameters are strongly dependent on the temporal resolution of the source data. 
Hence, an SNN model that has been trained using data with a given temporal resolution may not necessarily perform well in inference on data from another sensory source even if it is recording of the same phenomena, e.g., see Section~\ref{sec:numericalResults}.

Training on sequences with high temporal resolution is desired due to the possible inclusion of more information and therefore better performance of the model \cite{gehrig2024low}. However, this comes with a cost. Difficulties arise when training SNNs on long sequences such as exploding and vanishing gradients and other instabilities due to the recurrent and saturating nature of SNNs \cite{zucchet2024recurrentneuralnetworksvanishing}. Another limitation while training on long sequences is the significant training time as the temporal resolution dimension approximately linearly increases the training time and computational complexity \cite{pereznieves2022sparse}. Hence, the development of temporal adaptation methods enables training models in a fast and stable manner on low temporal resolution data, while still ensuring high performance during inference on high temporal resolution data. This setting supports energy, time, and computation efficient training pipeline.

Considering both the temporal domain adaptation problem as well as challenges of training on sequences with fine temporal resolution, the main contributions of this paper are:
\begin{itemize}
    
    \item We propose three novel methods for adapting SNN parameters that govern the neuron dynamics for changes in temporal resolution of the data without using any data samples with new temporal resolution.
    
    \item The performance of the proposed methods is demonstrated using the audio datasets SHD  \cite{shddataset}  and MSWC \cite{mswcdataset} and the vision NMNIST dataset \cite{orchard2015converting} .
    
    \item The results show that in a wide range of scenarios, e.g. coarse-to-fine scenarios for all the above datasets,  our proposed methods provide significant performance gains compared to using no temporal resolution domain adaptation method or using the existing benchmark method from the literature.
    
    \item We show that utilizing version of the data with low temporal resolution for pre-training, constitutes a promising training approach for improving computational training efficiency in SNNs.
\end{itemize}

We now provide an overview of the organization of the paper. Section \ref{sec:related_work} provides a review of the related work.  The main problem studied in this article, i.e., the temporal domain adaptation problem, is formulated in Section \ref{sec:problem_statement}.  Section \ref{sec:proposed_method_intro} presents the proposed methods. Numerical results illustrating the performance of our methods are provided in Section \ref{sec:numericalResults} and Section \ref{sec:ablation_study}.  
Finally, in Section \ref{sec:discussions} and Section \ref{sec:conclusions} we present further discussions and a summary of our conclusions, respectively. 

\section{Related Work}
\label{sec:related_work}

\begin{figure}
    \centering
    \includegraphics[width=1\linewidth]{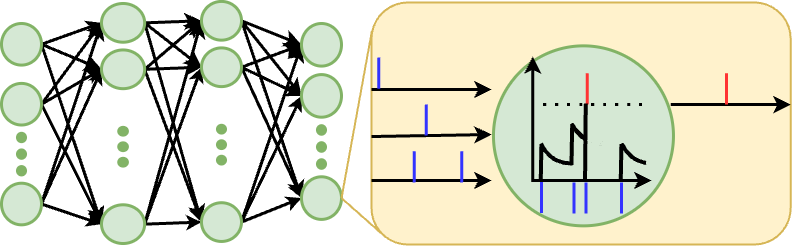} 
    \caption{Feedforward SNN model architecture consisting of four layers, each with arbitrary number of neurons. An example neuron dynamics given as a zoom-in on a neuron.}
    \label{fig:snn_neuron_diagram}
\end{figure}

The Leaky-Integrate-and-Fire (LIF) neuron model is one of the most simple and widely used neuron models in the context of SNNs. When input stimuli in the form of spikes are applied to the neuron, the membrane potential increases and when a certain threshold is reached it produces an output spike. On the other hand, in the absence of input stimuli, the membrane potential decays exponentially \cite{bittar2022surrogate}, see Figure \ref{fig:snn_neuron_diagram}. 
The vanilla LIF neuron has only one learnable parameter, which is the membrane potential decay (the leak), and this parameter has an explicit exponential dependence on the time step length of the data. Hence, when dealing with a change of temporal resolution, intuitive way to adapt the LIF neuron dynamic is to scale the leak parameter according to the change. 
In \cite[Table 12]{he2020comparing} the authors explore this method of temporal domain adaptation using the simple LIF neuron. 
On the other hand, in \cite{caccavella2023lowpower}, the authors use Integrate-and-Fire (IF) neuron which is a special simpler case of LIF neuron without leak. Hence alternative to parameters adaptation, they employ a graded spike network and different normalization techniques for domain adaptation with promising results. 
However, in order to capture more complex temporal information in data, more complex neuron models such as the adaptive LIF (adLIF) neurons have shown superior performance \cite{bittar2022surrogate, baronig2025advancingspatiotemporalprocessingspiking, adLIFNature, ma2025spikingneuralnetworkstemporal}. In the case of more complex neuron models however, the parameter might not have explicit dependence on the time-length of the data, or even any dependence at all.

In this case, it is unclear how these parameters need to be scaled to depict the change of temporal resolution, and to the best of our knowledge, this issue has not been thoroughly addressed in the existing literature.

An important building block of our proposed approach is State Space Models (SSMs), which is a general  framework for  modeling dynamic systems. Their neural network implementations, i.e. deep SSMs, 
raised as an attractive alternative to the conventional deep recurrent models.  
It was shown that for appropriate conditioning on the state matrix dynamics, deep SSMs can handle long-range dependencies while offering stable and computationally efficient training
\cite{wang2024state}. 
Limited explorations of temporal resolution change have been performed using SSMs and traditional data \cite[Table 2]{smith2022simplified}. 
Deep SSMs have been used for event-based data in a number of works \cite{schöne2024scalable, zubić2024state}
Similar to the leak adaptation in LIF neurons, explicit correlation can be addressed between significantly simplified version of SSM and the temporal dynamics between spikes. \cite{schöne2024scalable} explores this concept by processing incoming spikes with adapted leak based on the time-length between consecutive spikes. 
Alternatively, \cite{zubić2024state} explored varying-resolution aware training of deep SSMs for vision dataset, where the training includes masking inputs outside specific frequency ranges and incorporating an additional loss function to penalize high-frequency activities.

Interesting intersection of work is emerging between SNNs and deep SSMs. 
One common line of work is embedding spiking dynamics in deep SSMs where at each time step the output activity of the SSM is processed by a firing threshold function
\cite{stan2023learning, bal2024rethinkingspikingneuralnetworks, karilanova2025statespacemodelinspiredmultipleinput, karilanova_reset}.
Alternatively, in \cite{du2024spiking} they combine SNN with SSMs by adding a spiking layer to the SSM.
Nevertheless, the challenge of drawing lines of correspondence between SNNs and linear SSMs is still present due to the non-linearity when it comes to SNNs. Our proposed methods for temporal domain adaptation for SNNs are based on this drawn parallel and the theory developed for adaptation in SSMs. See Table~\ref{tab:related_work_summary} for a summary of the prior works discussed in this section.

\begin{table*}[]
    \centering
    \newcolumntype{l}{>{\centering\arraybackslash}p{3.3cm}}
    \newcolumntype{m}{>{\centering\arraybackslash}p{3cm}}
    \newcolumntype{n}{>{\centering\arraybackslash}p{3.9cm}}
    \begin{tabular}{lmmmn}
    \hline
        Method 
        & Neuron model
        & Adaptation 
        & Datasets 
        & Metrics \\
    \hline 
    This article &  Generalized spiking neuron including adLIF, LIF, IF, SSM  & Expectation, Integral, and Euler Methods & SHD, MSWC, N-MNIST &  Classification accuracy in varying temoral resolution \\
    \hline 
    Comparing SNNs and ANNs \cite[Table 12]{he2020comparing} & LIF, rate-coded RNN and rate-coded LSTM  & No-adaptation & N-MNIST &  Classification accuracy in varying temporal resolution \\
    \hline
    Event-driven inference \cite{caccavella2023lowpower} & IF & Layer normalization & class ``faces''  of N-CALTECH101 & mAP on varying time-window of integration \\
    \hline
    EventSSM \cite{schöne2024scalable} & SSM & Event-by-event processing & SHD, SSC, DVS-Gesture & Classification accuracy (both in train and test stage event-by-event processing used) \\
    \hline
    SSM for event cameras \cite{zubić2024state} & SSM & Low-pass bandlimiting, and SSM-ViT block &Gen1 and 1 Mpx& mAP on varying evaluations frequency\\ 
    \hline
    \end{tabular}
    \caption{Overview of works which include aspects of zero-shot temporal adaptation in SNN or deep-SSM models.}
    \label{tab:related_work_summary}
\end{table*}

\section{Problem Statement}
\label{sec:problem_statement}

In Section \ref{sec:popular_neurons} we present a popular neuron model commonly used in the SNN literature to establish a foundation for the generalized neuron model in Section \ref{sec:general_neuron}. 
We outline the temporal domain adaptation problem that this article focuses on in Section \ref{sec:prob_statement_temp_res}.

\subsection{Popular spiking neuron model}
\label{sec:popular_neurons}
The discrete-time adaptive Leaky-Integrate and Fire (adLIF) neuron is defined as \cite{bittar2022surrogate}
\begin{subequations}
\label{eqn:adLIFneuron_intro}
\begin{alignat}{10}
\! \Umem{t+1} =& \alpha (\Umem{t} - \theta \Sil{t} )
                + (1-\alpha) (\Wbf \Slbefore{t} +\Vbf \Sl{t}) \notag \\
                &- (1-\alpha) \Uad{t} \label{eqn:adLIF_mem} \\ 
\! \Uad{t+1} =& a \Umem{t} + \beta \Uad{t}
                    + b \Sil{t} 
                    \label{eqn:adLIF_adap}\\
\Sil{t} =& g_\theta(\Umem{t})  \label{eqn:adLIF_spk}
\end{alignat}
\end{subequations}
where $\Slbeforeo \in \Z^{L_b \times 1}$ are the input spikes from the neurons in the previous layer and $\Wbf \in \R^{1 \times L_b}$ is the weight matrix associated with their connections to the neuron of observation, while $\Slo \in \Z^{L_c \times 1}$ are the output spikes of all the other neurons in the current layer and $\Vbf \in \R^{1 \times L_c}$ with zero diagonal is their connections to the neuron of observation. Here, $L_c$ and $L_b$ are the number of neurons in the current and the previous layer, respectively. Furthermore, $g_\theta(\cdot)$ describes the spiking behavior with the firing threshold $\theta \in \R^{1 \times 1}$; $\alpha=\exp(\Delta / \tau_{mem})$ represents the decay of the membrane potential over time such that $\Delta$ corresponds to the duration of a timestep and $\tau_{mem}$ is a membrane time constant; $\beta=\exp(\Delta / \tau_{syn})$ represents the decay of the adaptive variable with adaptation constant $\tau_{syn}$; the parameters $a$ and $b$ are without explicit dependence on the discrete time step length $\Delta$. All parameters $\alpha,\beta,a,b$ are generally learnable during training within given ranges.

Note, the weight vector $\Vbf$ corresponds to the recurrent connections between the neurons in the same layer. Recurrence represented by $\Vbf$ is a characteristic of the network rather than the neuron.

\subsection{General spiking neuron model}
\label{sec:general_neuron}
In this section, we introduce a discrete-time general $n$-dimensional spiking neuron model, which generalizes the two-dimensional adLIF neuron \eqref{eqn:adLIFneuron_intro}. 
This generalized model is central to the proposed methods of this paper. It is defined as
\begin{subequations}
\label{eqn:general_LIF_intro}
\begin{align}
\! \vbf[t+1] &=  \Hbfv \vbf[t] 
                    + \Hbfs \Sil{t}
                    + \Hbff \Wbf \Slbefore{t}
                    + \Hbfr \Vbf \Sl{t} 
                    \label{eqn:general_LIF_intro:line1}\\
 \Sil{t} &= g_{\Thetabf}(\vbf[t]) 
         =\begin{cases} 
          1  \text{ if }\vbf[t] \in \Thetabf \\
          0  \text{ otherwise}
          \end{cases}
\label{eqn:general_LIF_intro:line2}
\end{align}
\end{subequations}
where
$\vbf \in \R^{n \times 1}$ is the state variable, while $\Hbfv \in \R^{n \times n}$ and  $\Hbfs, \Hbff, \Hbfr \in \R^{n \times 1}$ are matrices containing the parameters of the individual neuron.
The subscripts of the $\Hbf$ matrices, i.e. $v,f,i$, and $r$,  indicate state, feedback, input and recurrence, respectively. The spiking behavior is described by $g_{\Thetabf}(\cdot)$, where the neuron spikes when the state enters in the region described by $\Thetabf$. The region $\Thetabf$ depends on the neuron model.

Many popular spiking neuron models can  be expressed as special cases of the general spiking neuron model in \eqref{eqn:general_LIF_intro} including
the LIF neuron \cite{Neuronal_Dynamics_book} with $n=1$ state variable, 
the adLIF neuron \cite{bittar2022surrogate} with $n=2$, 
the Izikevich neuron \cite{izhikevich2003simple} with $n=2$, 
the Hodgkin-Huxley \cite{Hodgkin1952} with $n=4$, 
multi-compartment neurons with $2$ compartments each with $n=2$    \cite{Zhang_Yang_Ma_Wu_Li_Tan_2024}. Below, we show how the $n=2$ dimensional adLIF neuron can be expressed as a special case of \eqref{eqn:general_LIF_intro}.

The adLIF neuron, in matrix form, can be written as:
\begin{align}
\label{eqn:adLIFneuron_rearanged}
\! \begin{bmatrix} \Umem{t+1} \\ \Uad{t+1} \end{bmatrix}
 =
&\begin{bmatrix}\alpha & -(1-\alpha)\\ a &\beta\end{bmatrix}
\begin{bmatrix} \Umem{t} \\ \Uad{t} \end{bmatrix}
+
\begin{bmatrix} - \alpha \theta \\ b \end{bmatrix}
\Sil{t} \notag
\\
& +
\begin{bmatrix} 1-\alpha \\ 0\end{bmatrix}
\Wbf \Slbefore{t}
+
\begin{bmatrix} 1-\alpha \\ 0\end{bmatrix}
\Vbf \Sl{t}.
\end{align}
Thus,  the adLIF neuron \eqref{eqn:adLIFneuron_intro} with two state variables $\vbf=\begin{bmatrix} \UmemNoArg, \UadNoArg \end{bmatrix}^T$ is a special case of the general model in \eqref{eqn:general_LIF_intro} where
\begin{align}
\label{eqn:adLIF_H_matrices}
\Hbfv &= \begin{bmatrix}\alpha & -(1-\alpha)\\ a &\beta\end{bmatrix},
\Hbfs = \begin{bmatrix} - \alpha \theta \\ b \end{bmatrix}, 
\Hbff = \begin{bmatrix} 1-\alpha \\ 0\end{bmatrix}, 
\text{and} \notag \\
\Hbfr &= \begin{bmatrix} 1-\alpha \\ 0\end{bmatrix}.
\end{align}

\begin{remark}
The general spiking neuron model in \eqref{eqn:general_LIF_intro} does not encompass all possible neuron formulations. For instance, it does not capture:
(i) hard reset mechanism, where the neuron state is set to a predefined reset state value when spiking occurs \cite{davies_Loihi_2018} rather then subtracting the  spike using $\Hbfs$;
(ii) varying adaptive spiking threshold, where the threshold varies over time \cite{CHEN2022189}, rather than being defined by a fixed spiking region $\Thetabf$.
\end{remark}

\subsection{Problem Formulation - Temporal resolution domain adaptation for general spiking neuron model}
\label{sec:prob_statement_temp_res}

\subsubsection{Event-based data}
\label{sec:event_based_data}

\balert{In this section, we state the event-based data representation we focus on in this article.} 
Event-based data can be represented as asynchronous or synchronous \balert{\cite{Gallego2022, chakravarthi2024recent}}. In the asynchronous representation, data is provided as a list of events each with  their precise time of occurrence. In the synchronous representation, data is provided in a time-binned format as spike sequences of a given length, each bin  populated with all events occurring during this period of time. The first type of representation is naturally adopted when data comes from event-based sensors, such as NMNIST \cite{orchard2015converting} and the second type of data representation is typically used after post-processing of data obtained by conventional sensors, such as Spiking Heidelberg Digits (SHD) dataset \cite{shddataset}, or by post-processing of data from event-based sensors. 

The standard approach for processing event-based data with SNNs is to use digital clock-based SNNs, which require the data to be in an effectively synchronous form for processing.  Hence, if the data is asynchronous, it is converted into a synchronous form, such as by binning events on regular time intervals. 
\balert{In this article, we focus on such synchronous data representations where events that occur during a temporal window of fixed length are accumulated to create sequences of spikes \cite{Gallego2022, liu2018adaptive}. This type of representations have been used in a large number of earlier SNN works,  including studies utilizing temporal bins for audio signals \cite{bittar2022surrogate, hammouamri2023learningdelaysspikingneural} and frame-based event representation  for  spatio-temporal data \cite{zubić2024state, caccavella2023lowpower, he2020comparing}.
}

\subsubsection{Temporal resolution}
\label{sec:temporal_resolution}

\balert{In this section, we define the notion of temporal resolution used in this article.}
Consider a continuous-time signal $\sbf(t)$, $0 \leq t \leq t_c$ with a duration $t_c$ seconds, representing a phenomenon of interest, such as a spoken keyword.  We denote by $\sbf \in \Z^{N \times 1}$ an integer-valued discrete-time sequence extrapolated from $\sbf(t)$ where $N$ is the number of time steps. Hence, for a regular discretization of the time-axis, which we focus on here, a single time step in $\sbf$ will represent a time duration of $\Delta_N=t_c/N$ seconds.

Suppose that we have two discrete-time data sequences associated with the same continuous signal, each with a different number of time steps. Let us refer to them as  
$\sbf_{S} \in \Z^{\dtNS \times 1}$ and $\sbf_{T} \in \Z^{\dtNT \times 1}$ with $\dtNS \neq \dtNT$, and thus $\DeltaS \neq \DeltaT$. Hence, $\sbf_{S}$ and $\sbf_{T}$  correspond to sequences with different time resolutions.
The sequences $\sbf_{S}$ and $\sbf_{T}$ can be obtained from $\sbf(t)$ in various ways. One is to obtain both sequences directly from $\sbf(t)$ using different encoding parameters. Another is to obtain $\sbf_{S}$ directly from $\sbf(t)$, while obtaining $\sbf_{T}$ by transforming $\sbf_{S}$, or vice versa.

\subsubsection{Model adaptation}
\label{sec:Model_adaptation_assumption}

We consider the scenario where the SNN model is pre-trained on data with time resolution $\DeltaS$ but we would like to use this model on data with temporal resolution $\DeltaT$, where $\DeltaT \neq \DeltaS$. Consequently, we refer to the training dataset as source data and the latter as target data. 
Similarly, we refer to $\DeltaS$ and $\DeltaT$ as the temporal resolution of the source and target data, respectively.

During SNN training, two groups of parameters are learned: parameters of the neuron dynamics for each neuron, $\HbfvS,\HbfsS, \HbffS, \HbfrS$, and the  weights of the synaptic connections between the neurons, $\Wbf, \Vbf$.  
 %
Our main goal in this article is to adjust the model parameters such that we obtain high-accuracy results on target data.
We propose to achieve this goal by keeping the dynamics of the state variable $\vbf$ of neurons, see \eqref{eqn:general_LIF_intro}, as similar as possible under the source and target data. This means that the dynamics of the state variable $\vbf$ is preserved even when the temporal resolution of the data changes.

We consider this problem under two constraints:  
First, we are only allowed to adapt the neuron's dynamic parameters, $\Hbf_{k}$ for $k=v,i,f,r$. Hence, SNN architecture and synaptic connections are unchanged. 
Second, the only known information about the target data is the ratio between the target and the source temporal resolution. Hence, we do not rely on any post-deployment re-training on the target dataset, which can be costly.
%
Our results in Section~\ref{sec:numericalResults} show that promising performance can be obtained even under this constrained scenario, supporting the central role of neuron dynamics in capturing time-dependent behavior.  

In summary, the main goal of this article is to find a mapping from the source to the target feature space in the form of 
\begin{align}
    \label{eqn:mapping_formal}
    \HbfvT, &\HbffT, \HbfsT, \HbfrT \notag \\ 
    &= {\cal{M}}(\DeltaRatio, \HbfvS, \HbffS, \HbfsS, \HbfrS),
\end{align}
where the mapping ${\cal{M}}(.)$ represents the adaptation of the temporal resolution domain, and the parameter $\DeltaRatio$ is given by
\begin{align}
\label{eqn:rho_definition}
\DeltaRatio = \frac{\DeltaT}{\DeltaS} = \frac{\dtNS}{\dtNT} = \frac{1}{\NRatio}
\end{align} 
 defining the ratio between the temporal resolutions. Here, we have also defined $\NRatio$, the  reciprocal of $\DeltaRatio $, for a more convenient notation in later chapters.
We consider the following scenarios: 
\begin{itemize}
    \item Coarse-to-Fine deployment: The model is pre-trained on source data with coarse temporal resolution and tested on target data with fine temporal resolution, i.e.,  $\DeltaS > \DeltaT$, thus $\DeltaRatio < 1$.
    \item Fine-to-Coarse deployment: The model is pre-trained on source data with fine temporal resolution and tested on target data with coarse temporal resolution, i.e.,  $\DeltaS < \DeltaT$, thus $\DeltaRatio > 1$.
\end{itemize}

\section{Methods}
\label{sec:proposed_method_intro}

As a background for our work, in Section \ref{sec:map_snn_non_lssm}, we begin by expressing the SNN neuron model as a non-linear SSM model. This provides a foundation for Section \ref{sec:map_snn_lssm}, where we propose an approximate correspondence between the SNN neuron model and linear SSM model. The proposed temporal resolution domain adaptation methods, presented  in Section \ref{sec:proposedMethods:main}, are based on this approximate correspondence. In Section \ref{sec:Method_TimeConst} we present a simple benchmark method.

\subsection{Preliminaries - Correspondence between SNNs and non-linear SSMs}
\label{sec:map_snn_non_lssm}
In this section, we express the SNN neuron model as a specific case of non-linear SSMs. 

A general discrete-time nonlinear SSM can be written as \cite{OkuyamaYoshifumi2014DCS}:
\begin{subequations}
\label{eqn:non-lssm:generic}
\begin{align}
\vbf[t+1] &= \FuncF(\vbf[t], \fbf[t]) \\
\ybf[t] &= \FuncG(\vbf[t], \fbf[t]) 
\end{align}
\end{subequations}
where $\vbf[t] \in \R^{n \times 1}, \fbf[t] \in  \R^{r \times 1}$, $\ybf[t] \in \R^{p \times 1}$ represent the state vector, the input to the system and the output of the system, respectively, and 
$\FuncF(\cdot)$ and $\FuncG(\cdot)$ are possibly non-linear functions. Here, $n$ is the dimension of the system's state, $r$ is the dimension of the system's input, and $p$ is the dimension of the system's output. 
The general neuron model in \eqref{eqn:general_LIF_intro} can be expressed as a non-linear SSM as in \eqref{eqn:non-lssm:generic} using
\begin{subequations}
\label{eqn:non-lssm_and_snn:correct_correspondence}%
\begin{align}
\!\FuncF(\vbf[t], \fbf[t]) &= \Hbfv \vbf[t] 
    + \Hbfs g_{\Thetabf}(\vbf[t])
    + \begin{bmatrix} \Hbff, \Hbfr \end{bmatrix} \fbf[t] \\
\!\FuncG(\vbf[t], \fbf[t]) &= g_{\Thetabf}(\vbf[t])
\end{align}
\end{subequations}
where $\fbf[t]=\begin{bmatrix}\Wbf \Slbefore{t}, \Vbf \Sl{t} \end{bmatrix}^T$ is considered external input since $\Slbefore{t}$ parameterize the spikes from the previous layer and $\Sl{t}$ parameterize the spikes from all the other neurons in the current layer. Here, $\ybf$ represents the output spike of the neuron.

Since our ultimate goal  is to find a correspondence between SNNs and linear SSMs, we move one step closer to a linear SSM by considering  separable $\FuncF(\cdot)$ and $\FuncG(\cdot)$,  i.e., 
\begin{subequations}
\label{eqn:non-lssm:constaint}
\begin{align}
\vbf[t+1] &= \FuncF_A(\vbf[t]) + \FuncF_B(\fbf[t])\\
\ybf[t] &= \FuncG_C(\vbf[t]) + \FuncG_D(\fbf[t])
\end{align}
\end{subequations}
where $\FuncF_A(\cdot), \FuncF_B(\cdot), \FuncG_C(\cdot)$ and $\FuncG_D(\cdot)$ are possibly non-linear functions.
The generalized neuron in \eqref{eqn:general_LIF_intro} can be  expressed as a non-linear SSM with separable functions as in \eqref{eqn:non-lssm:constaint} using
\begin{subequations}
\label{eqn:non-lssm-constraint_and_snn:correspondence}
\begin{align}
    \FuncF_A(\vbf[t]) &= \Hbfv \vbf[t] + \Hbfs g_{\Thetabf}(\vbf[t])\\
    \FuncF_B(\fbf[t]) &= \begin{bmatrix} \Hbff, \Hbfr \end{bmatrix} \fbf[t] \\
    \FuncG_C(\vbf[t]) &= g_{\Thetabf}(\vbf[t])   \label{eqn:non-lssm-constraint_and_snn:correspondence:C} \\
    \FuncG_D(\fbf[t]) &= \zerobf
    \label{eqn:non-lssm-constraint_and_snn:correspondence:D}
\end{align}
\end{subequations}
where $\fbf[t]=\begin{bmatrix}\Wbf \Slbefore{t}, \Vbf \Sl{t} \end{bmatrix}^T$.

\subsection{Preliminaries - Correspondence between SNNs and linear SSMs}
\label{sec:map_snn_lssm}
As linear SSMs enable analytical investigations and form the basis of our adaptation methods, in this section, we propose an approximate correspondence between the general SNN neuron model in \eqref{eqn:general_LIF_intro} and linear SSMs. 
Despite SNNs being strictly non-linear SSMs, this approximate correspondence can be utilized to achieve promising results for temporal domain adaptation, as demonstrated by the results in Section \ref{sec:numericalResults}.

A linear SSM in discrete-time can be written as \cite{Gajic}:
\begin{subequations}
\label{eqn:lssm:generic}
\begin{align}
    \vbf[t+1] = \Abf \vbf[t] + \Bbf \fbf[t]
\\
    \ybf[t] = \Cbf \vbf[t] + \Dbf \fbf[t],
\end{align} 
\end{subequations}
where the vectors $\vbf[t], \fbf[t], \ybf[t]$ are respectively the state vector, input vector and output vector. 
Here, the matrix $\Abf \in \R^{n \times n}$ describes the internal behaviour of the system, while matrices $\Bbf \in \R^{n \times r}, \Cbf \in \R^{p \times n}, \Dbf \in \R^{p \times r}$ represent connections between the external world and the system. These matrices represent arrays of real scalar numbers. The system is assumed to be time invariant which means the scalar numbers are constant over time.

To find an approximate correspondence between the general neuron model in \eqref{eqn:general_LIF_intro} and the linear SSM in \eqref{eqn:lssm:generic},  we first map the nonlinear function in the separable non-linear SSM \eqref{eqn:non-lssm:constaint} to the matrices describing linear relation in \eqref{eqn:lssm:generic}. The mapping is intuitive, 
\begin{subequations}
\label{eqn:non-lssm-constraint_and_SSM:correspondence}
\begin{align}
\Abf \vbf[t] &\partof \FuncF_A (\vbf[t]) \\
\Bbf \fbf[t] &\partof \FuncF_B (\fbf[t])\\
\Cbf \vbf[t] &\partof \FuncG_C (\vbf[t])\\
\Dbf \fbf[t] &\partof \FuncG_D (\fbf[t]) 
\end{align}
\end{subequations}
where the symbol $\partof$ indicates that the left-hand side linear term is best described as part of the dynamics of the right-hand side possibly nonlinear function.
Next, we associate the general neuron matrices $\Hbfv,\Hbfs, \Hbff, \Hbfr$ in  \eqref{eqn:general_LIF_intro}, to the linear matrices $\Abf, \Bbf, \Cbf, \Dbf$ in \eqref{eqn:lssm:generic}. 
Using the correspondences in \eqref{eqn:non-lssm-constraint_and_snn:correspondence} and \eqref{eqn:non-lssm-constraint_and_SSM:correspondence}, the following mapping arises as a candidate: $\Abf=\Hbfv + \Hbfs \Gbf_{\Thetabf}$ and $\Bbf=\begin{bmatrix}\Hbff, \Hbfr\end{bmatrix}$, where $\Gbf_{\Thetabf}\in \R^{1\times n}$ represents a linearization of the non-linear spiking function $g_{\Thetabf}(\cdot)$.
However,  we propose to make a deviation from this mapping and interpret the spike of the neuron itself as an external input to the neuron state. This deviation avoids the linearization of the non-linear spiking function. Hence, we use $\Abf=\Hbfv$, and $\Bbf=\begin{bmatrix}\Hbfs, \Hbff, \Hbfr\end{bmatrix}$.
To visualize this correspondence, we rewrite the general spiking neuron in \eqref{eqn:general_LIF_intro} as 
\begin{subequations}
\label{eqn:general_LIF_rearange}
\begin{align}
\! \vbf[t+1] &=  \Hbfv \vbf[t] 
           +\begin{bmatrix}
           \Hbfs, \Hbff, \Hbfr 
            \end{bmatrix}
            \begin{bmatrix}
            \Sil{t} \\ \Wbf \Slbefore{t} \\ \Vbf \Sl{t} 
            \end{bmatrix} \\
 \Sil{t} &= g_{\Thetabf}(\vbf[t])
\end{align}
\end{subequations}
Hence, we propose  the following approximate correspondence between the general SNN neuron model and the linear SSM framework: 
\begin{align}
\label{eqn:state:evaluation:withSNNconnection}
\vbf[t+1] &=  \Abf \vbf[t] + \Bbf \fbf[t] 
\end{align}
where
$\vbf \in \R^{n \times 1}, 
\Abf = \Hbfv \in \R^{n \times n}, 
\Bbf = \begin{bmatrix} \Hbfs, \Hbff, \Hbfr  \end{bmatrix} \in \R^{n \times r}$, and
$\fbf[t] =  \begin{bmatrix}
            \Sil{t}, \Wbf \Slbefore{t}, \Vbf \Sl{t} 
            \end{bmatrix}^T \in \R^{r \times 1}$
such that $\fbf$ combines all input spikes to the neuron, including the effect of its own output. 
Since only the $\Abf$ and $\Bbf$ matrices are needed for adapting the state variable dynamics, which is our main interest here, we leave  \eqref{eqn:non-lssm-constraint_and_snn:correspondence:C} and  \eqref{eqn:non-lssm-constraint_and_snn:correspondence:D} as they are, since these functions are associated with the output spikes $\ybf[t]$ and are not required for adaptation of state dynamics.

We now consider adLIF neuron  as an example. For adLIF, using  \eqref{eqn:adLIF_H_matrices}, we obtain the following as $\Abf$ and  $\Bbf$ of \eqref{eqn:state:evaluation:withSNNconnection}: 
\begin{subequations}
 \label{eqn:adLIF_AandB}
\begin{align}
    \Abf &= \Hbfv = \begin{bmatrix}\alpha & -(1-\alpha)\\ a &\beta\end{bmatrix}, \\
    \Bbf &= \begin{bmatrix} \Hbfs, \Hbff, \Hbfr  \end{bmatrix} =\begin{bmatrix} -\alpha\theta & 1-\alpha & 1-\alpha \\ b&0&0\end{bmatrix}.
\end{align}
\end{subequations}

\subsection{Proposed Methods for Temporal Resolution Domain Adaptation}\label{sec:proposedMethods:main}
In this section, we define our proposed temporal resolution domain adaptation methods for the generalized SNN neuron model in \eqref{eqn:general_LIF_intro}.
As explained in Section \ref{sec:prob_statement_temp_res}, the proposed methods suggest a mapping ${\cal{M}}(.)$ from the source domain with temporal resolution $\DeltaS$ to the target domain with temporal resolution $\DeltaT$, where $\DeltaRatio = \frac{\DeltaT}{\DeltaS}$.

Note that all the methods presented below have been derived in detail in the form of propositions in Section \ref{sec:propositions}. A step-by-step pseudo-code outlining the usage of the methods is provided in Algorithm~\ref{alg:source-to-target-adapt}.

\noindent \textbf{Integral approximation adaptation method} is derived from the Integral approximation method for obtaining discrete-time SSM from a continuous-time system model. Following Proposition \ref{Integral_Method_Prop} in Section \ref{sec:propositions},  the proposed mapping is 
\begin{subequations}
\label{eqn:H_mapping:integral}
\begin{align} 
\! \HbfvT 
 &= \left( \HbfvS \right)^ \DeltaRatio  \\
 \HbfiT
 &= (\HbfvT - \Ibf) (\HbfvS - \Ibf)^{-1} \HbfiS 
\end{align}
\end{subequations}
for $k=s,f,r$.

\noindent \textbf{Euler adaptation method} is derived from the Euler approximation method for obtaining discrete-time SSM from a continuous-time system model. Following Proposition \ref{Euler_Method_Prop} in Section \ref{sec:propositions},  the proposed mapping is 
\begin{subequations}
\label{eqn:H_mapping:euler}
\begin{align} 
\! \HbfvT 
 &= \Ibf + \DeltaRatio(\HbfvS - \Ibf) \\
 \HbfiT
 &= \DeltaRatio \HbfiS 
\end{align}
\end{subequations}
for $k=s,f,r$.

\noindent \textbf{Expectation adaptation method} is based on the assumption  the expected values of the state variable vector between different time resolutions are the same.
This adaptation method assumes that the ratio of the time steps ($\Delta$) of the lower and higher resolutions is an integer. Specifically, if $\DeltaRatio>1, \DeltaRatio \in \Z_{+}$, while if
$\DeltaRatio<1, 1/\DeltaRatio \in \Z_{+}$.
Following Proposition \ref{Method_Expectation_Prop} in Section \ref{sec:propositions} the proposed mapping is 
\begin{subequations}
\label{eqn:H_mapping:exp}
\begin{align} 
\noindent \! 
\HbfvT &= \left( \HbfvS \right) ^ \DeltaRatio\\
\! \HbfiT &=\begin{cases} 
          (\sum^{\DeltaRatio}_{j=1}\left( \HbfvS \right)^{\DeltaRatio-j}) \HbfiS & \DeltaRatio \geq 1 \\
          (\sum^{1/\DeltaRatio}_{j=1}\left( \HbfvT \right)^{1/\DeltaRatio-j})^{-1} \HbfiS & \DeltaRatio < 1
        \end{cases}
 \end{align} 
 \end{subequations}
for $k=s,f,r$.

\begin{remark}
In the Integral and Expectation adaptation methods,
$\left( \HbfvS \right)^ \DeltaRatio$ represents a matrix raised to a possibly fractional power which can be computed using the Schur–Padé algorithm \cite{MatrixFractionalPower}. 
In general, fractional powers of a real-valued matrix may be complex-valued. 
By restricting the parameters of the adLIF neuron \eqref{eqn:adLIFneuron_intro} to the ranges
$\alpha,\beta, a \in (0, 1)$, the possibly fractional power $\DeltaRatio=1/\NRatio$ of the real matrix $\HbfvS$ used in our experiments remains real-valued \cite{MatrixFractionalPower}.
\end{remark}

\begin{algorithm*}[t]
\caption{Pseudo-code for the work-flow with the proposed adaptation methods}
\label{alg:source-to-target-adapt}
\DontPrintSemicolon
\SetKwInOut{Input}{Input}\SetKwInOut{Output}{Output}

\BlankLine
\BlankLine
\textbf{(1) Train on source resolution}\;
\Input{Training dataset $\mathcal{D}_S$ with source resolution $\DeltaS$.}
Initialize model randomly $SNN_S=(\HbfvS^{(0)}, \HbffS^{(0)}, \HbfsS^{(0)}, \HbfrS^{(0)}, \Wbf_S^{(0)}, \Vbf_S^{(0)})$. \;
\For{epoch $=1,\dots,E$}{
  Update the $SNN_S$ model parameters using $\mathcal{D}_S$.\;
} 
\Return $SNN_S=(\HbfvS, \HbffS, \HbfsS, \HbfrS, \Wbf_S, \Vbf_S)$. 

\BlankLine
\BlankLine
\textbf{(2) Perform adaptation}\;
\Input{Trained $SNN_S$, $\rho$, i.e., the ratio $\frac{\DeltaT}{\DeltaS}$, choice of adaptation method, i.e., Integral, Euler or Expectation.} 
Adapt neuron model parameters of $SNN_S$ i.e. set 
$\HbfvT \HbffT, \HbfsT, \HbfrT
= {\cal{M}}(\DeltaRatio, \HbfvS, \HbffS, \HbfsS, \HbfrS)$ using \eqref{eqn:H_mapping:integral} for Integral 
\eqref{eqn:H_mapping:euler} for Euler, or 
\eqref{eqn:H_mapping:exp} for Expectation. \; 
Create the model $SNN_T=(\HbfvT, \HbffT, \HbfsT, \HbfrT, \Wbf_S, \Vbf_S)$. \;
\Return  $SNN_T$. 

\BlankLine
\BlankLine
\textbf{(3) Use on data with target resolution}\;
\Input{$SNN_T$, testing dataset $\mathcal{D}_T$ with target resolution $\DeltaT$.} 
\Output{Performance of $SNN_T$ on $\mathcal{D}_T$ i.e., $SNN_T(\mathcal{D}_T)$.}
\end{algorithm*}

\subsection{A Simple Benchmark Method: Time-constant Adaptation Method}\label{sec:Method_TimeConst}
We now  introduce a simple method that will serve as a benchmark for our proposed methods. 
This benchmark method is motivated by the explicit dependence of various neuron parameters on the sampling period in the popular LIF neuron and adaptive LIF neuron models, and has been utilized in the literature for the LIF neuron \cite{he2020comparing}.  
Assume that the elements of the matrices $\HbfiS$ for $k=v,s,f,r$ belong to one of three groups:
\begin{enumerate}[label=\roman*]
     \item in the form of $\hS=\exp^{-\DeltaS/\tau}$ where $\tau \in \R_{+}$; \label{item:gr1}
     \item linear transformation of $\hS$; \label{item:gr2}
     \item has no explicit dependency on $\DeltaS$. \label{item:gr3}
\end{enumerate}
The time-constant adaptation methods obtain $\HbfiT$ for $k=v,s,f,r$, by replacing elements of $\HbfiS$ belonging to the group \ref{item:gr1}, \ref{item:gr2}, \ref{item:gr3} by:
\begin{enumerate}[label=\roman*]
    \item $\hT= \left( \hS \right)^{\DeltaRatio}$ 
    \item the same linear transformation of  $\hT$
    \item rest of the terms are unchanged
\end{enumerate}
To illustrate the method, we now look at the popular adaptive LIF neuron from Section \ref{sec:popular_neurons}. 
In the adaptive LIF neuron, all of the three types of elements are present. Here,  
$\alphaT = \alphaS^{\DeltaRatio}$ and $\betaT = \betaS^{\DeltaRatio}$ due to their explicit exponential dependence on the step length $T$. 
The other parameters are either scaled accordingly due to the dependence with the already scaled parameters, for example $-(1-\alphaT)$, or kept the same (in the case of the parameters $a$ and $b$).

\begin{remark}
The proposed methods of Section~\ref{sec:proposedMethods:main} do not require the parameters of the neuron model to  explicitly depend on the time resolution and hence they are applicable to a wide range of neuron models. This is in contrast to the benchmark method of Section~\ref{sec:Method_TimeConst}, where the method is best suited to the neuron models where all neuron parameters have explicit dependence on the time resolution. 
\end{remark}

\subsection{Computational Complexity}
\label{sec:computational_complexity}
In this section, we discuss the computation complexity of the proposed methods based on the number of multiply–accumulate (MAC) operations needed.

\subsubsection{Computational complexity of proposed adaptation methods mapping}
\label{sec:cost_of_adaptation}

In this section, we calculate the scaling of MAC operations for the proposed adaptation methods in terms of the number of neurons, neuron state dimensions, and target-to-source temporal resolution ratio $\rho$.  

For the basic matrix operations, we use the following conventions: 
    i) multiplying  a $(n\times n)$ matrix and a $(n\times 1)$ vector requires $\propto n^2 $ MACs;
    ii)  inverting a $(n\times n)$ matrix requires $\propto n^3$ MACs;
    iii) finding $\rho>1$ integer power of a $(n\times n)$ matrix, using repeated matrix power multiplication,  requires $\propto (\rho-1) n^3 \propto \rho n^3$ MACs; 
    iv) finding $\rho<1$ fractional power of $(n\times n)$ matrix requires $\propto (1/\rho)  n^3$ MACs \cite{MatrixFractionalPower}, \cite[Sec.4.2]{Higham_AlMohy_2010}. 

Using the general spiking neuron model in  \eqref{eqn:general_LIF_intro} and $\Hbfv \in \R^{n \times n}$ and $\Hbfs, \Hbff, \Hbfr  \in \R^{n \times 1}$, we obtain the following scaling  of MAC operations per neuron for $\rho \geq 1$:
\begin{itemize}
    \item Integral method in \eqref{eqn:H_mapping:integral}: $\rho\, n^3$ for $\HbfvT$ and $n^3$ for $\HbfiT$ for $k=s,f,r$.
    \item Euler method in \eqref{eqn:H_mapping:euler}: $n^2$ for $\HbfvT$ and $n$ for $\HbfiT$ for $k=s,f,r$.
    \item Expectation method in \eqref{eqn:H_mapping:exp}: $\rho\, n^3$ for $\HbfvT$ and $\rho\, n^3$ for $k=s,f,r$. 
\end{itemize}
For $\rho<1$,  $\rho$ is replaced with $1/\rho$. We continue with the convention $\rho \geq 1$ in the below. 

The above calculations are per neuron. For the whole network, these numbers are multiplied by  the number of neurons, i.e.,  $\sum^{L}_{l=1} h_l$,  where $h_l$ is the number of neurons in the  hidden layer $l$ and $L$ is the total number of hidden layers. Hence, assuming all hidden layers have $h_l$ neurons, number of MAC operations scale with 
$\rho\,  n^3 h_l$  for Integral method, $n^2 h_l$ for Euler method and $ \rho  n^3  h_l$ for Expectation method. 

We note that only the ratio of the target and source sequence lengths, i.e. $\rho$ (in \eqref{eqn:rho_definition}),  affect the computational cost. In other words,  the length of the source or target sequence does not directly impact the computational complexity of our adaptation methods.

\subsubsection{Computational complexity of inference}
\label{sec:cost_of_inference}

The adaptation of the model is performed once. After the adaptation, inference proceeds normally. Our methods operate by replacing existing neuron parameters with the adapted values, without introducing any additional parameters. Thus, the total parameter count of the network remains unchanged, and the number of operations per time step during inference is identical to that of the original, non-adapted model. 

We now provide an overview of the computational complexity cost during inference, i.e., forward pass.  
The dominant MAC cost during inference comes from the weight matrix multiplications which scales with  $\sum^{L}_{l=1} h_{l-1}h_{l}$ for a feedforward network and $\sum^{L}_{l=1} h_{l-1}h_{l} +  h_{l}^2$ for a  recurrent network. This is due to the fact that in a typical SNN, the state dimension of neurons $n$, is much smaller than the number of neurons in the layer $h_l$,  i.e. $h_l \gg n$, hence the typical computational cost due to the update of dynamics is insignificant. Hence,  the MAC per time step during inference grows with $h_l^2$.
For a sequence with $T$ number of time steps, the inference cost then becomes $T  h_l^2$.
Hence, the computational cost of inference, under both the adapted and non-adapted models, scales with  $T  h_l^2$.

\subsubsection{Comparison between cost of adaptation and re-training}
\label{sec:cost_comparison_adapt_and_train}

In this article, we focus on a zero-shot setup where the target data is unavailable and re-training cannot be performed. For comparison purposes, here we discuss the computational cost of re-training, and compare it with the cost of our adaptation methods.

The computational complexity of re-training depends on the computational cost of backward pass. 
A common approximation is that the computational backward pass requires about twice the cost of the forward pass \cite{wiedemann2020dithered} \cite[Fig. 3]{narayanan2021efficient}. Thus, for sequence of length $T$, since inference complexity scales with $\propto T  h_l^2$  (see Section \ref{sec:cost_of_inference}), training complexity scales with $3T  h_l^2$ including both inference and backward pass. Moreover, since training is typically performed over multiple data samples,  the scaling of cost of re-training becomes $3 B T h_l^2$ per training step where $B$ is the number of samples. 

We now compare the cost of adaptation, which scales with $\rho  n^3 h_l$ at worst,  with the cost of re-training, which  scales with $3BT h_l^2$ for a single epoch. We note that in SNNs typically $h_l \gg n$, and $B,T \gg 1$ so that it is typically the case that the scaling satisfies $3BT h_l^2 \gg \rho n^3 h_l$. This represents a substantial difference in scaling of the complexity for SNNs with large $h_l$, highlighting the efficiency fo scaling of our proposed methods of adaptation over re-training. Moreover, in contrast to re-training, the computational cost of our proposed adaptation methods are independent of the sequence length. As a result, our approach scales much better than full re-training when dealing with higher-resolution data or long sequences.

To illustrate the possible difference in the scaling of computational cost, we now provide typical values for $n, h_l, T$ and $B$. For both SHD and MSWC datasets, this article and the prior works \cite{bittar2022surrogate, yik2024neurobench} use RadLIF neurons with $n=2$ and hidden layers with $h_l=1024$ neurons. In our setting,  sequence length in its highest resolutions for SHD dataset is $T=100$, and for MSWC, it is $T=201$. Under these settings, for our zero-shot adaptation, we have $\rho n^3 h_l=\rho \times 2^3\times 1024$, whereas the scaling of re-training per epoch per sequence is $T h_l^2=100\times 1024^2$ for SHD dataset. Moreover, training a single epoch over the full training dataset would multiply the cost of training by $B=8332$ for SHD, and $B=5\times 10^4$ for MSWC dataset.

The developments in this section provide approximate proportional scalings of the computational costs. Their reflection in practice depends on variable ranges, hardware specifics and implementation details. In~Section~\ref{sec:results:time_efficient_training}, we provide empirical comparisons based on actual wall-clock times.
\section{Results}
\label{sec:numericalResults}

In Section \ref{sec:num:setting:maintext}, we define our experimental setup. 
Section \ref{sec:illustrative_example} illustrates the performance of the proposed methods on a single neuron using a synthetic dataset. Section \ref{sec:results:baseline} presents the baseline performances in various temporal resolutions with event-based datasets. Sections \ref{sec:results:L2H} and
\ref{sec:results:H2L} present the performance of the proposed temporal domain adaptation methods on these datasets. This motivates us to illustrate in Section \ref{sec:results:time_efficient_training} that the proposed methods can be used to support time-efficient training.

The research questions targeted by these experiments are the following:
\begin{itemize}
    \item Can the proposed methods for zero-shot adaptation without re-training achieve satisfactory accuracy levels on varying temporal resolution levels?
    \item How do the proposed methods perform relative to the benchmark methods?
    \item Can we train efficiently at low temporal resolution and infer at high resolution with minimal performance loss using the proposed methods?
\end{itemize}

\subsection{Preliminaries}\label{sec:num:setting:maintext}

\subsubsection{Datasets}
We use the following event-based datasets:

The Spiking Heidelberg Digits (SHD) dataset \cite{shddataset} is an audio dataset consisting of non-signed spikes generated using the mathematical artificial cochlea model called Lauscher. The dataset consists of $20$ classes of spoken digits from $0$ to $9$ in both German and English language with total of  $8156$ train samples, $2264$ test samples. Each sample consists of $700$ input channels each with $100$ time steps, where each time step represents summed spikes over the time window of length $1$ms. The benchmark for SHD dataset is $95.1\%$ test classification accuracy \cite{hammouamri2023learningdelaysspikingneural} where SNN with learnable delays are used. The closest setup to our work achieves $94.6\%$ accuracy \cite{bittar2022surrogate}, using the same SNN architecture and recurrent adaptive LIF neuron. However, their best model was selected based on the test dataset rather than a separate validation dataset, making the benchmark overly optimistic.

The Multilingual Spoken Word Corpus (MSWC) dataset \cite{mswcdataset} is a large and growing audio dataset. In this paper we use a subset of samples and encoding as proposed in the NeuroBench initiative \cite{yik2024neurobench}. The subset consists of $100$ classes, $20$ classes per $5$ languages. We use $600$ train, and $100$ test samples per class. Each audio sample is converted to signed spikes using the Speech2Spikes (S2S) \cite{speechtospikes} preprocessing algorithm resulting in $20$ input channels and $201$ time steps. The current baseline for this subset of MSWC is $93.48\%$ classification accuracy \cite{yik2024neurobench}.

The NMNIST dataset \cite{orchard2015converting} is a neuromorphic version of the MNIST vision dataset. It consists of $10$ classes representing the digits $0$ to $9$ in an event-based manner. Samples are obtained by recording the static MNIST images shown on a monitor using a mobile event based image sensor \cite{orchard2015converting}. The dataset consists of $60 000$ train and $10 000$ test samples. We use the Tonic library \cite{tonic} for converting each sample into $(34 \times 34 \times 2, N)$ spike pattern, where the first dimension represents spatial information such as channel and height and widths of the frame, while the second dimension  is the number of time steps. We take $N\approx300$ where the exact number differs due to the Tonic denoise filter. The state of the art for NMNIST dataset is $99.6\%$ accuracy using  convolutional SNNs \cite{samadzadeh2021convolutionalspikingneuralnetworks}.

\subsubsection{Overview of experimental procedure}
We define the data in the above datasets as having the fine temporal resolution, and refer them as the scenario of bin size $1$, i.e. $b=1$. We create data for  coarser resolutions ($b=2, 3, 4, 10$) using the sum-binning procedure described in Section \ref{sec:num:setting:sum_binning}. 
We have two sets of experiments: Coarse-to-Fine and Fine-to-Coarse deployment. 
In the Coarse-to-Fine experiments, the models are trained on a simulated coarse-resolution version of the train dataset, and evaluated on the test dataset with the unaltered fine resolution. 
In the Fine-to-Coarse experiments, the models are trained on the unaltered fine-resolution dataset, and evaluated on the simulated coarser-resolution version of the test dataset.

All the values reported in this paper are average performance of $10$ different initializations of the models.

\subsubsection{Model Architecture and Training} 
For all experiments we use a SNN with adaptive LIF (adLIF) neuron dynamics \cite{bittar2022surrogate} with $2$ hidden layer of $1024$ neurons. In the results, this setup is referred to as adLIF for the model without recurrent connections and RadLIF for the model with recurrent connections. Each model is trained for $50$ iterations, and we report the test data accuracy at the final ($50$-th) iteration. Models are trained using BPTT with a box surrogate gradient \cite{neftci2019surrogate} as implemented by Bittar and Garner \cite{bittar2022surrogate}. Models are trained on Nvidia A100 GPU with 64GB system memory, 40GB VRAM and 16 CPU cores. See Section \ref{sec:HPO} for further details on hyperparameters.

\subsubsection{Data with coarse temporal resolution}
\label{sec:num:setting:sum_binning}
Given a sequence, we obtain a coarser time resolution version of it using sum-binning with a non-overlapping moving window along its time dimension. For example, given a sequence of length $4$, say $(x, y, z, w)$ where $x, y, z, w \in \Z$, by applying non-binary sum binning with bin size $b=2$,  
we obtain $(x + y, z + w)$.
Note that increasing the bin size smooths over temporal features, leading to a loss of temporal detail and reduced information in the data.

\subsubsection{Batchnorm-scaling} 
One dimensional Batch Normalization layers \cite{ioffe2015batchnormalizationacceleratingdeep}
are used before each spiking layer in the SNNs. 
These layers are adapted due the change of temporal resolution between target and source data,  see Section \ref{sec:num:setting:batchnorm_scaling} for details.

\subsubsection{Benchmark Methods for Temporal Resolution Domain Adaptation}
\label{sec:num:setting:benchmarks}

In both Fine-to-Coarse and Coarse-to-Fine experiments, two benchmarks are used: (i) No-adaptation of the model;  (ii) Time-constants adaptation benchmark of Section \ref{sec:Method_TimeConst}.

\subsection{Illustrative Example: Neuron Dynamics under Proposed Methods}
\label{sec:illustrative_example}
We now investigate the performance of our proposed temporal domain adaptation methods at a single neuron level. 
We use the adLIF neuron as in \eqref{eqn:adLIFneuron_intro}. Specifically, as we have a single neuron without recurrent connections to other neurons, 
\eqref{eqn:adLIF_mem} becomes 
\begin{equation}
 \Umem{t+1} = \alpha (\Umem{t} - \theta \Sil{t} ) + (1-\alpha) i[t] - (1-\alpha) \Uad{t},
\end{equation}
while 
\eqref{eqn:adLIF_adap} and \eqref{eqn:adLIF_spk} remain the same.
As input to the neuron, we use an excitation signal commonly used in signal processing and system identification for its ability to excite a wide range of system dynamics,  i.e., we use the input
\begin{equation}
    i[t] = \sum^K_{k=1}A_k \sin[\omega_k t +\pi/\phi_k]
    \label{eq:sin_input}
\end{equation}
which represents a linear combination of $K$ sin waves, each with a different amplitude $A_k$, frequency $\omega_k$, and phase shift $\pi/\phi_k$.
The maximum number of time steps is set to $T=100$ i.e. $t=0, 1,..., 100$.
We then also create $i_{2}[t]$,  a sum binned version of $i[t]$ with $b=2$ and $T=50$.  
The general objective is to explore to which extent the dynamic behavior of the neuron  can be preserved  despite the changes in temporal resolution of the input data. \\
In the Fine-to-Coarse scenario, i.e. {source bin $\bS=1$ to target bin $\bT=2$}, first the high resolution $i[t]$ is given as input to the adLIF neuron and $\UmemNoArg[t]$ is recorded. We refer to this as reference dynamics. We then apply different model adaptations for the parameters of the adLIF neuron and record  $\UmemNoArg[t]$
when sending $i_{2}[t]$ as input to the neuron. 
In the  Coarse-to-Fine scenario, i.e. {$\bS=2$ to $\bT=1$}, $i_{2}[t]$ constitutes the reference dynamics, and the performance of different model adaptations under the high resolution input $i[t]$ is evaluated. \\  
To quantitatively inspect the mismatch between the reference and the adapted dynamics,  we use the quality functions $\Qone$ and $\Qtwo$, which are associated with the normalized relative square error, and  the correlation coefficient, respectively.  For both $\Qone$ and $\Qtwo$, having a value close to $1$ indicates a better reconstruction compared to a value close to $0$.  See Section \ref{sec:num:setting:quality_fnc} for exact definitions.

\begin{figure}
\centering
     \begin{subfigure}[b]{1\linewidth}
     \centering
     \caption{Fine-to-Coarse}
     \includegraphics[width=\textwidth]{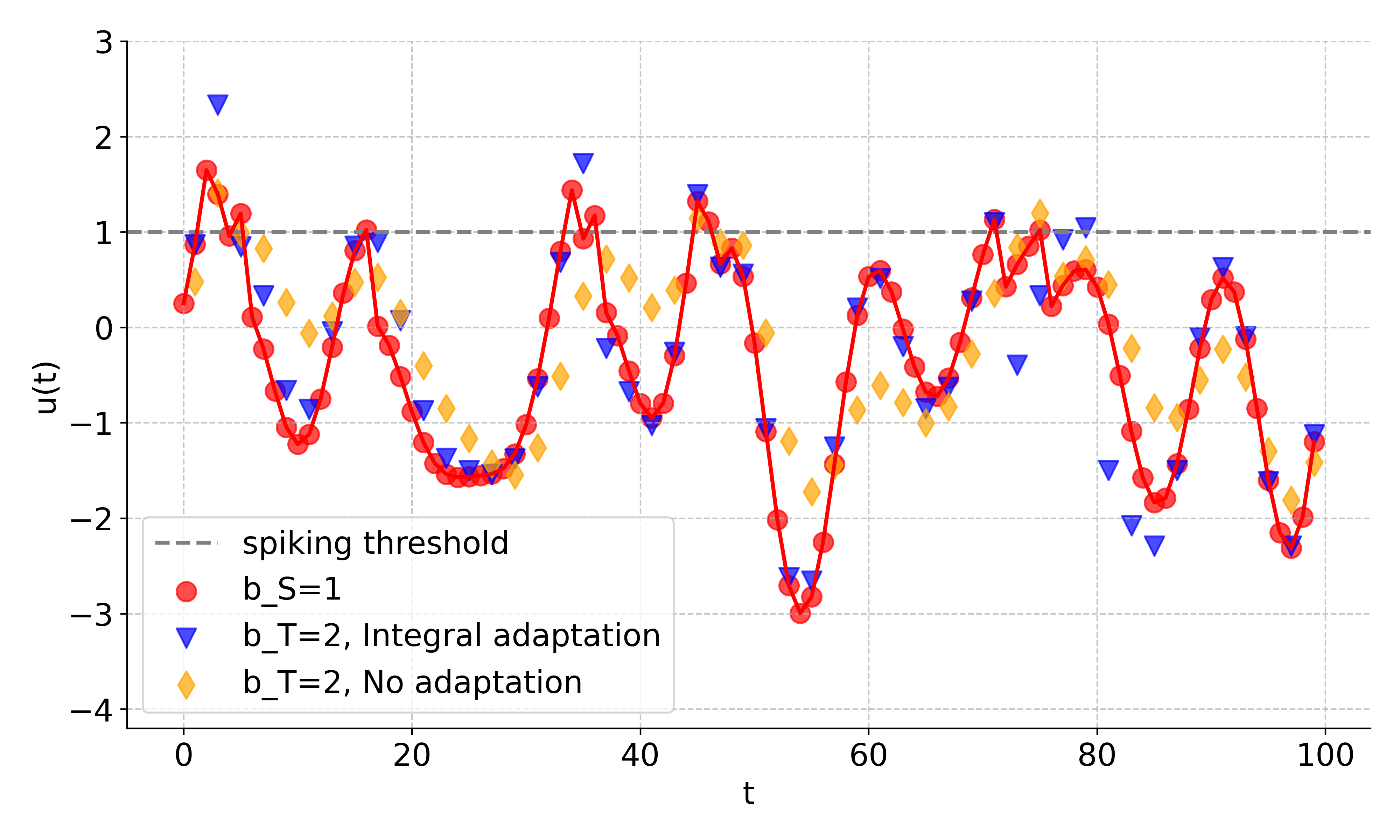}
     \label{fig:H2L:Single_Neuron_Scaling}
     \end{subfigure}
\hfill
     \begin{subfigure}[b]{1\linewidth}
     \centering
     \caption{Coarse-to-Fine}
     \includegraphics[width=\textwidth]{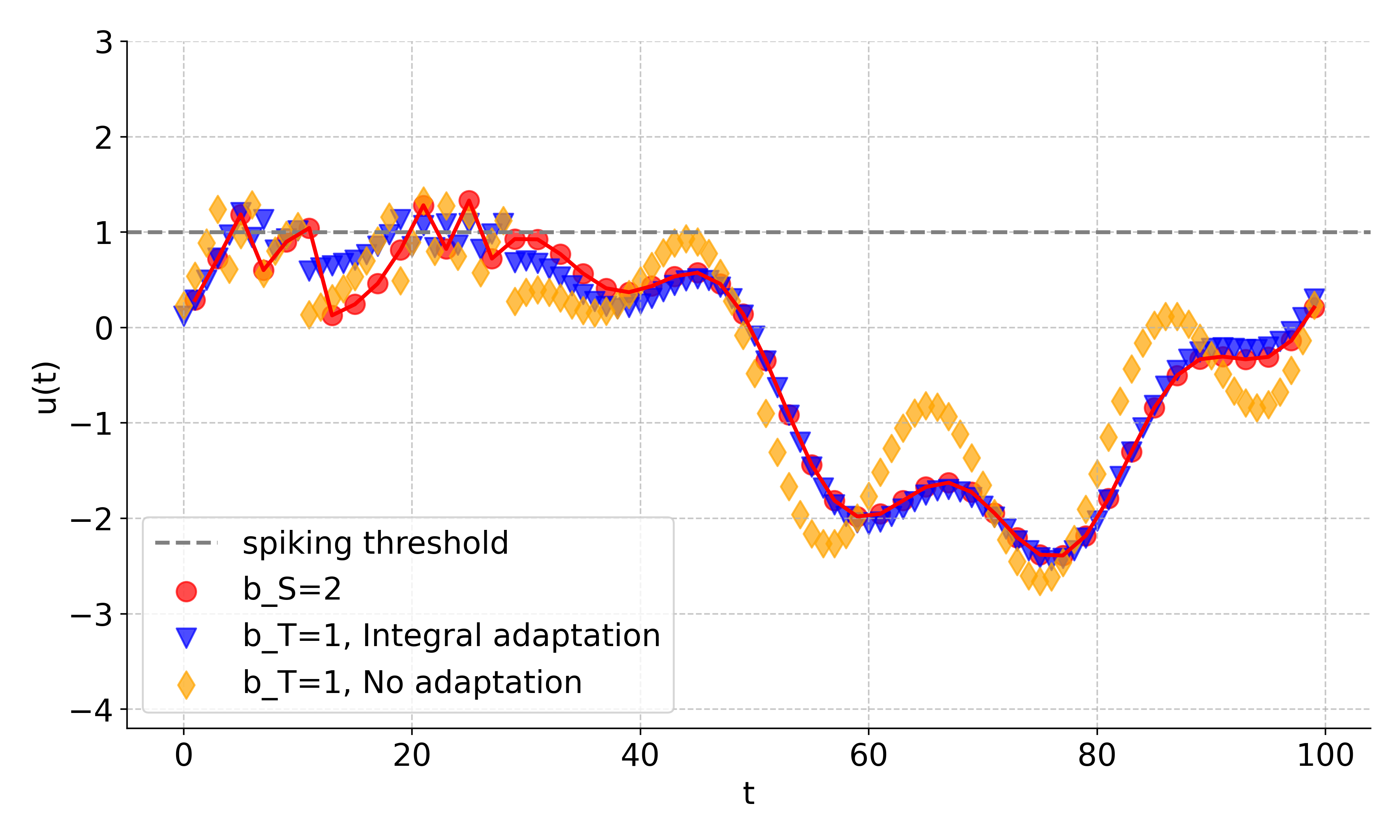}
     \label{fig:L2H:Single_Neuron_Scaling}
     \end{subfigure}
\caption{Membrane potential dynamics of a single adLIF neuron over time for different adaptation methods and time resolutions where $\bS$ (red circular) is the reference dynamics.}
\label{fig:Single_Neuron_Scaling}
\end{figure}

The input $i[t]$ is defined as in \eqref{eq:sin_input} with $K=3$, $A_k\in[0.1, 0.2]$, $\omega_k \in[1, 10]$, and $\phi_k \in [1, 20]$, and  adLIF neurons with $\alpha\in[0.6, 0.98]$, $\beta\in[0.6, 0.98]$, $a\in[0.2, 0.5]$, and $b\in[0.2, 0.5]$.  We simulate $1000$ random pairs of input sequences and adLIF neurons from these intervals and calculate the average $\Qone$ and $\Qtwo$ over all instances. The resulting values are presented in  Table~ \ref{tab:matching_stats}.  
Two illustrative cases with their corresponding membrane potential $\UmemNoArg[t]$ are provided in Figure \ref{fig:Single_Neuron_Scaling}.
On average, the results in Table~\ref{tab:matching_stats} indicate a better overall match under the three proposed adaptation methods (Integral, Expectation and Euler) compared to both the Time-constant adaptation baseline and the No adaptation case. 
Time-constant adaptation method performs relatively poor compared to the proposed methods according to $\Qone$, and relatively close to the these methods according to $\Qtwo$.   
Figure \ref{fig:Single_Neuron_Scaling} illustrate that although the integral adaptation method may match the reference dynamics better than no adaptation, noticeable deviations compared to the original signal may also occur.

\begin{table}
\centering
\caption{Qualitative match of the voltage dynamics for various adaptation methods. Each value is average of $1000$ different combination of adLIF neuron and input sequence.$\bS$ and $\bT$ stand for source and target bin size.}
\label{tab:matching_stats}
\newcolumntype{l}{>{\centering\arraybackslash}p{2.5cm}}
\newcolumntype{m}{>{\centering\arraybackslash}p{2cm}}
\begin{subtable}{1\linewidth}
\centering
\vspace{0.15cm}
 \small
\caption{$\bS=1$ to $\bT=2$}
\begin{tabular}{lmm}
 \hline
 Model Adapt. & $\Qone$ & $\Qtwo$  \\
 \hline
 None &0.66 $\pm$ 0.16&0.84 $\pm$ 0.08\\ 
 Integral  &\textbf{0.92 $\pm$ 0.10}&\textbf{0.97 $\pm$ 0.04}\\ 
 Expectation & 0.91 $\pm$ 0.11&0.96 $\pm$ 0.04\\ 
 Euler &0.87 $\pm$ 0.11&0.95 $\pm$ 0.03\\  
 Time-const. &0.77 $\pm$ 0.11&0.95 $\pm$ 0.03\\ 
\hline
\end{tabular}
\end{subtable}
\begin{subtable}{1\linewidth}
\centering
\vspace{0.15cm}
\small
\caption{$\bS=2$ to $\bT=1$}
\begin{tabular}{lmm}
 \hline
 Model Adapt. & $\Qone$ & $\Qtwo$   \\
 \hline
 None &0.57 $\pm$ 0.25&0.84 $\pm$ 0.08\\ 
 Integral  &\textbf{0.97 $\pm$ 0.05}&\textbf{0.99 $\pm$ 0.01}\\ 
 Expectation &\textbf{0.97 $\pm$ 0.03}&\textbf{0.99 $\pm$ 0.01}\\ 
 Euler &0.96 $\pm$ 0.06&\textbf{0.99 $\pm$ 0.02}\\  
 Time-const. &0.84 $\pm$ 0.07&0.93 $\pm$ 0.03\\ 
\hline
\end{tabular}
\end{subtable}
\end{table}

\subsection{Baseline Performance for Different Temporal Resolutions}
\label{sec:results:baseline}

In Table \ref{tab:learn_from_scratch_bins_result} we present the performance of SNN models trained and tested on the same time resolution, i.e., the source and target bin sizes are equal, $\bS=\bT$. We refer to these results as baseline performance since they show the accuracy levels and computational time that would be achieved if the models were trained with source data whose temporal resolution is the same as in the target data.

Note that lowering the time resolution might result in drop in accuracy due to potentially removing finer details in the time sequence by eliminating or deforming important features needed for the classification task. 
However, higher time resolutions require more backpropagation steps during training, which may negatively affect performance due to the limitations of surrogate gradients in BPTT for SNNs. This phenomenon can also be seen in\cite[Table VI, VII, VIII]{he2020comparing}.
Table \ref{tab:learn_from_scratch_bins_result} illustrates both of these possibilities: for the SHD dataset (Table \ref{tab:sub:learn_from_scratch_SHD_Dataset}), accuracy decreases monotonically with lower resolution, while for the MSWC dataset (Table \ref{tab:sub:learn_from_scratch_MSWC_Dataset}), accuracy initially increases slightly before declining.

Compared to the SHD and MSWC datasets, NMNIST shows a smaller decline in accuracy as the bin size increases. This is consistent with the static nature of the NMNIST data, in contrast to the dynamic audio data, where temporal features play a key role in encoding information.

Although various anti-aliasing filters can be used to achieve lower temporal resolution, sum-binning offers advantages in terms of simplicity, real-time application, and preserving the event-based nature of the data. As shown in Table \ref{tab:learn_from_scratch_bins_result}, the accuracy remains largely unaffected for bin sizes of $2$ to $4$, indicating minimal impact from aliasing noise or the exclusion of high-frequency content. However, a significant performance drop is observed with a bin size of $10$, suggesting that
important temporal content is lost under $b=10$.


\begin{table}
\caption{Baseline results for training SNN models using various bin sizes, where the source and target bin sizes are equal $\bS=\bT$. Two different neuron models are used: adLIF and RadLIF. Each value is average over $10$ trials. The benchmark results from the literature presented in Section \ref{sec:num:setting:maintext} are obtained using the highest temporal resolution data i.e. $b=1$.}
\label{tab:learn_from_scratch_bins_result} 
\newcolumntype{l}{>{\centering\arraybackslash}p{0.65cm}}
\newcolumntype{m}{>{\centering\arraybackslash}p{2cm}}
\begin{subtable}[h]{1\linewidth}
\centering
\vspace{0.15cm}
\caption{Results for SHD dataset.}
\label{tab:sub:learn_from_scratch_SHD_Dataset}
\resizebox{\columnwidth}{!}{%
\begin{tabular}{llllmlm} 
 \hline
  &&& \multicolumn{2}{c}{adLIF} & \multicolumn{2}{c}{RadLIF} \\
   \cmidrule(lr){4-5} \cmidrule(lr){6-7}
 Bin  & Steps   & $\Delta$ & Train  & Accuracy & Train & Accuracy\\
 (b) & (N) & (ms)  &  (min) & & (min)&  \\
\end{tabular}
}
\end{subtable}
\begin{subtable}[h]{1\linewidth}
\centering
\resizebox{\columnwidth}{!}{%
\begin{tabular}{llllmlm}
 \hline
1 & 100 & 1 & 6.3 & 91.5 $\pm$ 0.6 \% & 7.7 & 91.8 $\pm$ 0.7 \% \\
2 & 50 & 2& 2.7 & 90.3 $\pm$ 0.6 \% & 3.6 & 91.0 $\pm$ 1.0 \% \\
3 & 33 & 3& 1.8 & 89.8 $\pm$ 0.8 \% & 2.7 & 90.7 $\pm$ 0.7 \% \\
4 & 25 & 4& 1.6 & 89.1 $\pm$ 0.8 \% & 2.3 & 88.9 $\pm$ 0.9 \% \\
10 & 10 & 10&  1.1 & 81.8 $\pm$ 1.1 \% & 1.7 & 79.2 $\pm$ 1.5 \% \\
100 & 1 & 100& 0.7 & 50.2 $\pm$ 1.8 \% & 1.3 & 6.3 $\pm$ 2.8 \% \\
\hline
\end{tabular}
}
\end{subtable}
\begin{subtable}[h]{1\linewidth}
\centering
\vspace{0.15cm}
\caption{Results for MSWC dataset.}
\label{tab:sub:learn_from_scratch_MSWC_Dataset}
\resizebox{\columnwidth}{!}{%
\begin{tabular}{llllmlm}
 \hline
1 & 201 & 5 & 106.2 & 92.4 $\pm$ 0.3 \% & 117.0 & 95.7 $\pm$ 0.1 \% \\
2 & 100 & 10& 36.9 & 93.5 $\pm$ 0.2 \% & 41.7 & 95.9 $\pm$ 0.1 \% \\
3 & 67 & 15& 21.8 & 94.1 $\pm$ 0.2 \% & 24.7 & 95.7 $\pm$ 0.2 \% \\
4 & 50 & 20& 15.1 & 93.8 $\pm$ 0.2 \% & 17.5 & 95.6 $\pm$ 0.2 \% \\
10 & 20 & 50& 6.3 & 90.7 $\pm$ 0.3 \% & 7.3 & 92.8 $\pm$ 0.4 \% \\
201 & 1 & 1000&  2.3 & 1.0 $\pm$ 0.1 \% & 2.6 & 1.0 $\pm$ 0.1 \% \\
\hline
\end{tabular}
}
\end{subtable}
\begin{subtable}[h]{1\linewidth}
\centering
\vspace{0.15cm}
\caption{Results for NMNIST dataset.}
\label{tab:sub:learn_from_scratch_NMNIST_Dataset}
\resizebox{\columnwidth}{!}{%
\begin{tabular}{llllmlm}
 \hline
1 & 300 & 1 & 433.1 & 98.6 $\pm$ 0.1 \% & 487.9 & 98.5 $\pm$ 0.1 \% \\
2 & 150 & 2& 223.3 & 98.7 $\pm$ 0.1 \% & 245.4 & 98.5 $\pm$ 0.1 \% \\
3 & 100 & 3& 150.9 & 98.7 $\pm$ 0.1 \% & 178.1 & 98.5 $\pm$ 0.1 \% \\
4 & 75 & 4& 112.6 & 98.6 $\pm$ 0.0 \% & 143.4 & 98.5 $\pm$ 0.1 \% \\
10 & 30 & 10& 104.4 & 98.7  $\pm$ 0.1 \% & 87.4 & 98.3 $\pm$ 0.1 \% \\
300 & 1 & 300& 84.4 & 92.5 $\pm$ 0.5 \% & 84.9 & 11.3 $\pm$ 0.0 \% \\
\hline
\end{tabular}
}
\end{subtable}
\end{table}

\begin{table*}
\centering
\caption{Performance under various domain adaptation methods on target data with fine temporal resolution ($\bT=1$), where the models were trained on source data with coarser temporal resolution ($\bS=2, 3, 4, 10$). Each value in the table is average of $10$ trials.}
\label{tab:L2H}
\newcolumntype{l}{>{\centering\arraybackslash}p{2.4cm}}
\newcolumntype{m}{>{\centering\arraybackslash}p{2.2cm}}
\begin{subtable}[h]{1\linewidth}
\centering
\vspace{0.15cm}
\caption{Results using SHD Dataset. Baseline accuracy ($\bS=\bT=1$) from Table  \ref{tab:sub:learn_from_scratch_SHD_Dataset} is $91.4 \%$ for adLIF and $91.6 \%$ for RadLIF.}
\label{tab:sub:L2H_SHD_Dataset}
\resizebox{\textwidth}{!}{%
\begin{tabular}{mllllllll}
 \hline
     & \multicolumn{2}{c}{$\bS=2$ to $\bT=1$} 
     & \multicolumn{2}{c}{$\bS=3$ to $\bT=1$} 
     & \multicolumn{2}{c}{$\bS=4$ to $\bT=1$} 
     & \multicolumn{2}{c}{$\bS=10$ to $\bT=1$}  \\
 \cmidrule(lr){2-3} \cmidrule(lr){4-5} \cmidrule(lr){6-7} \cmidrule(lr){8-9}
 Model Adapt. &adLIF&RadLIF &adLIF&RadLIF &adLIF&RadLIF &adLIF&RadLIF \\
\end{tabular}
}
\end{subtable}
\begin{subtable}[h]{1\textwidth}
\centering
\resizebox{\textwidth}{!}{%
\begin{tabular}{mllllllll}
 \hline
 None & 58.1 $\pm$ 1.9 \% & 31.8 $\pm$ 4.6 \% & 30.3 $\pm$ 2.1 \% & 5.3 $\pm$ 0.4 \% & 20.5 $\pm$ 1.7 \% & 6.1 $\pm$ 1.5 \% & 9.6 $\pm$ 1.6 \% & 6.7 $\pm$ 2.1 \% \\
Expectation & \textbf{89.5 $\pm$ 0.6 \%} & \textbf{87.7 $\pm$ 2.3 \%} & \textbf{87.4 $\pm$ 0.4 \%} & \textbf{81.6  $\pm$ 1.9 \%} & \textbf{84.3 $\pm$ 0.6\%} & \textbf{72.8 $\pm$ 2.6 \%} & \textbf{70.4 $\pm$ 1.6 \%} & \textbf{45.4 $\pm$ 3.8 \%} \\
Integral & \textbf{89.5 $\pm$ 0.6 \%} & \textbf{87.7 $\pm$ 2.3 \%} & \textbf{87.4 $\pm$ 0.4 \%} & \textbf{81.6 $\pm$ 1.9 \%} & \textbf{84.3 $\pm$ 0.6 \%} & \textbf{72.8 $\pm$ 2.6 \%} & \textbf{70.4 $\pm$ 1.7 \%} & \textbf{45.4 $\pm$ 3.9 \%} \\
Euler & 84.6 $\pm$ 0.6 \% & 76.8 $\pm$ 7.0 \% & 77.8 $\pm$ 1.2 \% & 57.5 $\pm$ 5.7 \% & 71.3 $\pm$ 2.1 \% & 50.5 $\pm$ 7.4 \% & 49.7 $\pm$ 1.2 \% & 45.3 $\pm$ 4.4 \% \\
Time-const. &53.0 $\pm$ 3.2 \% & 7.5 $\pm$ 2.9 \% & 19.9 $\pm$ 3.9 \% & 5.6 $\pm$ 1.5 \% & 14.3 $\pm$ 4.1 \% & 5.0 $\pm$ 0.2 \% & 5.0 $\pm$ 0.0 \% & 5.0 $\pm$ 0.1 \% \\
\hline
\end{tabular}
}
\end{subtable}
\begin{subtable}[h]{1\textwidth}
\centering
\vspace{0.15cm}
\caption{Results using MSWC Dataset. Baseline accuracy ($\bS=\bT=1$) from Table  \ref{tab:sub:learn_from_scratch_MSWC_Dataset} is $92.4\%$ for adLIF and $95.7 \%$ for RadLIF.}
\label{tab:sub:L2H_MSWC_Dataset}
\resizebox{\textwidth}{!}{%
\begin{tabular}{mllllllll}
 \hline
None& 40.5 $\pm$ 1.9 \% & 22.2 $\pm$ 2.2 \% & 10.2 $\pm$ 0.5 \% & 3.4 $\pm$ 0.6 \% & 4.2 $\pm$ 0.2 \% & 1.7 $\pm$ 0.3 \% & 1.6 $\pm$ 0.2 \% & 1.3 $\pm$ 0.2 \% \\
Expectation& \textbf{93.6 $\pm$ 0.3 \%} & 94.2 $\pm$ 0.3 \% & \textbf{94.0 $\pm$ 0.2 \%} & \textbf{92.9 $\pm$ 0.4 \%} & \textbf{93.8 $\pm$ 0.1 \%} & \textbf{92.1 $\pm$ 0.9 \%} & \textbf{86.8 $\pm$ 1.0 \%} & \textbf{61.8 $\pm$ 9.1 \%} \\
Integral& \textbf{93.6 $\pm$ 0.3 \%} & {94.2 $\pm$ 0.3 \%}& \textbf{94.0 $\pm$ 0.2 \%} & \textbf{92.9 $\pm$ 0.4 \%} & \textbf{93.8 $\pm$ 0.1 \%} & \textbf{92.1 $\pm$ 0.9 \%} & \textbf{86.8 $\pm$ 1.0 \%} & \textbf{61.8 $\pm$ 9.2 \%} \\
Euler& 92.8 $\pm$ 0.2 \% &\textbf{ 94.9$\pm$ 0.2\%} & 92.6 $\pm$ 0.3 \% & 92.3 $\pm$ 0.6 \% & 91.7 $\pm$ 0.2 \% & 87.7 $\pm$ 1.0 \% & 74.4 $\pm$ 1.1 \% & 23.9 $\pm$ 3.8 \% \\
Time-const.&  38.8 $\pm$ 2.8 \% & 1.1 $\pm$ 0.2 \% & 7.1 $\pm$ 1.5 \% & 1.0 $\pm$ 0.1 \% & 1.9 $\pm$ 0.8 \% & 1.1 $\pm$ 0.2 \% & 1.2 $\pm$ 0.2 \% & 1.0 $\pm$ 0.1 \% \\
\hline
\end{tabular}
}
\end{subtable}
\begin{subtable}[h]{1\textwidth}
\centering
\vspace{0.15cm}
\caption{Results using NMNIST Dataset. Baseline accuracy ($\bS=\bT=1$) from Table  \ref{tab:sub:learn_from_scratch_NMNIST_Dataset} is $98.6 \%$ for adLIF and $98.5\%$ for RadLIF.}
\label{tab:sub:L2H_NMNIST_Dataset}
\resizebox{\textwidth}{!}{%
\begin{tabular}{mllllllll}
 \hline
None &95.4 $\pm$ 0.7 \% & 97.2 $\pm$ 0.4 \% & 83.9 $\pm$ 2.3 \% & 93.3 $\pm$ 2.5 \% & 72.8 $\pm$ 3.8 \% & 86.3 $\pm$ 6.2 \% & 47.9 $\pm$ 3.9 \% & 18.5 $\pm$ 10.7 \% \\
Expectation &  \textbf{98.5 $\pm$ 0.1 \%} & \textbf{98.5 $\pm$ 0.1 \%} & 98.1 $\pm$ 0.2 \% & \textbf{98.4 $\pm$ 0.2 \%} & 94.9 $\pm$ 0.9 \% & \textbf{98.3 $\pm$ 0.2 \%} & 76.9 $\pm$ 5.2 \% & \textbf{96.0 $\pm$ 1.2 \%} \\
Integral &  \textbf{98.5 $\pm$ 0.1 \%} & \textbf{98.5 $\pm$ 0.1 \% }& 98.1 $\pm$ 0.2 \% & \textbf{98.4 $\pm$ 0.2 \%} & 94.9 $\pm$ 0.9 \% & \textbf{98.3 $\pm$ 0.2 \%} & 76.9 $\pm$ 5.2 \% & \textbf{96.0 $\pm$ 1.2 \%} \\
Euler & 98.4 $\pm$ 0.1 \% & 98.3 $\pm$ 0.1 \% & \textbf{98.5 $\pm$ 0.1 \%} & 98.2 $\pm$ 0.2 \% & \textbf{98.5 $\pm$ 0.1 \%} & 98.2 $\pm$ 0.2 \% & \textbf{97.7 $\pm$ 0.2 \%} & 94.6 $\pm$ 2.4 \% \\
Time-const. & 97.2 $\pm$ 0.3 \% & 73.8 $\pm$ 15.5 \% & 89.4 $\pm$ 5.3 \% & 23.6 $\pm$ 17.8 \% & 46.0 $\pm$ 13.0 \% & 13.4 $\pm$ 5.1 \% & 20.8 $\pm$ 6.3 \% & 10.5  $\pm$ 0.9 \% \\
\hline
\end{tabular}
}
\end{subtable}
\end{table*}

\begin{table*}
\centering
\caption{Performance under various domain adaptation methods on target data with coarser temporal resolution ($\bT=2, 3, 4, 10$), where the models were trained on source data with fine time resolution ($\bS=1$). Each value in the table is average of $10$ trials.}
\label{tab:H2L}
\newcolumntype{l}{>{\centering\arraybackslash}p{2.4cm}}
\newcolumntype{m}{>{\centering\arraybackslash}p{2.2cm}}
\begin{subtable}[h]{1\linewidth}
\centering
\vspace{0.15cm}
\caption{Results using SHD dataset.}
\label{tab:sub:H2L_SHD_Dataset}
\resizebox{\textwidth}{!}{%
\begin{tabular}{mllllllll}
 \hline
     & \multicolumn{2}{c}{$\bS=1$ to $\bT=2$} 
     & \multicolumn{2}{c}{$\bS=1$ to $\bT=3$} 
     & \multicolumn{2}{c}{$\bS=1$ to $\bT=4$} 
     & \multicolumn{2}{c}{$\bS=1$ to $\bT=10$}  \\
 \cmidrule(lr){2-3} \cmidrule(lr){4-5} \cmidrule(lr){6-7} \cmidrule(lr){8-9}
 Model Adapt.&adLIF&RadLIF &adLIF&RadLIF &adLIF&RadLIF &adLIF&RadLIF \\
\end{tabular}
}
\end{subtable}
\begin{subtable}[h]{1\textwidth}
\centering
\resizebox{\textwidth}{!}{%
\begin{tabular}{mllllllll}
 \hline
  None & 56.9 $\pm$ 1.6 \% & 45.4 $\pm$ 2.5 \% & 26.7 $\pm$ 2.2 \% & 21.5 $\pm$ 1.4 \% & 18.4 $\pm$ 1.9 \% & 13.0 $\pm$ 1.7 \% & 10.7 $\pm$ 1.5 \% & 8.2 $\pm$ 2.0 \% \\
Expectation & 75.1 $\pm$ 2.2 \% & \textbf{76.0 $\pm$ 2.6 \%} & 52.0 $\pm$ 2.8 \% & \textbf{45.6 $\pm$ 4.0 \%} & 34.3 $\pm$ 2.5 \% & \textbf{29.3 $\pm$ 3.9 \%} & 14.4 $\pm$ 1.8 \% & \textbf{9.0 $\pm$ 1.1 \%} \\
Integral  & 75.1 $\pm$ 2.2 \% & \textbf{76.0 $\pm$ 2.6 \%} & 52.0 $\pm$ 2.8 \% & \textbf{45.6 $\pm$ 4.0 \%} & 34.3 $\pm$ 2.5 \% & \textbf{29.3 $\pm$ 3.9 \%} & 14.4 $\pm$ 1.8 \% & \textbf{9.0 $\pm$ 1.1 \%} \\
Euler  & 55.5 $\pm$ 4.7 \% & 50.0 $\pm$ 9.3 \% & 7.1 $\pm$ 1.0 \% & 7.3 $\pm$ 1.7 \% & 5.8 $\pm$ 0.8 \% & 5.7 $\pm$ 0.8 \% & 6.2 $\pm$ 1.0 \% & 4.7 $\pm$ 1.1 \% \\
Time-const.  & \textbf{80.1 $\pm$ 2.2 \%} & 54.4 $\pm$ 11.2 \% & \textbf{55.0 $\pm$ 4.3 \%} & 19.5 $\pm$ 8.2 \% & \textbf{38.4 $\pm$ 3.7 \%} & 11.2 $\pm$ 4.4 \% & \textbf{17.3 $\pm$ 2.5 \%} & 6.8 $\pm$ 1.4 \% \\
\hline
\end{tabular}
}
\end{subtable}
\begin{subtable}[h]{1\textwidth}
\centering
\vspace{0.15cm}
\caption{Results using MSWC dataset.}
\label{tab:sub:H2L_MSWC_Dataset}
\resizebox{\textwidth}{!}{%
\begin{tabular}{mllllllll}
 \hline
None & 54.1 $\pm$ 2.6 \% & 32.0 $\pm$ 3.3 \% & 13.4 $\pm$ 1.4 \% & 2.5 $\pm$ 0.5 \% & 4.6 $\pm$ 0.4 \% & 1.4 $\pm$ 0.2 \% & 1.4 $\pm$ 0.2 \% & 1.1 $\pm$ 0.1 \% \\
Expectation  & \textbf{59.4 $\pm$ 12.5 \%} & 42.7 $\pm$ 5.0 \% & \textbf{25.9 $\pm$ 8.7 \%} & 4.0 $\pm$ 1.1 \% & \textbf{13.1 $\pm$ 3.3 \%} & 2.0 $\pm$ 0.4 \% & \textbf{2.7 $\pm$ 0.3 \%} & 1.1 $\pm$ 0.2 \% \\
Integral  & \textbf{59.4 $\pm$ 12.5 \%} & 42.7 $\pm$ 5.0 \% & \textbf{25.9 $\pm$ 8.7 \%} & 4.0 $\pm$ 1.1 \% & \textbf{13.1 $\pm$ 3.3 \%} & 2.0 $\pm$ 0.4 \% & \textbf{2.7 $\pm$ 0.3 \%} & 1.1 $\pm$ 0.2 \% \\
Euler  & 45.6 $\pm$ 5.1 \% & \textbf{49.2 $\pm$ 4.4 \%} & 6.9 $\pm$ 2.2 \% & \textbf{5.0 $\pm$ 1.2 \%} & 2.5 $\pm$ 0.8 \% & \textbf{2.4 $\pm$ 0.5 \%} & 1.4 $\pm$ 0.3 \% & \textbf{1.2 $\pm$ 0.1 \%} \\
Time-const. &55.5 $\pm$ 15.3 \% & 6.8 $\pm$ 1.3 \% & 19.9 $\pm$ 9.2 \% & 1.5 $\pm$ 0.3 \% & 7.9 $\pm$ 3.7 \% & 1.2 $\pm$ 0.2 \% & 2.0 $\pm$ 0.5 \% & 1.0 $\pm$ 0.2 \% \\
\hline
\end{tabular}
}
\end{subtable}
\begin{subtable}[h]{1\textwidth}
\centering
\vspace{0.15cm}
\caption{Results using NMNIST dataset.}
\label{tab:sub:H2L_NMNIST_Dataset}
\resizebox{\textwidth}{!}{%
\begin{tabular}{mllllllll}
 \hline
None  & 96.4 $\pm$ 0.7 \% & 97.5 $\pm$ 0.3 \% & 92.7  $\pm$ 1.8\% & \textbf{96.3 $\pm$ 0.6 \%} & 88.3 $\pm$ 2.8 \% & \textbf{94.8 $\pm$ 1.2 \%} & 60.1 $\pm$ 5.8 \% & 74.2 $\pm$ 6.3 \% \\
Expectation  & \textbf{98.3 $\pm$ 0.2 \%} & \textbf{98.0 $\pm$ 0.3 \%} & 97.2 $\pm$ 0.5 \% & \textbf{96.3 $\pm$ 1.1 \%} & 95.3 $\pm$ 1.1 \% & 94.3 $\pm$ 2.0 \% & 83.7 $\pm$ 4.0 \% & \textbf{89.9 $\pm$ 3.5 \%} \\
Integral & \textbf{98.3 $\pm$ 0.2 \%} & \textbf{98.0 $\pm$ 0.3 \%} & 97.2 $\pm$ 0.5 \% & \textbf{96.3 $\pm$ 1.1 \%} & 95.3 $\pm$ 1.1 \% & 94.3 $\pm$ 2.0 \% & 83.7 $\pm$ 4.0 \% & \textbf{89.9 $\pm$ 3.5 \%} \\
Euler  & 88.9 $\pm$ 3.6 \% & 75.6 $\pm$ 12.6 \% & 88.6 $\pm$ 4.2 \% & 75.3 $\pm$ 11.3 \% & 87.6 $\pm$ 3.7 \% & 78.8 $\pm$ 10.3 \% & 73.5 $\pm$ 7.3 \% & 80.4 $\pm$ 9.3 \% \\
Time-const.  & 98.2 $\pm$ 0.2 \% & 81.2 $\pm$ 14.4 \% & \textbf{97.9 $\pm$ 0.2 \%} & 71.5 $\pm$ 17.0 \% & \textbf{97.7 $\pm$ 0.3 \%} & 70.6 $\pm$ 16.0 \% & \textbf{96.5 $\pm$ 0.7 \%} & 77.3 $\pm$ 8.6 \% \\
\hline
\end{tabular}
}
\end{subtable}
\end{table*}

\subsection{Coarse-to-Fine Temporal Resolution Domain Adaptation}
\label{sec:results:L2H}

In Table \ref{tab:L2H} we present the  performance of the proposed adaptation methods under Coarse-to-Fine deployment.
In all cases and datasets inspected, the proposed adaptation methods significantly outperform no-adaptation scenario and the benchmark Time-constant adaptation method. As discussed in Section \ref{sec:propositions}, Expectation and Integral adaptation methods provide identical scaling under certain common settings, hence their adaptation performance is almost identical. 
In most cases, the accuracy provided by Euler adaptation method is lower compared to Expectation and Euler, but still significantly higher than no-adaptation and Time-constant benchmark methods. 
As shown in Section \ref{sec:propositions}, the Euler method is based on the first-order approximation of the derivative, and a simpler approach compared to the Expectation and Integral method. The lower accuracy of Euler method compared to these two methods is consistent with higher information loss due to its simpler approach.

For all model adaptation methods, in general, the performance declines as the difference between the source and target temporal resolutions increases. This is consistent with the growing mismatch between the source and target feature space, which hinders the model’s ability to capture essential patterns, either because these features are not present or are inadequately represented in the source data.

We now compare the adLIF and RadLIF columns of Table \ref{tab:L2H} and the corresponding baseline performance in Table \ref{tab:learn_from_scratch_bins_result}. We observe that in the majority of the cases, the temporal domain adaptation accuracy in the recurrent SNNs, i.e. RadLIF columns, has a larger drop of performance compared to the non-recurrent SNNs, i.e. adLIF columns. This difference might be due to the fact that we have associated the recurrent connection in the neuron dynamics in the recurrent SNNs with a linear external input in SSMs, see Section~\ref{sec:map_snn_lssm}. When there is no recurrent connection in SNN (adLIF) this association is not required, and hence the correspondence between SNNs and linear SSMs provides a (still approximate but) closer match.

We now compare the model adaptation performance under NMNIST with other datasets.
The accuracy drop, i.e. the gap between the accuracy in Table \ref{tab:L2H} and the baseline accuracy in Table \ref{tab:learn_from_scratch_bins_result}, is typically smaller for NMNIST than that for SHD and MSWC. For example, for the scenario of bin size $\bS=10$ to bin size $\bT=1$ with the best performing adaptation method under adLIF,  the accuracy drop is $21\%$, $5.6\%$, $0.9\%$, for SHD, MSWC and NMNIST, respectively. 
Possible reason for this behavior is again the static nature of the NMNIST data, where the time dimension does not encode any inherent temporal information unlike in the audio datasets where the time behavior is important for distinguishing between the different classes. 
However, including the NMNIST dataset in the evaluation has shown that our methods can be applied and expected to perform in different mediums of data origin.

\subsection{Fine-to-Coarse Temporal Resolution Domain Adaptation}
\label{sec:results:H2L}
In Table \ref{tab:H2L}, we present the Fine-to-Coarse deployment performance of the proposed parameters adaptation methods.
Comparing the results with the baseline performance in Table~\ref{tab:sub:learn_from_scratch_SHD_Dataset}, we observe that there is a relatively large gap in performance for all scenarios with SHD and MSWC, which suggests that the fine-to-coarse deployment with spiking neurons is challenging.  
Now we compare our proposed method with the no-adaptation baseline in Table \ref{tab:H2L}.
The proposed Integral and Expectation methods consistently outperform the no-adaptation baseline or  demonstrate statistically equivalent performance considering the standard deviations. 
In contrast, the average performance of the Euler method is generally inferior to that of the no-adaptation baseline, although in some cases it remains within the standard deviation range of the baseline. 
Now we compare our proposed method with the time-constant adaptation method.
Under adLIF, time-constant adaptation method provides the best performance in several cases whereas under RadLIF the proposed methods perform better than time-constant adaptation method in all scenarios, indicating that the proposed methods can address the challenge of scaling the recurrent external feedback in a better manner.

\subsection{Time efficient training}
\label{sec:results:time_efficient_training}

We now investigate the trade-offs between accuracy and training time for pre-training the model on various temporal resolutions, i.e. varying $\bS$,  while evaluating its performance on $\bT=1$.

In the previous sections, the pre-trained models were obtained using a fixed number of training iterations ($50$) over the training dataset. To reveal the accuracy-complexity trade-offs better, here we introduce an additional pre-training setup that uses a validation set.
(If the dataset does not have  a pre-defined validation data, we extract a validation set from the training data, matching the size of the test set.) We implement early stopping training, where training continues until convergence based on validation data loss. Specifically, training is allowed up to $100$ iterations, with early termination if the reduction in validation loss is less than $10^{-5}$ for $10$ consecutive iterations. The best-performing model in the validation set is then saved, and its accuracy on the test dataset is reported.

\begin{figure}
     \centering
     \includegraphics[width=1\linewidth]{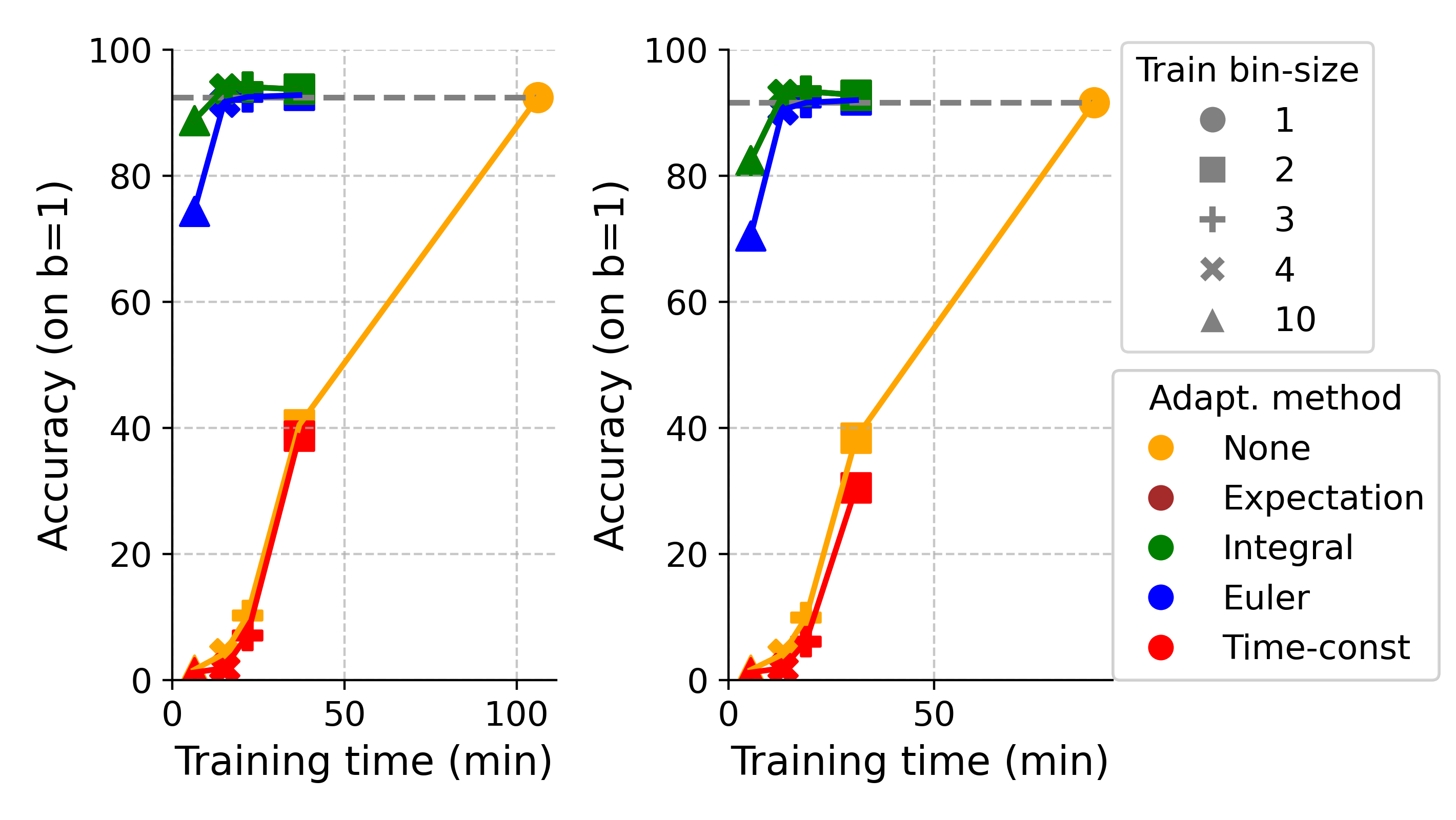} 
    \caption{Both plots show training + adaptation (wall-clock) time vs accuracy for the MSWC dataset and adLIF neuron, where the left plot uses fixed 50 epochs pre-training (results in Table \ref{tab:sub:L2H_MSWC_Dataset}), while the right plot uses early stopping pre-training. 
    The gray dashed line represent the baseline performance of pre-training the model on the target resolution. 
    Note that lines for the Integral and Expectation adaptation methods  are overlapping.}
    \label{fig:comp_complex}
\end{figure}

We focus on the MSWC dataset and adLIF neuron for illustration. In Figure~\ref{fig:comp_complex}, the accuracy on bin size $\bT=1$ is plotted against the total training time, where the total training time is the sum of the training time at source resolution $b_s$ (Step 1 of Algorithm~\ref{alg:source-to-target-adapt}) and the adaptation time if it is applied (Step 2 of Algorithm~\ref{alg:source-to-target-adapt}).

We observe that for both figures, i.e. the fixed epochs pre-training and the early-stopping pre-training, the general characteristics of the curves are the same. In particular, they show that significant gains can be obtained by pre-training lower resolutions.
For instance, consider the results with the early stopping procedure. 
Instead of training a model on data with bin size $1$ until convergence, an alternative approach is to train on bin size $\bS=4$ and apply the Integral adaptation method to achieve a model with high-accuracy on data with bin size $\bT=1$.
In particular, we obtain an accuracy of $93.8 \% $ instead of the baseline performance of $92.4 \%$ while training $106/15 \approx 7$ times faster.

We note that the adaptation procedure (Step~2 of Algorithm \ref{alg:source-to-target-adapt}) takes a small fraction of the total time. For example, for the setting in Figure~\ref{fig:comp_complex}, it takes a maximum of $0.012$ minute to adapt the parameters of all neurons in the network using any of our proposed methods, whereas training for a single epoch with bin size $1$ requires $\approx 106/50 = 2.12$ minutes. Hence, consistent with the discussions in Section \ref{sec:cost_comparison_adapt_and_train}, adaptation is significantly more efficient than re-training.

These results suggest that training on low temporal resolution datasets and adapting models using the proposed methods may significantly reduce training time with minimal to no impact on accuracy when source and target resolutions are close, with no re-training on the target dataset.

\begin{table*}
\centering
\small
\caption{Comparison between the performance contribution of model adaptation and normalization layer adaptation, for Coarse-to-Fine scenario with MSWC dataset.}
\label{tab:ablation_study}
\newcolumntype{l}{>{\centering\arraybackslash}p{1.5cm}}
\newcolumntype{n}{>{\centering\arraybackslash}p{1.5cm}}
\newcolumntype{m}{>{\centering\arraybackslash}p{2.5cm}}
\resizebox{\textwidth}{!}{%
\begin{tabularx}{1\linewidth}{lnlmmmm}
\hline
 \multirow{2}{*}{Model}  & \multirow{2}{*}{Stats used}  & \multirow{2}{*}{Norm.}  & \multicolumn{2}{c}{$\bS=1$ to $\bT=2$} & \multicolumn{2}{c}{$\bS=2$ to $\bT=1$} \\
 \cmidrule(lr){4-5} \cmidrule(lr){6-7}
 Adapt.  & for norm. & Adapt.  & adLIF & RadLIF & adLIF & RadLIF \\
 \hline
Integral & Source & None &37.7 $\pm$ 1.7&51.4 $\pm$ 3.4&25.5 $\pm$ 2.5&20.2 $\pm$ 7.1\\
Integral & Source & Adapt &75.4 $\pm$ 2.3&75.8 $\pm$ 2.8&89.5 $\pm$ 0.6&87.6 $\pm$ 2.4\\
Integral & Target & N/A &83.2 $\pm$ 1.5&86.5 $\pm$ 1.0&91.5 $\pm$ 0.6&90.2 $\pm$ 1.7\\
None & Source & None &29.2 $\pm$ 2.5&30.0 $\pm$ 2.7&21.2 $\pm$ 2.4&8.2 $\pm$ 3.1\\
None & Source & Adapt &57.1 $\pm$ 1.7&45.3 $\pm$ 2.5&58.2 $\pm$ 1.7&31.6 $\pm$ 4.6\\
None & Target & N/A &71.2 $\pm$ 1.7&60.0 $\pm$ 2.1&66.7 $\pm$ 1.3&45.8 $\pm$ 2.4\\
\hline
\end{tabularx}
}
\end{table*}

\section{Ablation study}
\label{sec:ablation_study}

In Section~\ref{sec:ablation_study:norm_layer} we investigate the effect of adapting the normalization layer to the change of temporal resolution in the data. In Section~\ref{sec:ablation_study:input_adapt} we investigate different types of adaptation, where instead of adapting the models as in our proposed methods, we adapt the input to the model.
In Section~\ref{sec:various_to_coarse_tranformations}, we investigate the performance when the coarse resolution sequences are obtained using different methods.
We evaluate the impact of spike sparsity on the performance of our methods in Section~\ref{sec:sparsity}.
We discuss an alternative method for linearization of the neuron dynamics in Section~\ref{sec:deviation_Hf_in_A}.

\subsection{Normalization layer adaptation}
\label{sec:ablation_study:norm_layer}

In the previous section, we utilize model adaptation of Section \ref{sec:proposedMethods:main} and normalization layer adaptation, of Section \ref{sec:num:setting:batchnorm_scaling} together. We now evaluate the effect of these components separately. 

Table \ref{tab:ablation_study} provides the results for Coarse-to-Fine ($\bS=2$ to $\bT=1$) and Fine-to-Coarse ($\bS=1$ to $\bT=2$) deployment for SHD dataset. ``Stats used for normalization'' indicates the data statistics used in the normalization layers of the model. Recall that results presented previously correspond to the row ``Integral-Source-Adapt''. Note that the label ``target'' represents the case where statistics of the target dataset is used for normalization. This represents a potentially impractical but useful high-performance baseline since target data statistics provide the ideal normalization for the target dataset but cannot be always calculated.

We now focus on the cases where target data is used for normalization. We observe that when the Integral model adaptation method is applied, a significant performance improvement  compared to the case of no model adaptation is obtained, e.g. an improvement from $66.7 \%$ to $91.5 \%$ under $\bS=2$ to $\bT=1$ with adLIF.  This indicates that accurate normalization alone cannot fully address the feature shift between source and target data.

Furthermore, examining the cases where the source data statistics are used with no normalization adaptation, for example in the case of $\bS=1$ to $\bT=2$ and RadLIF, we again observe a significant performance improvement from $30.0 \%$ to $51.4 \%$ when the Integral model adaptation method is applied compared to the no model adaptation. This indicates that even in the case of no normalization the proposed model adaptation method offer significant improvement.

These results suggest that, 
in practical applications where target data statistics are inaccessible, the combination of our model adaptation methods and proposed normalization adaptation should be preferred. This combination is used in the previous sections, such as in Table \ref{tab:L2H} and  Table \ref{tab:H2L}.

\subsection{Input Adaptation}
\label{sec:ablation_study:input_adapt}

Our proposed adaptation methods adapt the model to account for the change in the temporal resolution of the source and target data. In this section, we look into the scenario where the model is kept fixed but instead we transform the target data to mimic the temporal resolution of the source data. In this approach, the length of the target data is adjusted at deployment time to match the length of the source data. 

The normalization layers adaptation for the input adaptation method presented in this section are stated in Section \ref{sec:norm_adapt_in_experiment}.

\subsubsection{Coarse-to-Fine Input Adaptation}
\label{sec:input-adapt:L2H}

\begin{table}
\centering
\small
\caption{Performance of input adaptation for Course-to-Fine experiments on the SHD dataset.}
\label{tab:Ablation_study:L2H}
\newcolumntype{m}{>{\centering\arraybackslash}p{3cm}}
\newcolumntype{l}{>{\centering\arraybackslash}p{2cm}}
\resizebox{\columnwidth}{!}{%
\begin{tabular}{mll}
 \hline
 & \multicolumn{2}{c}{$\bS=2$ to $\bT=1$}  \\
 \cmidrule(lr){2-3}
Input Adapt.&adLIF&RadLIF \\
 \hline
Binary-sum-bin & 36.6 $\pm$ 2.1 \% & 40.6 $\pm$ 2.5 \% \\
Max-pool & 83.2 $\pm$ 1.0 \% & 85.2 $\pm$ 1.1 \% \\
\hline
\end{tabular}
}
\end{table}

In this case, the models has been trained on coarse resolution with $\NS$ time steps, but at deployment, we are given fine-resolution target data with $\NT$ time steps where $\NT=\NRatio\times \NS$ with $\NRatio\in\Z$. Hence we want to downsample the target dataset by a factor of $\NRatio$ so that it matches the number of time steps in the source data and mimics the coarse resolution. 
Specifically, we consider the following input transformations:
\begin{itemize}
    \item Binary-sum-binning Input Adaptation Method: This method, similar to the sum-binning in Section \ref{sec:num:setting:sum_binning}, sums every $\NRatio$ consecutive elements using a moving window (non-overlapping) and applies indicator function to the sum, hence for $\NRatio=2$, the sequence $(x, y, z, w)$ becomes $(I(x+y), I(z+w))$ where $I$ is the indicator function.
    
    \item Max-pooling Input Adaptation Method:  This method compares every $\NRatio$ consecutive elements using a non-overlapping moving window and replaces them with a single value representing the maximum of them.  Hence, for $\NRatio=2$, the sequence $(x, y, z, w)$ becomes $(\max(x, y), \max(z, w))$.
\end{itemize}

The results for these methods are presented in Table~\ref{tab:Ablation_study:L2H}. 
Sum-binning (as used during training) can also be viewed as one of the input adaptation methods. Hence, input adaptation via sum-binning corresponds to the case already covered in Table~\ref{tab:learn_from_scratch_bins_result}, where the source and target data are identical. This provides a baseline performance when the method for obtaining coarse resolution data may be matched during training and test.  
Recall from Table~\ref{tab:sub:L2H_SHD_Dataset} that for the case $\bS=2$ to $\bT=1$ using model adaptation yielded performance of $89.5\%$ for adLIF and $87.7\%$ for RadLIF. Comparing these results with the ones in Table~\ref{tab:Ablation_study:L2H}, we observe that Integral and Expectation model adaptation methods outperform the Binary-sum-bin and Max-pool input adaptation methods.

\subsubsection{Fine-to-Coarse Input Adaptation}
\label{sec:input-adapt:H2L}
In this case, the models have been train on fine resolution with $\NS$ time steps, but at deployment, we are given coarse-resolution target data with $\NT$ time steps where $\NS=\DeltaRatio\times \NT$ with $\DeltaRatio\in\Z$. Hence we want to upsample the target dataset by a factor of $\DeltaRatio$ so that it matches the number of time steps in the source data and thus mimics the fine resolution. Specifically, we consider the following input transformations:
\begin{itemize}
    \item Pad-zeros Input Adaptation Method: This method prefixes each element in the sequence with $\DeltaRatio-1$ zeros, hence for $\DeltaRatio=2$, the sequence $(x, y, z, w)$ becomes $(0, x, 0, y, 0, z, 0, w)$.
    \item Repeat-elements Input Adaptation Method: This method repeats each element in the sequence $\DeltaRatio-1$ additional times, hence for $\DeltaRatio=2$, the sequence $(x, y, z, w)$ becomes $(x,x, y,y, z,z, w,w)$. 
\end{itemize}

The results of input adaptation for the Fine-to-Coarse setting are presented in Table~\ref{tab:Ablation_study:H2L}. Recall from Table~\ref{tab:sub:H2L_SHD_Dataset} that for the case $\bS=1$ to $\bT=2$ using model adaptation yielded performance of $80.1\%$ for adLIF and $76.0\%$ for RadLIF. Comparing input adaptation with model adaptation, Repeat-elem method for adLIF;  and Pad-zero and Repeat-elem methods for RadLIF surpass the performance achieved through model adaptation. However, it is important to note that although input adaptation shows better performance than model adaptation, it comes with the drawback of longer processing times since a longer time-sequence has to be processed in order to obtain the classification result. For instance, when going from $\bS=1$ to $\bT=2$, using the input-adaptation method will double the inference time since the length of the sequence in target domain after input adaptation is $2\times\NT$ compared to using the proposed method where the length of the target sequence remains $\NT$.

\begin{table}
\centering
\small
\caption{Performance of input adaptation methods for Fine-to-Coarse experiments on the SHD dataset.}
\label{tab:Ablation_study:H2L}
\newcolumntype{l}{>{\centering\arraybackslash}p{2.5cm}}
\newcolumntype{m}{>{\centering\arraybackslash}p{2cm}}
\resizebox{\columnwidth}{!}{%
\begin{tabular}{mll}
 \hline
 & \multicolumn{2}{c}{$\bS=1$ to $\bT=2$}  \\
 \cmidrule(lr){2-3}
Input Adapt.&adLIF&RadLIF \\
 \hline
Pad-zeros & 72.0 $\pm$ 4.1 \% & 88.9 $\pm$ 0.9 \% \\
Repeat-elem & 89.3 $\pm$ 0.7 \% & 89.9 $\pm$ 0.4 \% \\
\hline
\end{tabular}
}
\end{table}

\begin{figure}
\centering
     \begin{subfigure}[b]{1\linewidth}
     \centering
     \includegraphics[width=0.9 \textwidth]{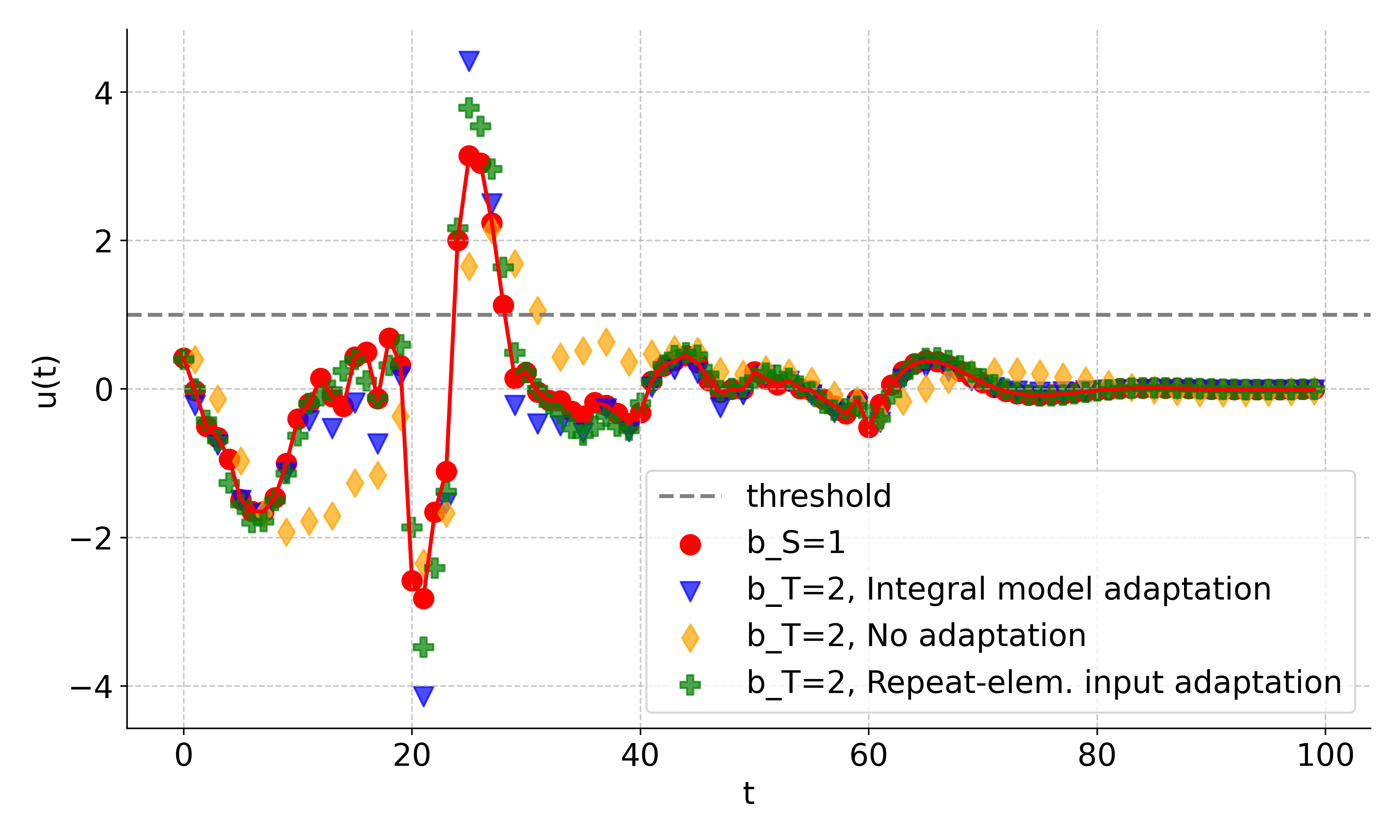}
     \caption{Neuron dynamics for a single neuron in the first hidden layer for one input sample of the SHD data.}
     \label{fig:H2L:Layer1-neuron}
     \end{subfigure}
\hfill
     \begin{subfigure}[b]{1\linewidth}
     \centering
     \includegraphics[width=0.9 \textwidth]{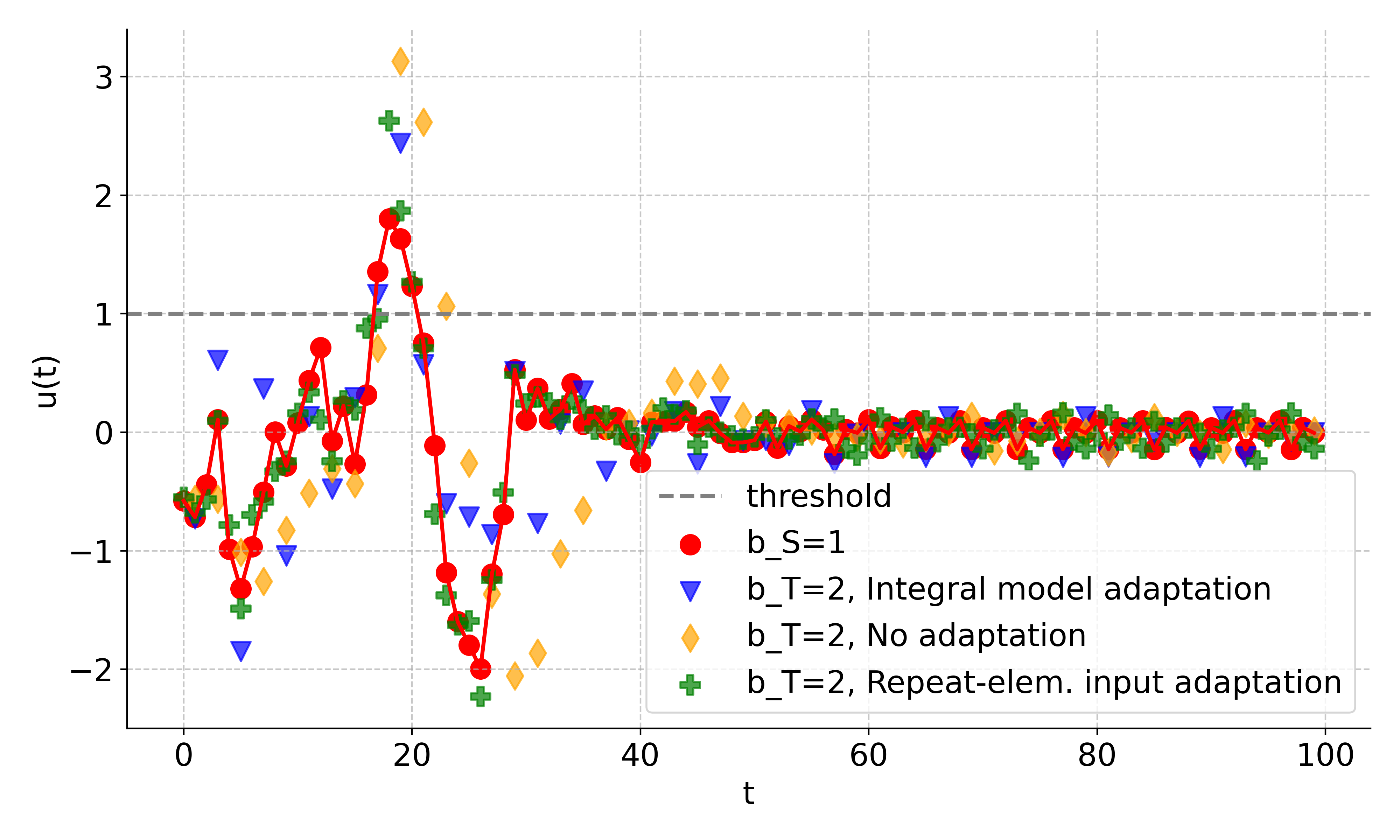}
     \caption{Neuron dynamics for a single neuron in the second hidden layer for one input sample of the SHD data.}
     \label{fig:H2L:Layer2-neuron}
     \end{subfigure}
\caption{Example of voltage dynamics over time in a pre-trained network on $\bS=1$. Figure shows voltage dynamics of Fine-to-Coarse $\bS=1$ to $\bT=2$ experiment for different adaptation methods and time resolutions.}
\label{fig:H2L:Single_Neuron_in_Network}
\end{figure}

\subsubsection{Neuron dynamics within the network}
To gain insights on   the  dynamics of the neurons under model and input adaptation, we now consider a neuron inside one of the networks used in this paper's experiments. In particular, we consider a model trained on the SHD dataset with $\bT=1$, and examine how it behaves when operating under $\bT=2$. Specifically, we take a sample from the SHD dataset and inspect the activity of one randomly selected neuron from the first hidden layer and one from the second hidden layer in  Figure~\ref{fig:H2L:Single_Neuron_in_Network}.

Figure~\ref{fig:H2L:Single_Neuron_in_Network} compares our Integral model adaptation, which processes coarse-resolution input, with the Repeat-elements input adaptation, which mimics fine-resolution data. 
As shown in Figure~\ref{fig:H2L:Layer1-neuron}, both the reference dynamics and the Repeat-elem method produce four output spikes, while the Integral method produces only two. Similarly, in Figure~\ref{fig:H2L:Layer2-neuron}, the reference and Repeat-elem generate three spikes, while the Integral method again outputs two.
These observations indicate that while both methods capture the main peaks in the neuron variable $\UmemNoArg[t]$, the Integral method is limited by the coarser temporal resolution, resulting in a lower number of spikes. In contrast, Repeat-elem operates on a longer sequence, providing more flexibility to match the finer details of the spiking pattern—reflected in the varying number of spikes across the two scenarios.

\subsection{Transformations for Obtaining Coarse Data}
\label{sec:various_to_coarse_tranformations}

In the previous experiments,  the sequences with coarse temporal resolution are obtained using sum-binning, see Section \ref{sec:num:setting:sum_binning}. In this section, we explore five alternative transformations for obtaining coarse temporal resolution sequences: Average-sum-bin, First-point-sample, Last-point-sample, Max-pool and Binary-sum-bin. Average-sum-bin is similar to sum-binning with an additional scaling of the coarse-resolution sequence by $\rho$. Hence,  Average-sum-bin is a scaled version of sum-bin. The First-point-sample and Last-point-sample perform simple index sampling of the fine-resolution sequence by using the first or last index in each bin, respectively. The Max-pool and Binary-sum-bin are  defined in Section \ref{sec:input-adapt:L2H}. The normalization layers adaptation for all these transformations is presented in Section \ref{sec:norm_adapt_in_experiment}.

As a baseline, 
we report the performance of SNN models trained and tested on the same time resolution, i.e., $\bS=\bT=2$, under different coarse resolution methods in Table \ref{tab:Ablation_study:VariousCoarseTransf:Baseline}. 
%
The table shows an average performance in the range of $87.3\%-91.0\%$ under all transformations.

We present the performance under the various coarse transformations using Integral method  and no adaptation for Fine-to-Coarse adaptation and  Coarse-to-Fine adaptation in Table \ref{tab:Ablation_study:VariousCoarseTransf:Fine-to-Coarse} and Table \ref{tab:Ablation_study:VariousCoarseTransf:Coarse-to-Fine}, respectively. 
We observe that performance of adaptation depends on the data representation in coarse resolution, where the adaptation performs weaker under Max-pool and Binary-sum-bin compared to other transformations.  
Nevertheless, our proposed adaptation method performs significantly better than no-adaptation for all coarse transformations and for both adLIF and RadLIF.
These experiments illustrate that the proposed methods may achieve higher performance compared to no-adaptation across different coarse temporal resolution transformations.

\begin{table}
\centering
\small
\caption{Baseline performance $\bS=2$ to $\bT=2$ on the SHD dataset for various Coarse data transformation.}
\label{tab:Ablation_study:VariousCoarseTransf:Baseline}
\newcolumntype{l}{>{\centering\arraybackslash}p{2cm}}
\newcolumntype{m}{>{\centering\arraybackslash}p{3.3cm}}
\resizebox{\columnwidth}{!}{%
\begin{tabular}{mll}
 \hline
Coarse Transformation & adLIF & RadLIF \\
\hline
Sum-bin & 91.0 $\pm$ 0.5 \% & 91.0 $\pm$ 0.7  \% \\
Average-sum-bin &  90.6 $\pm$ 0.5 \% & 90.8 $\pm$ 0.8  \% \\
First-point-sample & 87.3 $\pm$ 0.7 \% & 87.8 $\pm$ 0.6  \% \\
Last-point-sample &  88.3 $\pm$ 0.7 \% & 89.0 $\pm$ 0.9  \% \\
Max-pool & 90.7 $\pm$ 0.7 \% & 90.2 $\pm$ 0.5  \% \\
Binary-sum-bin & 87.9 $\pm$ 0.7 \% & 89.1 $\pm$ 0.3  \% \\
\hline
\end{tabular}
}
\end{table}

\begin{table*}
\centering
\small
\newcolumntype{l}{>{\centering\arraybackslash}p{2.5cm}}
\newcolumntype{m}{>{\centering\arraybackslash}p{4cm}}
\caption{Performance on the SHD dataset under coarse data obtained with varying transformations.}
\begin{subtable}[h]{1\linewidth}
    \caption{Fine-to-Coarse $\bS=1$ to $\bT=2$. }
    \label{tab:Ablation_study:VariousCoarseTransf:Fine-to-Coarse}
    \resizebox{\textwidth}{!}{%
    \begin{tabularx}{1\linewidth}{mllll}
    \hline
        & \multicolumn{2}{c}{Integral} 
        & \multicolumn{2}{c}{No-adaptation} \\
    \cmidrule(lr){2-3} \cmidrule(lr){4-5}
     Coarse Transformation  & adLIF & RadLIF & adLIF & RadLIF \\
    \hline
    Sum-bin & 75.4 $\pm$ 2.3 \% & 75.8 $\pm$ 2.8 \% & 57.1 $\pm$ 1.7 \% & 45.3 $\pm$ 2.5  \% \\
    Average-sum-bin & 75.4 $\pm$ 2.3 \% & 75.8 $\pm$ 2.8 \% & 57.1 $\pm$ 1.7 \% & 45.3 $\pm$ 2.5  \% \\
    First-point-sample & 69.6 $\pm$ 2.3 \% & 72.0 $\pm$ 2.9 \% & 55.3 $\pm$ 1.3 \% & 46.1 $\pm$ 2.1  \% \\
    Last-point-sample & 68.7 $\pm$ 3.0 \% & 72.3 $\pm$ 2.7 \% & 55.5 $\pm$ 1.2 \% & 44.7 $\pm$ 2.3  \% \\
    Max-pool & \, 54.5 $\pm$ 12.1 \% & 57.2 $\pm$ 9.5 \% & 41.1 $\pm$ 4.5 \% & 34.6 $\pm$ 4.0  \% \\
    Binary-sum-bin & 28.0 $\pm$ 4.8 \% & 25.9 $\pm$ 2.3 \% & 19.7 $\pm$ 3.6 \% & 17.2 $\pm$ 2.4  \% \\
    \hline
    \end{tabularx}
    }
\end{subtable}
\begin{subtable}[h]{1\linewidth}
    \caption{Coarse-to-Fine $\bS=2$ to $\bT=1$.}
    \label{tab:Ablation_study:VariousCoarseTransf:Coarse-to-Fine}
    \resizebox{\textwidth}{!}{%
    \begin{tabularx}{1\linewidth}{mllll}
    \hline
        & \multicolumn{2}{c}{Integral} 
        & \multicolumn{2}{c}{No-adaptation} \\
    \cmidrule(lr){2-3} \cmidrule(lr){4-5}
        Coarse Transformation & adLIF & RadLIF & adLIF & RadLIF \\
    \hline
    Sum-bin & 89.6 $\pm$ 0.6 \% & 89.2 $\pm$ 0.7 \% & 51.6 $\pm$ 1.3 \% & 32.8 $\pm$ 2.7 \% \\
    Average-sum-bin & 89.3 $\pm$ 0.6 \% & 89.5 $\pm$ 1.2 \% & 51.0 $\pm$ 1.4 \% & 33.8 $\pm$ 2.1 \% \\
    First-point-sample & 86.3 $\pm$ 0.6 \% & 87.0 $\pm$ 1.2 \% & 47.0 $\pm$ 1.0 \% & 25.9 $\pm$ 6.3 \% \\
    Last-point-sample & 87.9 $\pm$ 1.0 \% & 87.4 $\pm$ 1.1 \% & 45.3 $\pm$ 1.3 \% & 23.8 $\pm$ 4.1 \% \\
    Max-pool & 73.7 $\pm$ 1.2 \% & 68.4 $\pm$ 6.4 \% & 38.9 $\pm$ 1.8 \% & 17.2 $\pm$ 4.5 \% \\
    Binary-sum-bin & 60.0 $\pm$ 2.9 \% & 53.0 $\pm$ 2.2 \% & 33.5 $\pm$ 2.2 \% & 24.3 $\pm$ 2.4 \% \\
    \hline
    \end{tabularx}
    }
\end{subtable}
\end{table*}

\subsection{Spike Sparsity}
\label{sec:sparsity}

In this section, we evaluate the performance of the proposed adaptation methods across different spiking rates. As discussed in Section~\ref{sec:map_snn_lssm}, a neuron in a SNN reduces to a linear SSM when there is no nonlinearity in the neuron, where the spiking function of \eqref{eqn:general_LIF_intro:line2} is the primary source of nonlinearity. Hence, as neurons spike more frequently, the linear approximation is expected to become less accurate, which is expected to degrade the performance of the proposed methods. We now explore this phenomena.

\begin{figure}
\centering
     \begin{subfigure}[b]{0.95\linewidth}
     \centering
     \caption{Spike rate for both source and target resolution data}
     \includegraphics[width=\textwidth]{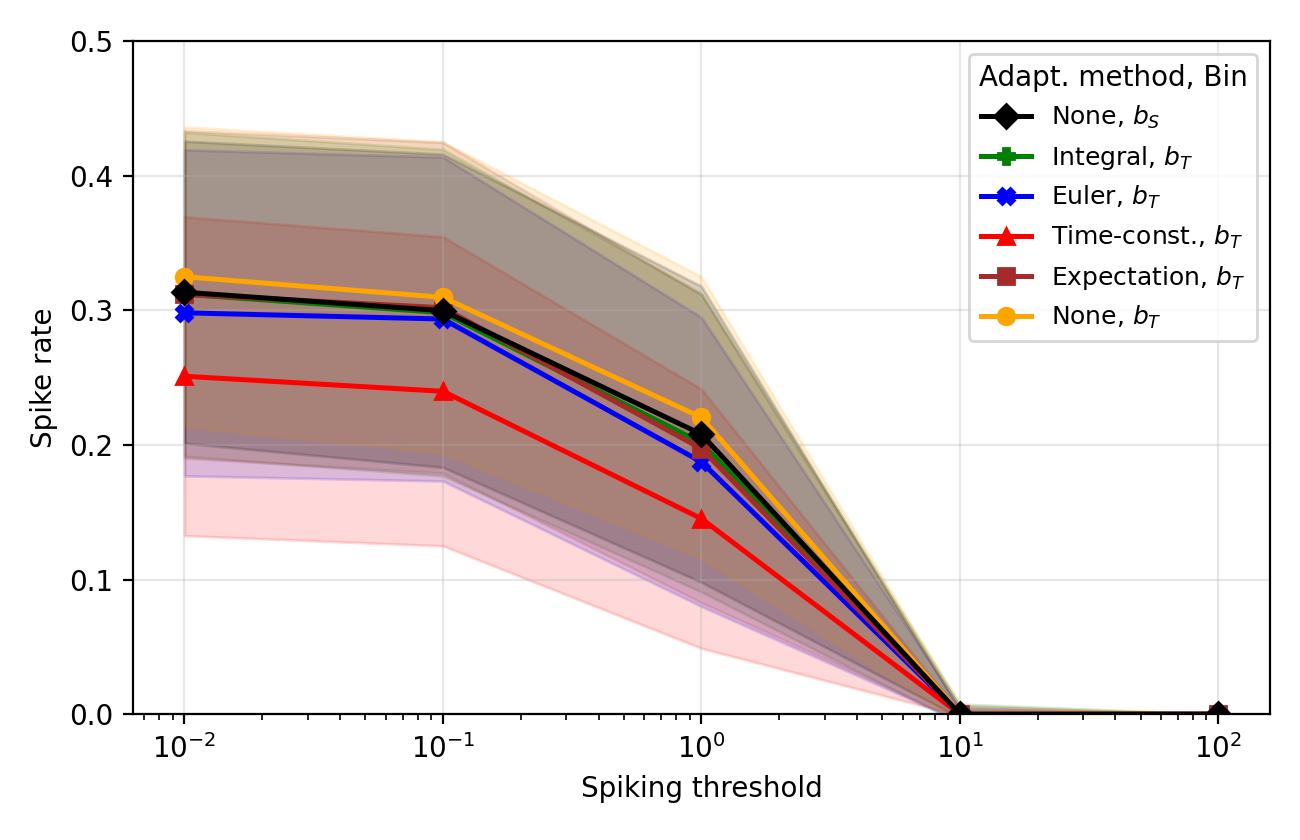}
     \label{fig:SpkRate:Coarse-to-Fine:T_spk}
     \end{subfigure}
\hfill
     \begin{subfigure}[b]{0.95\linewidth}
     \centering
     \caption{$\Qone$ between source and target dynamics}
     \includegraphics[width=\textwidth]{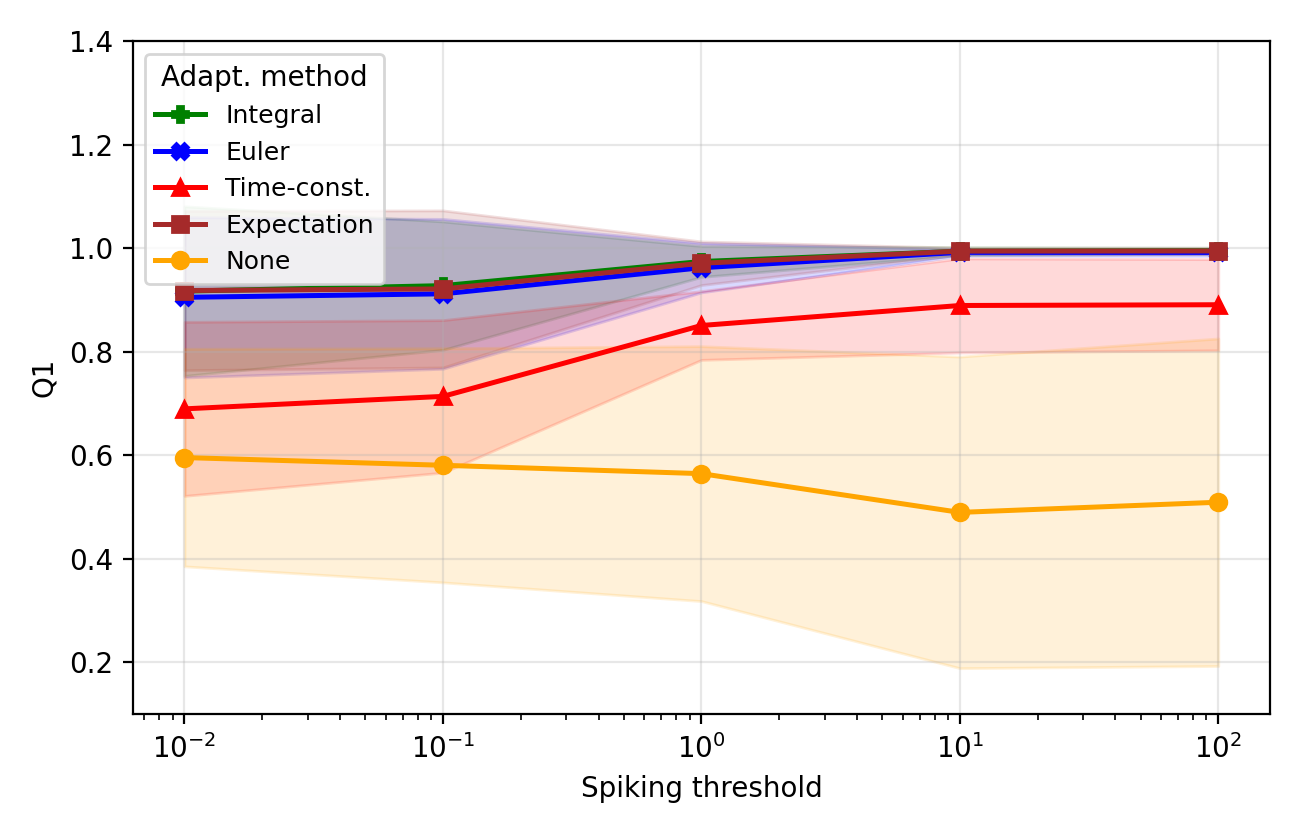}
     \label{fig:SpkRate:Coarse-to-Fine:Q1}
     \end{subfigure}
\hfill
     \begin{subfigure}[b]{0.95\linewidth}
     \centering
     \caption{$\Qtwo$ between source and target dynamics}
     \includegraphics[width=\textwidth]{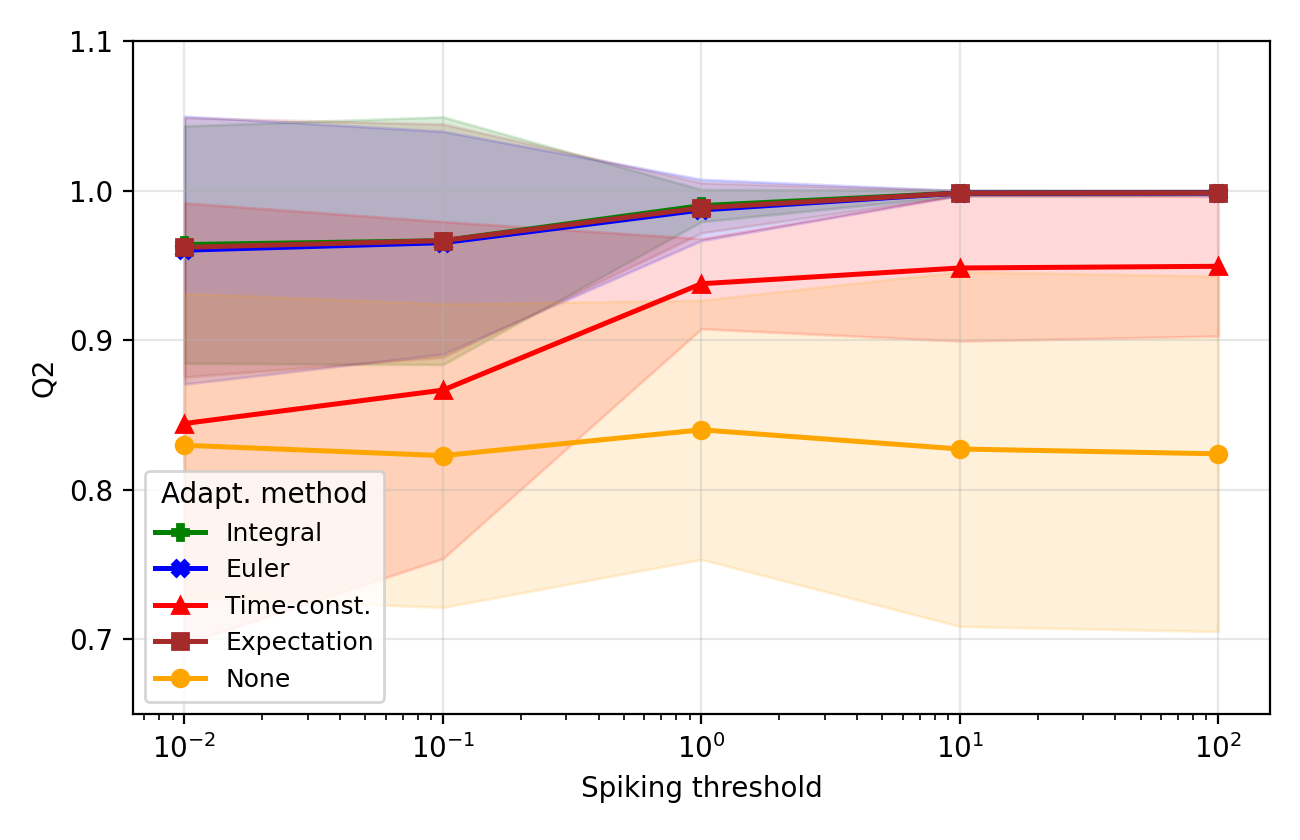}
     \label{fig:SpkRate:Coarse-to-Fine:Q2}
     \end{subfigure}     
\caption{Coarse-to-Fine scenario for $\bS=2$ and $\bT=1$. The solid line indicates the mean, and the shaded region represents one standard deviation around the mean.}
\label{fig:SpkRate:Coarse-to-Fine}
\end{figure}

\begin{figure}
\centering
     \begin{subfigure}[b]{0.95\linewidth}
     \centering
     \caption{Spike rate for both source and target resolution data}
     \includegraphics[width=\textwidth]{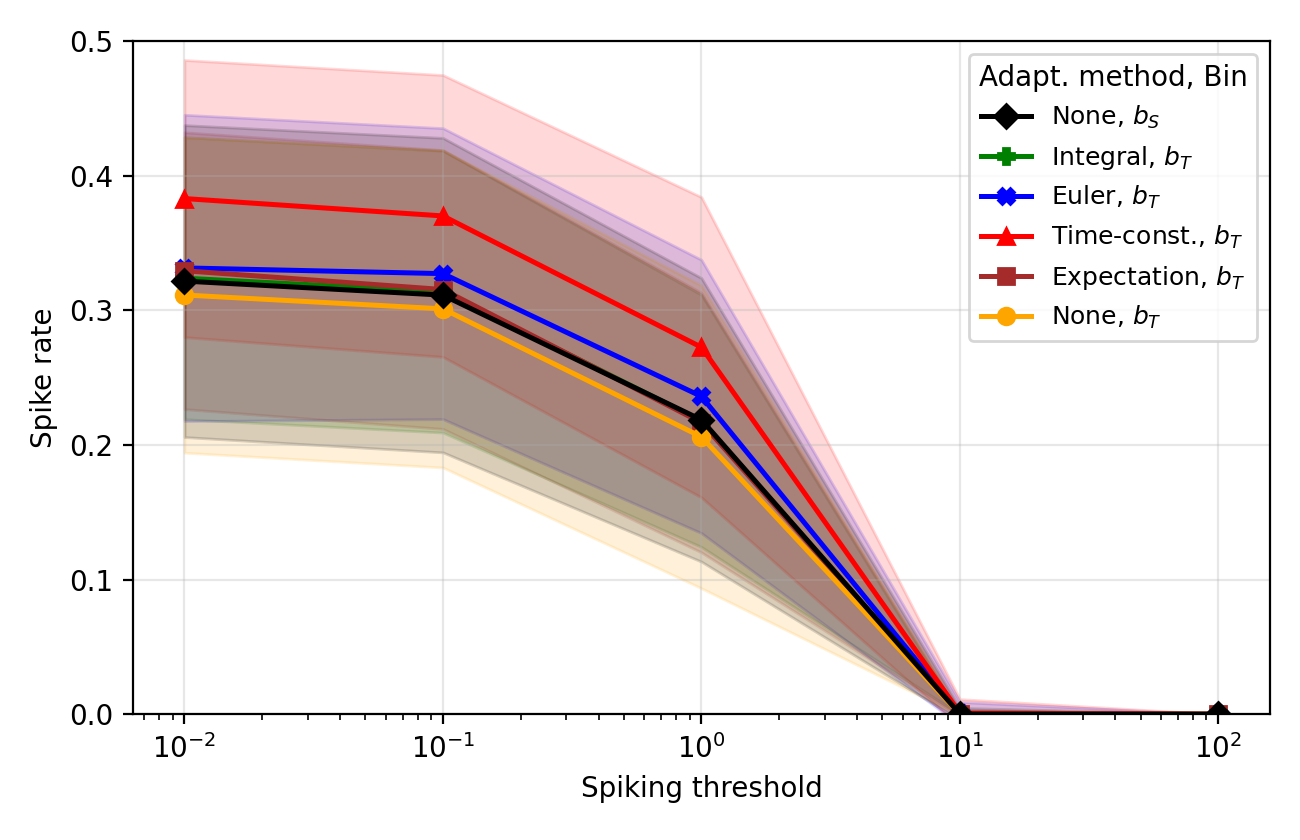}
     \label{fig:SpkRate:Fine-to-Coarse:T_spk}
     \end{subfigure}
\hfill
     \begin{subfigure}[b]{0.95\linewidth}
     \centering
     \caption{$\Qone$ between source and target dynamics}
     \includegraphics[width=\textwidth]{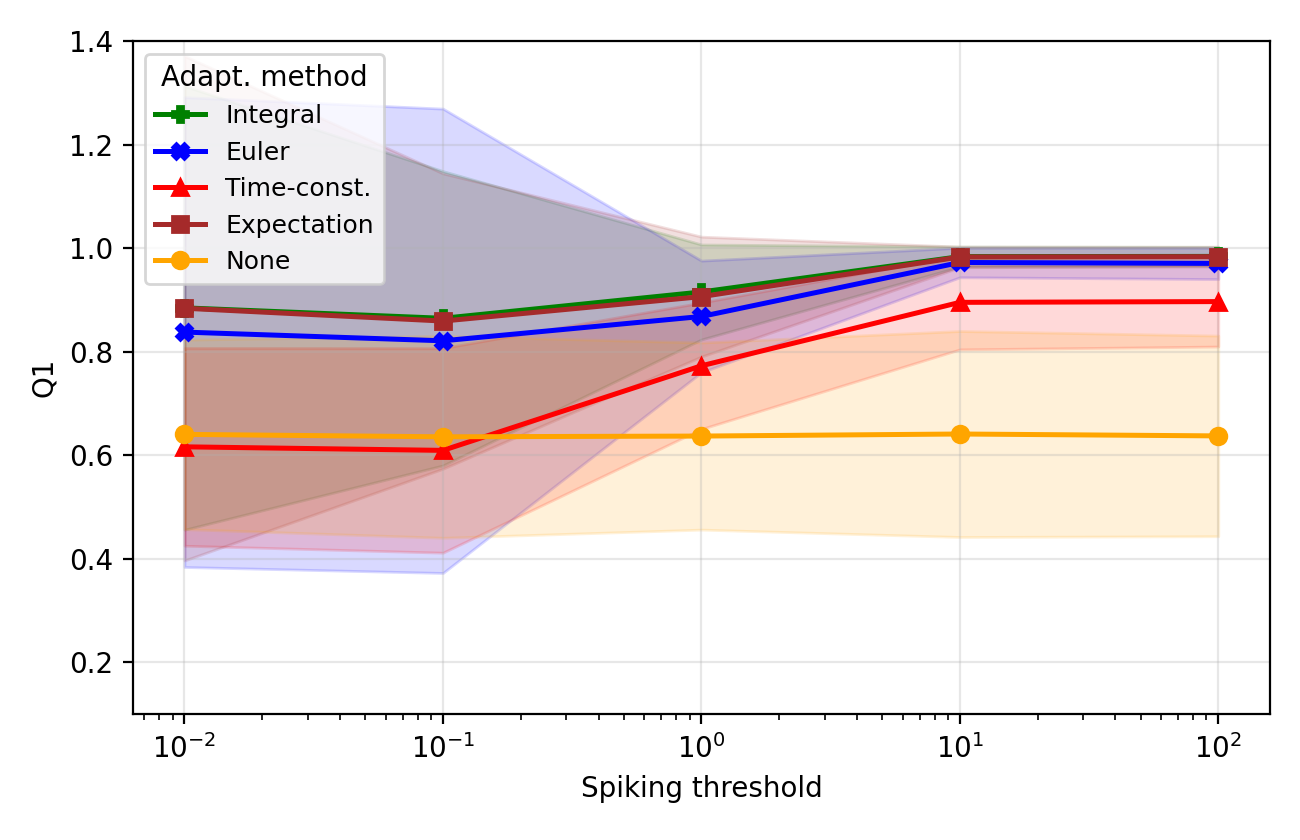}
     \label{fig:SpkRate:Fine-to-Coarse:Q1}
     \end{subfigure}
\hfill
     \begin{subfigure}[b]{0.95\linewidth}
     \centering
     \caption{$\Qtwo$ between source and target dynamics}
     \includegraphics[width=\textwidth]{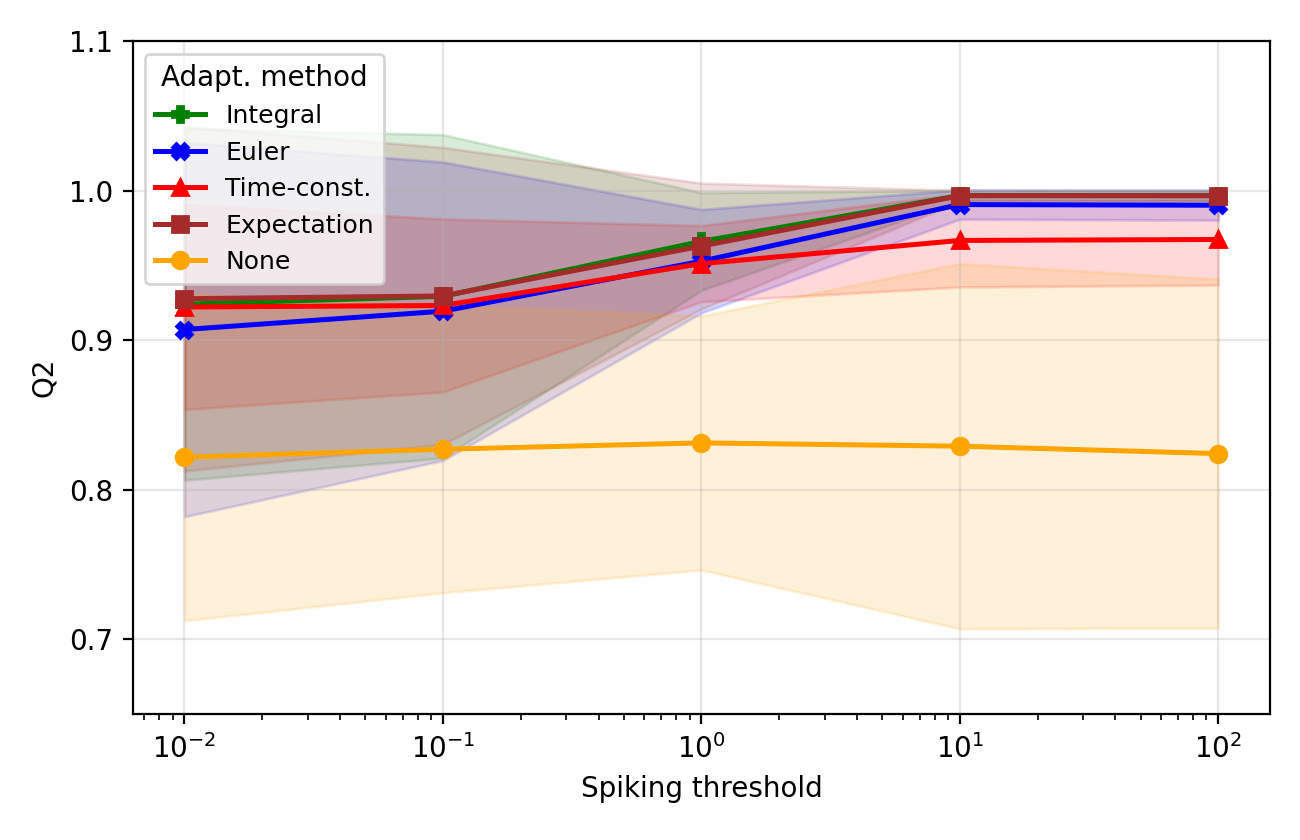}
     \label{fig:SpkRate:Fine-to-Coarse:Q2}
     \end{subfigure}     
\caption{Fine-to-Coarse scenario for $\bS=1$ and $\bT=2$. The solid line indicates the mean, and the shaded region represents one standard deviation around the mean.}
\label{fig:SpkRate:Fine-to-Coarse}
\end{figure}

We consider a single adLIF neuron with sinusoidal inputs, as introduced in Section~\ref{sec:illustrative_example}. 
Same as in Section~\ref{sec:illustrative_example}, we initialize $100$ distinct adLIF neurons and generate $100$ different sine inputs for each neuron, averaging the results across all trials. Again, we evaluate the match in dynamics for both Fine-to-Coarse for $\bS=1$ and $\bT=2$ and Coarse-to-Fine for $\bS=2$ and $\bT=1$.
To control the spiking activity, hence the level of nonlinearity,  we vary the spiking threshold: higher thresholds reduce spiking activity, whereas lower thresholds increase it.
This effect is illustrated in Figures~\ref{fig:SpkRate:Coarse-to-Fine:T_spk} and \ref{fig:SpkRate:Fine-to-Coarse:T_spk}, where the spike rate denotes the mean number of output spikes per neuron over the input sequence length. In these figures, we report results for both randomly initialized neurons with $\bS$ source temporal resolution input sequences, and for the adapted neurons with $\bT$ target temporal resolution input sequences.

Based on Figures~\ref{fig:SpkRate:Coarse-to-Fine:T_spk} and \ref{fig:SpkRate:Fine-to-Coarse:T_spk}, we now discuss how the spike rate changes under model adaption.  
 We observe that for both the Coarse-to-Fine and Fine-to-Coarse cases, the no-adaptation baseline and all three proposed methods yield to spike rates similar to the spike rate obtained with inputs at the source resolution.
In contrast, the Time-const. baseline produces relatively different spike-rate values. However, as shown earlier in Table~\ref{tab:sub:H2L_SHD_Dataset}, the Time-const. method can still outperform no-adaptation. This indicates that  the spike-rate is not necessarily a  good indicator of the performance of the adaptation methods.

Next, we discuss the match of neuron dynamics of the original and adapted neurons at different spiking thresholds; and hence different spiking levels. As in Section \ref{sec:illustrative_example}, we use the quality functions $\Qone$ and $\Qtwo$, where values close to $1$ indicate better reconstruction. Results for the Coarse-to-Fine case are shown in Figures~\ref{fig:SpkRate:Coarse-to-Fine:Q1} and \ref{fig:SpkRate:Coarse-to-Fine:Q2}, and for the Fine-to-Coarse case in Figures~\ref{fig:SpkRate:Fine-to-Coarse:Q1} and \ref{fig:SpkRate:Fine-to-Coarse:Q2}. 
Across all methods, higher thresholds generally lead to improved matches, while the no-adaptation baseline shows either no significant change or no trend that is consistent in all plots. 
The higher quality values for the proposed methods under higher thresholds are consistent with the fact that under lower spike rates, the neuron dynamics is closer to  linear, leading to a better performance of our proposed methods. This effect is most evident at the threshold of $10^2$ where the spiking rate drops to zero, and both $\Qone \approx 1$ and $\Qtwo \approx 1$. 

These experiments indicate that for the proposed adaptation methods, trained networks which have relatively lower spike rates may be more suitable. Hence,  the performance of the proposed methods may be improved by preferring such networks while training on source data,  e.g.,  by varying the spiking threshold or introducing regularization for spike rates. 

\subsection{Another Linearization of Neuron Dynamics}
\label{sec:deviation_Hf_in_A}

In the proposed adaptation methods of this article, we interpret the spike of the previous time step, i.e., $\Sil{t}$, as an external input to the neuron state and use the correspondence $\Abf=[\Hbfv]$, and $\Bbf=\begin{bmatrix}\Hbfs, \Hbff, \Hbfr\end{bmatrix}$, see Section~\ref{sec:map_snn_lssm}. However, as pointed out in Section \ref{sec:map_snn_lssm}, an alternative approach is to linearize the spiking function $g_{\Thetabf}(\cdot)$ and use $\Abf=\Hbfv + \Hbfs \Gbf_{\Thetabf}$, and $\Bbf=\begin{bmatrix}\Hbff, \Hbfr\end{bmatrix}$. 
In this section, we explore this approach.

Linearization of  $\Sil{t} = g_{\Thetabf}(\vbf[t])$ of \eqref{eqn:general_LIF_intro:line2} depends on the particular form of the function.   We now focus on the adLIF neuron   with $\Sil{t} = g_\theta(\Umem{t}) $ where $g_\theta(\cdot)$ is the Heaviside function with threshold $\theta$, and $\Umem{t}$ is the first component of the state vector $\vbf[t]$,   see \eqref{eqn:adLIFneuron_intro} and \eqref{eqn:adLIF_H_matrices}.
To linearize $g_\theta(\Umem{t})$,  one may use different approaches. Here, we  adopt   $\Sil{t} \approx 0.5 \Umem{t}$.  Hence, we now have 
\begin{align}
\label{eqn:general_LIF_rearange:new}
\! \vbf[t+1] &\approx  \Hbfv \vbf[t] +  \Hbfs  \begin{bmatrix} 0.5, & 0 \end{bmatrix}  \vbf[t] 
           +      \begin{bmatrix} \Hbff, \Hbfr  \end{bmatrix}
            \begin{bmatrix} \Wbf \Slbefore{t} \\ \Vbf \Sl{t}  \end{bmatrix}. 
\end{align}
Using \eqref{eqn:adLIF_H_matrices} and  \eqref{eqn:general_LIF_rearange:new},  we obtain the following mapping instead of the mapping in \eqref{eqn:adLIF_AandB} 
\begin{align}
\label{eqn:adLIF_AnadB:newlinearization}
\Abf \!=\! \begin{bmatrix}\alpha - 0.5\alpha \theta & -(1-\alpha)\\ a+0.5b &\beta\end{bmatrix}, \,\,
\Bbf \!=\! \begin{bmatrix} 1-\alpha & 1-\alpha \\ 0 & 0\end{bmatrix}. 
\end{align}

\begin{table*}
\centering
\small
\newcolumntype{l}{>{\centering\arraybackslash}p{2.5cm}}
\newcolumntype{m}{>{\centering\arraybackslash}p{4cm}}
\caption{Performance of adaptation methods on the SHD dataset under alternative linearization of neuron dynamics.}
\label{tab:Ablation_study:Alternative_linearization}
\resizebox{\textwidth}{!}{%
\begin{tabularx}{1\linewidth}{mllll}
\hline
    & \multicolumn{2}{c}{$\bS=2$ to $\bT=1$} 
    & \multicolumn{2}{c}{$\bS=1$ to $\bT=2$} \\
\cmidrule(lr){2-3} \cmidrule(lr){4-5}
Model Adapt. & adLIF & RadLIF & adLIF & RadLIF \\
\hline
Expectation &66.1 $\pm$ 2.6 \% & 41.4 $\pm$ 9.4 \% & 25.5 $\pm$ 3.9 \% & 21.7 $\pm$ 2.2 \% \\
Integral &66.1 $\pm$ 2.6 \% & 41.4 $\pm$ 9.4 \% & 25.5 $\pm$ 3.9 \% & 21.7 $\pm$ 2.2 \% \\
Euler &47.3 $\pm$ 3.2 \% & 17.1 $\pm$ 8.5 \% & 50.3 $\pm$ 4.5 \% & 44.1 $\pm$ 6.1 \% \\
\hline
\end{tabularx}
}
\end{table*}

Table~\ref{tab:Ablation_study:Alternative_linearization} presents the performance under  \eqref{eqn:adLIF_AnadB:newlinearization} and application of the adaptation methods using Section~\ref{sec:propositions} for the SHD dataset. 
Firstly, we discuss the Coarse-to-Fine scenario ($\bS=2$ to $\bT=1$), and compare with Table \ref{tab:sub:L2H_SHD_Dataset}. We observe that there is a significant drop in performance of all three adaptation approaches  under the alternative linearization in \eqref{eqn:adLIF_AnadB:newlinearization}. For example, a drop from $89.5 \%$ to $66.1 \%$  is observed for the Integral method. However, even with this performance drop, the Exponential and Integral method still outperform both the no-adaptation scenario and Time-const method.
%
Now we focus on the Fine-to-Coarse scenario ($\bS=1$ to $\bT=2$), and compare the results of Table~\ref{tab:Ablation_study:Alternative_linearization} with the results of Table \ref{tab:sub:H2L_SHD_Dataset}. We again observe a drop of performance under the alternative linearization of \eqref{eqn:adLIF_AnadB:newlinearization}. Here, in some cases the performance of adaptation is even worse than that of  no-adaptation and Time-const. method. 
We note that performance depends on the approach utilized for linearization of the non-linear spiking function;  and  other linearization approaches may improve  the performance. 

In conclusion, in the setting considered in this experiment, for both Coarse-to-Fine and Fine-to-Coarse scenarios, it is preferable to use the suggested correspondence in Section \ref{sec:map_snn_lssm}, i.e.,  treating the spike from the previous time step as external input to the neuron state as in \eqref{eqn:adLIF_AandB}, rather than the alternative linearization of \eqref{eqn:adLIF_AnadB:newlinearization}.

\section{Discussions}
\label{sec:discussions}
We now further discuss our results including the limitations of the proposed approaches. 

As noted in Section \ref{sec:general_neuron}, a variety of neuron models can be expressed within the generalized neuron model framework of Section~\eqref{eqn:general_LIF_intro}, which allows application of our proposed methods. However, in this paper, our numerical results focus on the adLIF model, as it provides a good balance between the simplicity of the basic LIF model and the complexity of more detailed biophysical models, while also being widely used in the machine learning–oriented SNN literature, e.g. see \cite{yik2024neurobench}.  
Nevertheless, investigating the performance of the proposed methods under other neuron models is a valuable direction for future research, as it will clarify the scope of their applicability in scenarios where models other than adLIF may be preferred.

As quantified in Sections \ref{sec:results:L2H} and \ref{sec:results:H2L}, the greater the difference between the resolution of source and target, the less successful the proposed adaptation methods are. This is consistent with the fact that the mismatch between the characteristics of the source and target data increases with increasing $\rho$.

The proposed model adaptation methods consistently outperform the existing model parameter adaptation methods benchmarks in both Coarse-to-Fine and Fine-to-Coarse settings. In the case of Fine-to-Coarse settings, the input adaptation methods we propose can outperform the model adaptation methods. This comes at the cost of longer inference time that scales scales with the ratio of the resolutions and has to be performed for each sample seen during inference. 

Our adaptation methods operate under the assumption that data is represented by events accumulated within fixed temporal windows, $\DeltaS$ for source data and $\DeltaT$ for target data. In such approaches, each time point in data sequence corresponds to a specific temporal interval used in accumulation of events. 
However, there are event representation formats where this relationship does not hold in a straightforward manner. 
For example, voxel-grid \cite{xu2025endtoendneuromorphiceventbased3d, chen2023dense} may include weighted interpolation of a set of events in the time domain, and time-surface \cite{lagorce2016hots, sironi2018hats} may encode recency via an exponential decay kernel at each pixel. In these cases, our methods are not directly applicable. An interesting avenue for future research is to investigate whether the use of intermediate fixed temporal-window based representations in the data processing chain, together with our adaptation methods, can yield satisfactory results for such cases.

An open question is the robustness of our adaptation methods to hardware non-idealities such as limited-precision representation of neuron parameters and synaptic weights, imperfect state updates, and general device noise. Since adaptation rules are derived under ideal device assumptions, their sensitivity to such effects remains to be studied. Exploring these aspects and possible mitigation strategies constitute an important future research direction to enable practical applications of our framework. 
\section{Conclusions}
\label{sec:conclusions}
We have established a bridge between SNNs and both linear and non-linear SSMs. We have formulated the problem of temporal domain adaptation for SNNs under an SSM framework.  We have proposed novel temporal domain adaptation methods using insights from the SSM framework.
These methods do not require access to data with the target resolution. 
The proposed methods significantly outperformed the existing primary approach in the current literature for temporal domain model adaptation for SNNs for all scenarios where the model was trained on data with coarse temporal resolution and used on data with fine temporal resolution. 

The proposed methods can be applied to any SNN neuron model. This addresses an important technical gap for time resolution adaptation in SNNs. In particular,  the existing method in the literature is best suited to neuron models with explicit dependence  on the temporal resolution, which is satisfied only for simpler neuron models.

Our results show that the proposed methods can be used to enable time-efficient training. In particular,  high accuracy on data with fine temporal resolution can be obtained by first training on coarse temporal resolution data, hence with a lower training time,  and then performing model adaptation using the proposed methods. 

Our findings illustrate that working with datasets with static origins, e.g. NMNIST, provides limited insight into importance of time resolution, since such datasets do not have the  temporal complexity inherent in real-world dynamic data. Hence, considering audio datasets where temporal dynamics play an important role, our work provides a  significant contribution to the literature on temporal resolution adaptation in contrast to approaches that focus solely on data with static origin.

The starting point for our framework is to match the neuron voltage dynamics on a single neuron level across different temporal resolutions. Investigation of spike-oriented approaches that focus on matching the spiking behavior on a layer level is considered a promising direction for future work. 
Other interesting research directions include exploring the performance of our proposed methods with other popular neuron models,
developing specialized temporal model adaptation techniques for other popular neuron model families, and evaluating the performance on non-ideal hardware.

\section{Acknowledgment}

This work was supported in part by   Swedish Research Council through grant agreement no. 2024-05194. 
S. Karilanova and A. Özçelikkale acknowledge the support of Centre for Interdisciplinary Mathematics (CIM), and AI4Research, both at Uppsala University. 

The computations were enabled by resources provided by the National Academic Infrastructure for Supercomputing in Sweden (NAISS), partially funded by the Swedish Research Council through grant agreement no. 2022-06725.

We thank Prof. Mikael Sternad from Uppsala University for his invaluable feedback on the manuscript and insightful discussions which contributed to the improvement of this work. We thank Petra Leferink for her parallel master's thesis on fine-tuning SNNs for domain adaptation, and Prof. Elisabetta Chicca for her insightful supervision.
%
\bibliographystyle{elsarticle-num} 
\bibliography{references} 
\appendix
\section{Appendix}

\subsection{Propositions for Proposed Temporal Resolution Domain Adaptation Methods}
\label{sec:propositions}
We now present our temporal-resolution domain adaptation methods as propositions and their proofs. 
In this section, the methods are developed using the standard linear SSM notation for the sake of clarity.  In Section~\ref{sec:proposedMethods:main}, we present them using the notation of the generalized SNN neuron model in \eqref{eqn:general_LIF_intro} based on the correspondence drawn in Section~\ref{sec:map_snn_lssm}.

We now provide some preliminaries. 
A continuous-time linear SSM can be represented as \cite{Gajic}:
\begin{subequations}
\label{eqn:lssm:cont-time}
\begin{align}
\frac{d}{d t} \vbf(t) & =\Abfc \vbf(t)+\Bbfc \fbf(t) \label{eqn:lssm:cont-time_line1} \\
\ybf(t) & =\Cbfc \vbf(t)+\Dbfc \fbf(t) \label{eqn:lssm:cont-time_line2}
\end{align}
\end{subequations}
where the vectors $\vbf(t), \fbf(t), \ybf(t)$ denote the state vector, input vector and output vector, respectively. Analogous to the discrete SSM defined in \eqref{eqn:lssm:generic}, the matrix $\Abfc \in \R^{n \times n}$ describes the internal behaviour of the system, while matrices $\Bbfc \in \R^{n \times r}, \Cbfc \in \R^{p \times n}, \Dbfc \in \R^{p \times r}$ represent connections between the input and the system, and the output.

We now define two methods from the literature used to approximate the continuous SSM defined in \eqref{eqn:lssm:cont-time} under a sampling period $\ctRo$ with the discrete SSM defined in \eqref{eqn:lssm:generic}: the Integral Approximation method, and the Euler method. The aim of both discretization methods is to find $\Abf, \Bbf, \Cbf, \Dbf$ matrices in \eqref{eqn:lssm:generic} such that the dynamics of the linear SSM and the continuous SSM are closely aligned. For clarity in this appendix, we denote these matrices with the subscript $r$, i.e. $\AbfRo, \BbfRo, \CbfRo, \DbfRo$ instead of $\Abf, \Bbf, \Cbf, \Dbf$.

\textbf{The Integral Approximation method} for discretization is based on the assumption that the system's input is constant during the given sampling period. Hence, a discrete-time linear SSM is obtained such that \cite{Gajic}:
\begin{subequations}
\label{eqn:lssm:integral}
\begin{align}
    \AbfRo &= e ^{\Abfc \ctRo} \label{eqn:integral_line1}\\
    \BbfRo &= (\AbfRo - \Ibf) \Abfc^{-1} \Bbfc \label{eqn:integral_line2} \\
    \CbfRo &= \Cbfc \\
    \DbfRo &= \Dbfc, 
\end{align}
\end{subequations}
which is valid under the assumption that $\Abfc$ is invertible.

\textbf{The Euler method} for discretization is known as a less accurate but simpler method compared to the integral approximation method \cite{Gajic}. Euler method is based on the approximation of the first derivative at the time instant $t=k\ctRo$ \cite{Gajic}:
\begin{align}
    \dfrac{d}{dt} \vbf(t) 
    \approx 
    \frac{1}{\ctRo} \left( \vbf((k+1)\ctRo) - \vbf(k\ctRo) \right).
\end{align}
Applying this approximation of the derivative, a discrete-time linear SSM is obtained such that \cite{Gajic}:
\begin{subequations}
\label{eqn:lssm:euler}
\begin{align}
    \AbfRo &= \Ibf + \ctRo \cdot \Abfc \label{eqn:euler_line1}\\
    \BbfRo &= T \cdot \Bbfc \label{eqn:euler_line2} \\
    \CbfRo &= \Cbfc \\
    \DbfRo &= \Dbfc.
\end{align}
\end{subequations}

\begin{asmptn}
\label{asmptn:two_discrete_sys}
Consider two discrete-time linear SSM \eqref{eqn:lssm:generic}, with matrix parameters
$\AbfR{1},\BbfR{1},\CbfR{1},\DbfR{1}$, and  $\AbfR{2}$, $\BbfR{2}$, $\CbfR{2},\DbfR{2}$, with associated time resolution $\ctR{1}$ and $\ctR{2}$, respectively. Assume $\ctR{2} = \DeltaRatio \ctR{1}$ for $\DeltaRatio  \in \R_{+}$. Additionally we assume $\CbfR{1} = \CbfR{2}$, and $\DbfR{1} = \DbfR{2}$.
\end{asmptn}

\begin{prop} \textbf{Integral temporal adaptation method:}
Consider two discrete-time linear SSM denoted with subscript $r_1$ and $r_2$ as defined in Assumption \ref{asmptn:two_discrete_sys} obtained from the continuous-time linear SSM \eqref{eqn:lssm:cont-time} using the Integral Approximation method with sampling periods $\ctR{2}$ and $\ctR{1}$, respectively. Assume that $(\AbfR{1} - \Ibf)$ is invertible, then:
\begin{subequations}
\begin{align} 
 \! \AbfR{2} &= \AbfR{1} ^ \DeltaRatio \\
 \! \BbfR{2} &= (\AbfR{2} - \Ibf) (\AbfR{1} - \Ibf)^{-1} \BbfR{1} 
 \end{align}
\end{subequations}
\label{Integral_Method_Prop}
\end{prop} 
\textbf{Proof of Proposition \ref{Integral_Method_Prop}:}
By \eqref{eqn:integral_line1} and \eqref{eqn:integral_line2} that
\begin{subequations}
\begin{align}
\! \AbfR{2} &= e ^{\Abfc \ctR{2}}
    = e ^{\Abfc \DeltaRatio \ctR{1}} 
    = (e ^{\Abfc \ctR{1}})^\DeltaRatio
    = \AbfR{1} ^ \DeltaRatio\\
\! \BbfR{2} &= (\AbfR{2} - \Ibf) \Abfc^{-1} \Bbf
          = (\AbfR{2} - \Ibf) (\AbfR{1} - \Ibf)^{-1} \BbfR{1}.
\end{align}
\end{subequations}

\begin{prop} \textbf{Euler temporal adaptation method:}
Consider two discrete-time linear SSM denoted with subscript $r_1$ and $r_2$ as defined in Assumption \ref{asmptn:two_discrete_sys} obtained from the continuous-time linear SSM \eqref{eqn:lssm:cont-time} using the Euler method with sampling periods $\ctR{2}$ and $\ctR{1}$, respectively. Then:
\begin{subequations}
\begin{align} 
 \AbfR{2} &= \Ibf + \DeltaRatio (\AbfR{1} - \Ibf) \\
 \BbfR{2} &= \DeltaRatio  \BbfR{1} 
\end{align}
\end{subequations}
\label{Euler_Method_Prop}
\end{prop}
\textbf{Proof of Proposition \ref{Euler_Method_Prop}:} It follows by \eqref{eqn:euler_line1} and \eqref{eqn:euler_line2} that
\begin{align}
    \AbfR{2} & = \Ibf + \ctR{2}  \Abfc 
            = \Ibf + \DeltaRatio \ctR{1} \Abfc  \\
            &= \Ibf + \DeltaRatio ((\ctR{1} \Abfc + \Ibf) - \Ibf)
            = \Ibf + \DeltaRatio ( \AbfR{1} - \Ibf)\\
    \BbfR{2} &= \ctR{2}  \Bbfc
            = \DeltaRatio \ctR{1} \Bbfc  
            = \DeltaRatio (\ctR{1}  \Bbfc) 
            = \DeltaRatio \BbfR{1}.
\end{align}

\begin{prop} 
\textbf{Expectation ~temporal ~adaptation}\\
\textbf{method:}
Consider two discrete-time linear SSM denoted with the subscripts $r_1$ and $r_2$ as defined in Assumption \ref{asmptn:two_discrete_sys} with the input sequences $\fbf[n]$ for $n \in {1,2,..., \dtR{1}}$, and $\fbfbar[m]$ for $m \in {1,..., \dtR{2}}$, respectively.  Here, $\NRatio N_{r_2}=N_{r_1}$,  $\NRatio \in \Z_{+}$.  Assume that $\E[\fbfbar[m]] = \E\left[\fbf[(m+1)\NRatio-k]\right] = \mubfx$ for $k=0,...,\NRatio-1$ for all $m$. Then, the following holds:
\begin{enumerate}
    \item If 
    \begin{subequations} \label{eqn:prop:Method_Expectation_Prop:1to2}
    \begin{align} 
     \AbfR{2} &= \AbfR{1}^{\NRatio} \\
     \BbfR{2} &= \sum^{\NRatio}_{j=1}\AbfR{1}^{\NRatio-j} \BbfR{1},
     \end{align}
     \end{subequations}
     then $\E[\vbfR{2}[m]] = \E[ \vbfR{1}[m\NRatio]]$ for all m. 
     
     \item Assume  that $(\sum^{\NRatio}_{j=1}\AbfR{2}^{\NRatio-j})$ is invertible. If 
     \begin{subequations} \label{eqn:prop:Method_Expectation_Prop:2to1}
     \begin{align} 
     \AbfR{1} &= \AbfR{2}^{\frac{1}{\NRatio}} \\
     \BbfR{1} &= (\sum^{\NRatio}_{j=1}\AbfR{1}^{\NRatio-j})^{-1} \BbfR{2}
     \end{align} 
     \end{subequations}
     then $\E[\vbfR{2}[m]] = \E[ \vbfR{1}[m\NRatio]]$ for all m. 
\end{enumerate}
\label{Method_Expectation_Prop}
\end{prop}
\textbf{Proof of Proposition \ref{Method_Expectation_Prop}}
We first express $\vbfR{1}[n+\NRatio]$ in terms of $\vbfR{1}[n]$ by performing $\NRatio$ recursive steps:
\begin{align}
\! \vbfR{1}[n+\NRatio] 
&= \AbfR{1} \vbfR{1}[n+\NRatio-1] + \BbfR{1} \fbf[n+\NRatio-1] \\
\! \vbfR{1}[n+\NRatio] 
&= \AbfR{1} (\AbfR{1} \vbfR{1}[n+\NRatio-2] + \BbfR{1} \fbf[n+\NRatio-2]) \notag \\
                                    & \quad + \BbfR{1} \fbf[n+\NRatio-1] \\    
&\shortvdotswithin{=}
\vbfR{1}[n+\NRatio] 
&= \AbfR{1}^{\NRatio} \vbfR{1}[n]+ \sum^{\NRatio}_{j=1}\AbfR{1}^{j-1} \BbfR{1} \fbf[n+\NRatio-j] 
\label{eqn:sum_b_steps}
\end{align}
We now consider the constraint on the expected dynamics of the state variable  i.e., $ \E[\vbfR{2}[m+1]] = \E[ \vbfR{1}[(m+1)\NRatio]]$. Rewriting both sides of the equation, we obtain  
\begin{align}
    & \quad \quad \E \left[\vbfR{2}[m+1]] = \E[ \vbfR{1}[(m+1)\NRatio] \right] \\
    &\E \left[ \AbfR{2} \vbfR{2}[m] + \BbfR{2} \fbfbar[m] \right]  = \notag \\
    & = \E \left[ \AbfR{1}^{\NRatio} \vbfR{1} \left[ m\NRatio \right]
    + \sum^{\NRatio}_{j=1}\AbfR{1}^{j-1} \BbfR{1} \fbf \left[(m+1)\NRatio-j \right] \right]  \\
    &\AbfR{2} \E \left[\vbfR{2}[m] \right] + \BbfR{2} \mubfx  =\notag \\
    & = \AbfR{1}^{\NRatio} \E \left[ \vbfR{1}[m\NRatio] \right]
    + \sum^{\NRatio}_{j=1}\AbfR{1}^{\NRatio-j} \BbfR{1} \mubfx 
\end{align}

Hence, if \eqref{eqn:prop:Method_Expectation_Prop:1to2} holds,  then $\E[\vbfR{2}[m]] = \E[ \vbfR{1}[m\NRatio]]$ and $\E[\vbfR{2}[m+1]] = \E[ \vbfR{1}[(m+1)\NRatio]]$.  Due to the generality of the previous equations, this holds for all $m \in {1,..., \dtR{2}}$. 
Similiarly, if $(\sum^{\NRatio}_{j=1}\AbfR{2}^{\NRatio-j})$ is invertible and \eqref{eqn:prop:Method_Expectation_Prop:2to1} holds, we have $\E[\vbfR{2}[m]] = \E[ \vbfR{1}[m\NRatio]]$ for all $m$.
This concludes the proof of Proposition~\ref{Method_Expectation_Prop}.

\begin{remark}
In Section~\ref{sec:proposedMethods:main}, we use \eqref{eqn:prop:Method_Expectation_Prop:1to2} and  \eqref{eqn:prop:Method_Expectation_Prop:2to1}, for adaptation in Fine-to-Coarse and Coarse-to-Fine scenarios, respectively. These equations are presented separately in Proposition~\ref{Method_Expectation_Prop} for clarity. 
\end{remark}

\begin{remark}\label{remark:ExpectationIntegral:assumptions}
We now compare Proposition~\ref{Method_Expectation_Prop}, i.e. Expectation temporal adaptation method, and Proposition~\ref{Integral_Method_Prop}, i.e.  Integral temporal adaptation method, in terms of their setting. 
Proposition \ref{Method_Expectation_Prop} assumes constant expected value of the input over an interval i.e. $E\left[\fbf[(m+1)\NRatio-k]\right]=\mubfx $ for $k=0,1,..,\NRatio$ for some arbitrary $\mubfx \in \R^{1 \times 1}$. 
Given this assumption, its goal is to match the expected values of the state variable at the time points of the coarse resolution, i.e. $\E[\vbfR{2}[m]] = \E[ \vbfR{1}[m\NRatio]]$.
On the other hand, the integral approximation method which lies under Proposition \ref{Integral_Method_Prop}, assumes that the input is constant over the interval i.e. $\fbf[(m+1)\NRatio-k]=c_b$ for $k=0,1,..,\NRatio$ for some arbitrary $c_b \in \R^{1 \times 1}$  \cite{Gajic}. 
Given this assumption, Proposition \ref{Integral_Method_Prop} matches the values of the state variable at the time points of the coarse resolution, i.e. $\vbfR{2}[m]= \vbfR{1}[m\NRatio]$. 
Note that matching the input and state variable implies matching the average behaviour of the input and the state variable.
Hence, the assumptions and the goals in the Integral temporal adaptation method imply the assumptions and the goals in the
Expectation temporal adaptation method. 
\label{remark:exp_int_input_assumption}
\end{remark}

\begin{remark}\label{remark:ExpectationIntegral:expressions}
Despite the different starting points for Expectation method, i.e., Proposition \ref{Method_Expectation_Prop}, and Integral method, i.e., Proposition \ref{Euler_Method_Prop}, the two methods are equivalent under the below assumptions:

Fine-to-Coarse deployment: Integral method makes the assumption that $\AbfR{1} - \Ibf$ is invertible, while Expectation method does not make such an assumption. However, if  $\AbfR{1} - \Ibf$ is also assumed to be invertible under Expectation method, then the two methods are equivalent. 
    
Coarse-to-Fine deployment: Integral method makes the assumption that $\AbfR{1} - \Ibf$ is invertible, while Expectation method assumes $(\sum^{\NRatio}_{j=1}\AbfR{2}^{\NRatio-j})$ is invertible. The two methods become equivalent if in the Expectation method, one also makes assumption that $\AbfR{1} - \Ibf$ and $\AbfR{2} - \Ibf$ is invertible.

Note that these equivalences are due  to the fact that  $\sum^{\NRatio}_{j=1}\Abf^{\NRatio-j}=(\Abf^{\NRatio}-\Ibf)(\Abf-\Ibf)^{-1}$ under  invertible $(\Abf-\Ibf)$,  $\NRatio \in \mathbb{Z}_+$.
\end{remark}

\subsection{Temporal domain adaptation for the normalization layers}
\label{sec:num:setting:batchnorm_scaling}

During training of the neural networks, normalization layers learn the statistics, i.e. the mean and variance, of source data, which are then typically used in inference without any change. On the other hand, in our setup, the model is used on target data, which has  a different time resolution. Hence, the statistics learned during training should be adjusted. In this section, we investigate how this adjustment should be made.

Denote a source sequence with $X_S=(x_1,x_2,...,x_{\NS})$ and the associated mean with $\muS = \frac{1}{\NS} \sum_{i=1}^{\NS} \E[x_i]$ and the variance with $\sigmaS^2 =\frac{1}{\NS} \sum_{i=1}^{\NS} \E[(x_i-\E[x_i])^2]$. Similarly, denote a target sequence with  $X_T=(x'_1,x'_2,...,x'_{\NT})$ and  the associated  mean and variance with $\muT$ and $\sigmaT^2$, respectively. Denote the ratio of the temporal resolutions with $\NRatio=\NT/\NS$.
Assume that the elements of source sequence, i.e., $x_i$ are i.i.d.

\subsubsection{Adaptation for sum binning transformation}
\label{sec:num:setting:batchnorm_scaling:sum_binning}
Let the target sequence be the sum-binned version of the source sequence i.e. $\NS=\NRatio \NT$ with $\NRatio \in \Z_{+}$. Then, under i.i.d. $x_i$, we have
$\muT = \E[x_i']=\E[ \sum^{\NRatio-1}_{j=0}  x_{i+j}]=\NRatio \E[x_i] = \NRatio\muS$, and 
$\sigmaT^2 = var(x_i') = var( \sum^{\NRatio-1}_{j=0} x_{i+j} ) = \NRatio^2 var(x_i) = \NRatio^2 \sigmaS^2$.

\subsubsection{Adaptation for repeat elements transformation}
\label{sec:num:setting:batchnorm_scaling:repeat_elems}
Let the target sequence be obtained by repeating each element of the source sequence  $\NRatio -1$ times. Then,
$\muT = \E[x'_i]=\E[x_i]= \muS$, and 
$\sigmaT^2 = var(x'_i) = var(x_i) = \sigmaS^2$.

\subsubsection{Adaptation for zero padding transformation}
\label{sec:num:setting:batchnorm_scaling:zero_pad}
Let the target sequence be obtained by adding $\NRatio-1$ zeros before each element of the source sequence. Then, 
$\muT =E[x_i'] =\frac{x_1+...+x_N}{\NRatio N}=\frac{\muS}{\NRatio}$, and using the identity $var(x_i')= E[x_i'^2] -(E[x_i'])^2$
\begin{align}
    \sigmaT^2         
      &=  \frac{1}{\NRatio \NS}\sum^{\NRatio \NS}_{i=1} \E[ x_i'^2] -  \frac{1}{\NRatio \NS}\sum^{\NRatio \NS}_{i=1} (\E[x_i'])^2 \\
    &=  \frac{1}{\NRatio \NS}\sum^{\NRatio \NS}_{i=1}\E[x_i'^2]  - (\frac{\muS}{\NRatio})^2 \\
    &= \frac{1}{\NRatio}  \frac{1}{\NS}\sum^{\NS}_{i=1} \E[(x_i)^2 ] - (\frac{\muS}{\NRatio})^2 \\
    &= \frac{1}{\NRatio} \left[ \sigmaS^2 + \muS^2  \right] - (\frac{\muS}{\NRatio})^2 \\
    &= \frac{1}{\NRatio} \sigmaS^2 + \frac{\muS^2(\NRatio-1)}{\NRatio^2}
\end{align}

\subsubsection{Temporal Domain adaptation in experiments}
\label{sec:norm_adapt_in_experiment}
We now present how adaptation of batch normalization in performed in the numerical experiments. 

In the Coarse-to-Fine scenario,
where the source data is the sum-binned data and the target data is the fine resolution data, we have $\NS=\NRatio \NT$ with $\NRatio<1$. 
The adaptations are approximated using the transformations in Section \ref{sec:num:setting:batchnorm_scaling:sum_binning}. Hence, we use $\muT = \NRatio\muS$ and $\sigmaT^2 = \NRatio^2 \sigmaS^2$. 
In the input adaptation of Section \ref{sec:input-adapt:L2H}, based on numerical experiments, we use $\muT = \NRatio\muS,\sigmaT^2=\NRatio \sigmaS^2$ for the max-pooling input adaptation method and $\muT = \NRatio\muS,\sigmaT^2=\NRatio^2 \sigmaS^2$ for the binary sum-binning input adaptation method.

In the Fine-to-Coarse scenario, where the source data is the fine data and the target data is the sum-binned resolution data, 
we have exactly the scenario of Section \ref{sec:num:setting:batchnorm_scaling:sum_binning}, and hence use $\muT = \NRatio\muS$ and $\sigmaT^2 = \NRatio^2 \sigmaS^2$.
In the Fine-to-Coarse experiments, for the input adaptation methods of Section~\ref{sec:input-adapt:H2L}, additional transformations are applied to the coarse target data such as repeat elements or zero padding, with the aim of mimicking fine resolution during inference. 
When the transformation for mimicking fine resolution is repeated elements, we use the scaling from the repeated elements (Section \ref{sec:num:setting:batchnorm_scaling:repeat_elems}), in addition to the sum binning (Section \ref{sec:num:setting:batchnorm_scaling:sum_binning}). Hence, we obtain $\muT = \NRatio\muS$ and $\sigmaT^2 = \NRatio^2 \sigmaS^2$.
When the transformation for mimicking fine resolution is zero-padding, we have repeated elements (Section \ref{sec:num:setting:batchnorm_scaling:zero_pad}), in addition to the sum binning (Section \ref{sec:num:setting:batchnorm_scaling:sum_binning}). Hence, we obtain the theoretical scaling as  $\muT = \NRatio\muS$ and $\sigmaT^2 = \NRatio\sigmaS^2 + \muS^2(\NRatio-1)$.
However, experiments showed poor performance with this theoretical scaling of the variance. Hence, we use $\sigmaT^2=\sigmaS^2$ in our results in Section~\ref{sec:numericalResults}.

For the transformations introduced in Section~\ref{sec:various_to_coarse_tranformations}, based on our numerical experiments, we use the following parameter correspondences: For binary-sum-bin and max-pool, we use $\muT = \NRatio\muS$ and $\sigmaT^2 = \sigmaS^2$. For first-point-sample, last-point-sample, and average-sum-bin, we use $\muT = \muS$ and $\sigmaT^2 = \sigmaS^2$. These relations are employed in both Coarse-to-Fine and Fine-to-Coarse scenarios.

Note that, in both Coarse-to-Fine and Fine-to-Coarse scenarios, we only adapt the first normalization layer in the network. This initial adjustment is considered to correct the shift introduced by the data transformation, reducing the need for additional scaling in subsequent layers.

\subsection{Performance criteria for assessing matching neuron dynamics under varying time resolution}
\label{sec:num:setting:quality_fnc}

Given time sequence $\vsig{1} \in \R^{\dtN{1}}$ with $\dtN{1}$ number of time steps, and signal $\vsig{2}  \in \R^{\dtN{2}}$ with $\dtN{2}$ number of time steps where we have either $\dtN{1}=\NRatio \dtN{2}$ or  $\dtN{2}=\NRatio \dtN{1}$ where $\NRatio \in \Z_{+}^{1 \times 1}{+}$. 
Our objective is to estimate the goodness of match between these two sequences of different temporal resolutions are describing assuming the $\vsig{1}$ is the true sequence.
Since the performance criteria functions that we will use require the two signals we compare to be of same time length, we uniformly sample the higher time resolution signal by taking every $\NRatio$-th sample. For demonstration we focus on the case where $\vsig{1}$ is of higher time resolution i.e. $\dtN{1}=\NRatio \dtN{2}$, hence we create the sampled signal $\vsigHat{1}$ such that $\vsigHat{1}[n]=\vsig{1}[\NRatio n]$ for $n=1,...,\dtN{2}$. Then the performance criteria functions are defined as
\begin{align}
    \Qone(\vsigHat{1}, \vsig{2})&= 
         1-\frac{\msefnc(\vsigHat{1}, \vsig{2} )}
               {\varfnc(\vsigHat{1})}  
\\
    \Qtwo(\vsigHat{1}, \vsig{2})&= 
        \frac{\covfnc(\vsigHat{1}, \vsig{2})}
             {\sqrt{\varfnc(\vsigHat{1}) \varfnc(\vsig{2})}}
\end{align}
where
\begin{align}
\msefnc(\vsigHat{1}, \vsig{2}) 
&= \frac{1}{\dtN{2}} \sum^{\dtN{2}}_{n=1} (\vsigHat{1}[n] - \vsig{2}[n])^2
\\
\covfnc(\vsigHat{1}, \vsig{2}) 
& = \frac{1}{\dtN{2}-1} \sum^{\dtN{2}}_{n=1} 
(\vsigHat{1}[n] - \overline{\vsigHat{1}})
(\vsig{2}[n] - \overline{\vsig{2}}).
\end{align}
Here $\overline{\vsigHat{1}}$ and $\overline{\vsig{2}}$ are the mean of $\vsigHat{1}$ and $\vsig{2}$, respectively, and  $\varfnc$ is the variance. 
Here, $\Qone$ gives the relative square error and $\Qtwo$ gives the correlation coefficient, hence both provide different measures of similarity between $\vsigHat{1}$ and $\vsig{2}$. When there is significant matching between $\vsigHat{1}$ and $\vsig{2}$, MSE is expected to be small compared to variance, and  both $\Qone$ and $\Qtwo$ are expected to take values $|\Qone|, |\Qtwo|\leq 1$ where $\Qone \approx 1$ and $\Qtwo \approx 1$ indicates better reconstruction compared to $\Qone \approx 0$ and $\Qtwo \approx 0$, respectively. Specifically $\Qone=1$ corresponds to perfect match, while $\Qtwo=1$ corresponds to perfect linear relationship between the two sequences.

\subsection{Hyperparameters}
\label{sec:HPO}

For the adLIF parameters, $\alpha, \beta, a$, and $b$ are uniformly initialized in the regions 
$[\exp(-\frac{1}{5}), -\frac{1}{25})]$,
$[\exp(-\frac{1}{30}), -\frac{1}{120})]$,
$[0, 1)$, and $[0, 2]$,  respectively. The clipping ranges while training for $a$ and $b$ are the same as their initialization ranges, while for both $\alpha$ and $\beta$ the clipping range used is $[0, 1)$.
In the training process, we use Adam optimizer with weight-decay$=0.0001$ and Step Learning Rate scheduler of base learning rate $0.01$, step-size $10$ and $\gamma=0.1$.

\end{document}